\documentclass[11pt]{report}
\usepackage[utf8]{inputenc} 
\usepackage[T1]{fontenc} 
\usepackage{mathpazo} 
\usepackage{bigfoot}
\usepackage{setspace}
\usepackage{amssymb}
\usepackage{amsmath}
\usepackage{mathrsfs}
\usepackage[skip=5pt plus1pt]{parskip}
\usepackage{graphicx}
\usepackage{listings}
\usepackage[dvipsnames]{xcolor}
\usepackage{sectsty}
\usepackage{booktabs}
\usepackage{ragged2e}
\chapterfont{\color{BrickRed}} 
\sectionfont{\color{BrickRed}}
\subsectionfont{\color{BrickRed}}
\subsubsectionfont{\color{BrickRed}}
\definecolor{codegreen}{rgb}{0,0.6,0}
\definecolor{codegray}{rgb}{0.5,0.5,0.5}
\definecolor{codepurple}{rgb}{0.58,0,0.82}
\definecolor{backcolour}{rgb}{0.95,0.95,0.92}
\lstdefinestyle{mystyle}{
    backgroundcolor=\color{backcolour},   
    commentstyle=\color{codegreen},
    keywordstyle=\color{magenta},
    numberstyle=\tiny\color{codegray},
    stringstyle=\color{codepurple},
    basicstyle=\ttfamily\footnotesize,
    breakatwhitespace=false,    
    breaklines=true,            
    captionpos=b,               
    keepspaces=true,            
    numbers=left,           
    numbersep=5pt,          
    showspaces=false,           
    showstringspaces=false,
    showtabs=false,             
    tabsize=2
}
\usepackage{hyperref}
\usepackage{hyperref}
\hypersetup{
  colorlinks   = true, 
  urlcolor     = BrickRed, 
  linkcolor    = BrickRed, 
  citecolor   = BrickRed 
}
\begin{document}

\begin{titlepage} 
	\newcommand{\HRule}{\rule{\linewidth}{0.5mm}} 
	\center
	\HRule\\[0.4cm]
	{\huge\bfseries \textcolor{BrickRed}{Dynamic Normativity}}\\[0.4cm]
	{\LARGE\bfseries \textcolor{BrickRed}{Necessary and Sufficient Conditions for Value Alignment}}\\[0.4cm] 
 
	\HRule\\[1.5cm]
        \begin{figure}[h]
        \centering
        \includegraphics[width=0.45\linewidth]{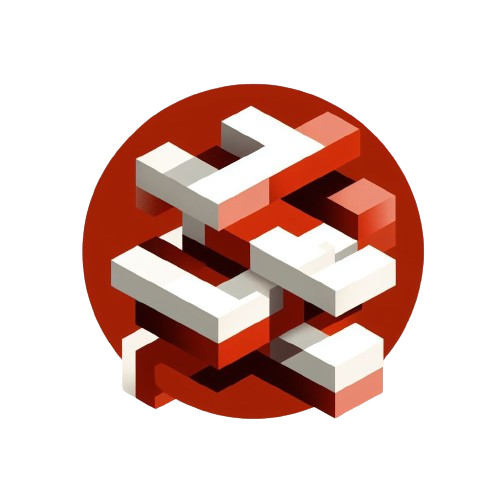}
        \label{fig:TeenyTinyLlama-logo}
        \end{figure}
    \vfill
    \textbf{\large\textsc{Nicholas Kluge Corrêa}}\\
    \vfill\vfill
	{\large 2024}
	\vfill
\end{titlepage}

\begin{titlepage} 
	\newcommand{\HRule}{\rule{\linewidth}{0.5mm}} 
	\center
	{\huge\bfseries Dynamic Normativity}\\[0.4cm]
	{\LARGE\bfseries Necessary and Sufficient Conditions for Value Alignment}\\[0.4cm]

    \vfill
    This inaugural dissertation was delivered by \textbf{Nicholas Kluge \textsc{Corrêa}} to the Institute of Philosophy of the University of Bonn (Germany) and the Graduate Program of Philosophy from the Pontifical Catholic University of Rio Grande do Sul (Brazil) in partial fulfillment of the requirements for the cotutelle degree of Doctor of Philosophy.
    \vfill
    \vfill
    {\large Bonn, 2024}
    \vfill
\end{titlepage}
\chapter*{Abstract}

\begin{singlespace}
    The critical inquiry pervading the realm of Philosophy, and perhaps extending its influence across all Humanities disciplines, revolves around the intricacies of morality and normativity. Surprisingly, in recent years, this thematic thread has woven its way into an unexpected domain, one not conventionally associated with pondering "what ought to be": the field of artificial intelligence (AI) research. Central to morality and AI, we find \textit{"alignment"}, a problem related to the challenges of expressing human goals and values in a manner that artificial systems can follow without leading to unwanted adversarial effects. More explicitly and with our current paradigm of AI development in mind, we can think of alignment as teaching human values to non-anthropomorphic entities trained through opaque, gradient-based learning techniques. This work addresses alignment as a technical-philosophical problem that requires solid philosophical foundations and practical implementations that bring normative theory to AI system development. To accomplish this, we propose two sets of necessary and sufficient conditions that, we argue, should be considered in any alignment process. While necessary conditions serve as metaphysical and metaethical roots that pertain to the permissibility of alignment, sufficient conditions establish a blueprint for aligning AI systems under a learning-based paradigm. After laying such foundations, we present implementations of this approach by using state-of-the-art techniques and methods for aligning general-purpose language systems. We call this framework \textit{\textcolor{BrickRed}{Dynamic Normativity}}. Its central thesis is that any alignment process under a learning paradigm that cannot fulfill its necessary and sufficient conditions will fail in producing aligned systems.
\end{singlespace}

\providecommand{\keywords}[1]
{
  \small	
  \textbf{\textit{Keywords---}} #1
}

\keywords{\textcolor{BrickRed}{Artificial Intelligence}, \textcolor{BrickRed}{Alignment}, \textcolor{BrickRed}{Value Learning}}

\chapter*{Resumo}

\begin{singlespace}
    A investigação crítica que permeia o campo da filosofia, e talvez estenda sua influência a todas as disciplinas de ciências humanas, gira em torno dos meandros da moralidade e da normatividade. Surpreendentemente, nos últimos anos, esse fio temático foi inserido em um domínio inesperado, que não é convencionalmente associado à reflexão sobre "o que deve ser": o campo de pesquisa da inteligência artificial (IA). No centro da moralidade e da IA, encontramos o \textit{"alinhamento"}, um problema relacionado aos desafios de expressar metas e valores humanos de uma forma que os sistemas artificiais possam seguir sem causar efeitos adversos indesejados. De forma mais explícita e com nosso paradigma atual de desenvolvimento de IA em mente, podemos pensar no alinhamento como o ensino de valores humanos a entidades não antropomórficas treinadas por meio de técnicas de aprendizado opacas e baseadas em gradiente. Este trabalho aborda o alinhamento como um problema técnico-filosófico que requer fundamentos filosóficos sólidos e implementações práticas que tragam a teoria normativa para o desenvolvimento do sistema de IA. Para isso, propomos dois conjuntos de condições necessárias e suficientes que, segundo nosso argumento, devem ser consideradas em qualquer processo de alinhamento. Enquanto as condições necessárias servem como raízes metafísicas e metaéticas relacionadas à permissibilidade do alinhamento, as condições suficientes estabelecem um plano para alinhar os sistemas de IA sob um paradigma baseado em aprendizado. Depois de estabelecer essas bases, apresentamos implementações dessa abordagem usando técnicas e métodos de última geração para alinhar sistemas de linguagem de uso geral. Chamamos essa estrutura de \textit{\textcolor{BrickRed}{Dinâmica Normativa}}. Sua tese central é que qualquer processo de alinhamento sob um paradigma de aprendizagem que não possa cumprir suas condições necessárias e suficientes falhará na produção de sistemas alinhados.
\end{singlespace}

\providecommand{\keywords}[1]
{
  \small	
  \textbf{\textit{Keywords---}} #1
}

\keywords{\textcolor{BrickRed}{Inteligência Artificial}, \textcolor{BrickRed}{Alinhamento}, \textcolor{BrickRed}{Aprendizagem de Valor}}
\chapter*{Preface}

\begin{flushright}
\textit{"The beginning is the most important part of the work."}

\textcolor{BrickRed}{― Plato, The Republic}
\end{flushright}

We commenced this work in early 2021. Many things in the field have changed in the last four years, especially in 2022-2024. ChatGPT was released and gained 100 million users within 2 months. Future of Life Institute pleads for a \href{https://futureoflife.org/open-letter/pause-giant-ai-experiments}{moratorium on training runs}\footnote{\hspace{1mm}\includegraphics[scale=0.025]{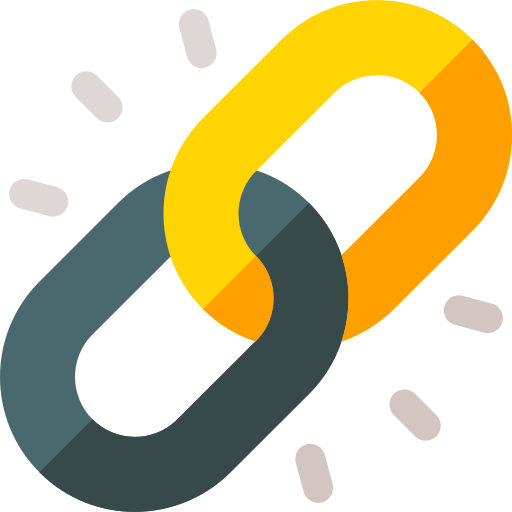}\hspace{1mm} \href{https://futureoflife.org/open-letter/pause-giant-ai-experiments}{futureoflife.org/open-letter/pause-giant-ai-experiments}} for AI systems stronger than GPT-4. The European Union passed the \href{https://artificialintelligenceact.eu}{AI Act}.\footnote{\hspace{1mm}\includegraphics[scale=0.025]{img/link.png}\hspace{1mm} \href{https://artificialintelligenceact.eu}{artificialintelligenceact.eu}} While AI systems have steadily broadened their scope of capabilities, including \href{https://openai.com/index/dall-e-3}{generating high-quality images},\footnote{\hspace{1mm}\includegraphics[scale=0.025]{img/link.png}\hspace{1mm} \href{https://openai.com/index/dall-e-3}{openai.com/index/dall-e-3}} \href{https://github.com/features/copilot}{writing code},\footnote{\hspace{1mm}\includegraphics[scale=0.025]{img/link.png}\hspace{1mm} \href{https://github.com/features/copilot}{github.com/features/copilot}} \href{https://openai.com/index/sora}{generating video},\footnote{\hspace{1mm}\includegraphics[scale=0.025]{img/link.png}\hspace{1mm} \href{https://openai.com/index/sora}{openai.com/index/sora}} and even tackling long-standing scientific problems like \href{https://deepmind.google/technologies/alphafold}{protein folding and predicting biomolecular interactions}.\footnote{\hspace{1mm}\includegraphics[scale=0.025]{img/link.png}\hspace{1mm} \href{https://deepmind.google/technologies/alphafold}{deepmind.google/technologies/alphafold}} 

Unlike philosophy's steady and slow currents, AI research jets forward rapidly. Consequently, many technical facets explored in this book are poised to become outdated as the field evolves. Yet, we aspire (perhaps very naively) for our philosophical underpinnings to endure longer. Therefore, we urge the reader to contextualize this work within the time frame of its creation. Much has transpired over the past four years. Alignment research has firmly entrenched itself in the mainstream. Unlike the bygone era of 2015-2020, when the subjects we will soon delve into were restricted to blog posts, unpublished manuscripts, and word-of-mouth lore, alignment has transcended mere academic curiosity. Nowadays, alignment research (to a great extent) has morphed into a force driving the development of products and services that are eagerly pursued by the market, and what was once a set of revolutionary and experimental ideas are now procedures accessible to anyone with minimal technical understanding, and almost no philosophical foundations. Nevertheless, this book is a testament to our endeavors throughout these years, offering a retrospective view of the journey that has led us to our current state.

\textit{Mea culpa} aside, in this work, we will address issues of Ethics and Normativity within the context of artificial intelligence, taking an applied stance toward the problems of the field.\footnote{By "applied", we mean that we will be taking an \textit{"ethics for design"} and \textit{"by design"} approach, as Virginia Dignum would put \cite{dignum_ethics_2018}, i.e., bringing normativity both to the human that builds and uses the machine as much as the machine that the human uses.}

But first, let us define Applied Ethics as the philosophical endeavor of applying normative ethical theories, i.e., theories that seek to differentiate right from wrong, to a specific context. In the case of the Ethics of AI (or AI Ethics for short), the applied context is the ethical evaluation of the development and use of AI systems and applications. Also, let us define the concept of AI research (outside the normative sphere) as the discipline involved in automating cognitive processes akin to those performed by humans, like our capabilities regarding visual perception and language understanding, among others. Don't worry; we will get back to \textit{"defining AI"}, in a more philosophically robust way, in later chapters.

Taking both a humanistic (i.e., philosophical) and technical (i.e., engineering-based) approach, this work strives for actual interdisciplinary development,\footnote{"The combination of multiple academic disciplines into one research activity" \cite{nissani1995fruits}.} though it remains for the reader to decide how well such a lofty goal has been approximated. As philosophers, we will seek to develop foundations for our work, taking inspiration from the many areas of Philosophy that support all other branches of knowledge, like Metaphysics and Metaethics. As Machine Learning (ML) Engineers, we will seek to implement our work to the best of our abilities, looking to bridge the gap between theory and practice.

Both approaches are complementary and essential to dealing with the problems we face. Both engineers and philosophers can (perhaps) agree that unaligned AI systems are undesirable and that an unaligned AI community is ill-equipped to solve its issues. But to address these problems, we first need to know what we mean when talking about AI, what we want to avoid, and where we wish to go. In the end, answering these questions, or simply attempting to, can help make AI research a more "humane" discipline. A goal that by itself should be enough.

Now, what can you expect as you leaf through the chapters of this book?

\begin{itemize}
    \item In \hyperref[chap1]{Chapter 1}, we will present an overview of AI Ethics as a field of active research, especially concerning how the normative discourse surrounding AI has taken form. This foundational section presents a descriptive and critical analysis of our current landscape and serves as an introduction to the uninitiated reader.
    \item In \hyperref[chap2]{Chapter 2}, we will dive into the land of unknown risks associated with Artificial General Intelligence (AGI). Given all the attention given to the possibility that AI may become an existential risk, we will bring these long-term concerns to our work and expose them to the reader.
    \item In \hyperref[chap3]{Chapter 3}, we will then define in technical/philosophical terms what we mean by \textit{"Alignment"} as a problem. Some of the questions we will address concern the limitations of our current paradigms in AI development and how Normative Ethics can help in this conundrum. At the end of this Chapter, we will present a set of necessary and sufficient conditions that will guide the rest of this project. This collection of conditions is what we will call \textcolor{BrickRed}{Dynamic Normativity}, which can be understood as a normative theory, or foundation, for developing aligned AI systems.
    \item In \hyperref[chap4]{Chapter 4}, we will then define our thesis's metaphysical and metaethical foundations. AI is a controversial term, and we must specify what we mean if this book is supposed (as we hope) to be a foundational work in the Philosophy of AI. At the same time, Chapter 4 will allow us to predefine our biases before diving deeper.
    \item While in Chapter 4, we defend the necessity and soundness of the conditions tied to Dynamic Normativity, the final three chapters (\hyperref[chap5]{5}, \hyperref[chap6]{6}, and \hyperref[chap7]{7}) present a minimal set of strategies and methodologies to tackle the value alignment problem. 
\end{itemize}

At many points in this book, the reader will be redirected to external materials, like dashboards, code repositories, demos, etc. (\includegraphics[scale=0.025]{img/link.png}), which should make its reading more dynamic and practical. In essence, this work sought to create tools and instructions so that other researchers could approach the alignment problem (and other ethical issues involving AI) on a practical/applied basis. While the book alone should be able to stand on its own regarding its narrative and intellectual constructions, there is much to gain if one is willing to open a browser and explore the additional content tied to this work.

Now, let us begin.

\tableofcontents
\chapter{Worldwide AI Ethics, Principles, and Blank Spots}
\label{chap1}

\begin{flushright}
\textit{"Once men turned their thinking over to machines in the hope that this would set them free. But that only permitted other men with machines to enslave them."}

\textcolor{BrickRed}{― Frank Herbert, Dune}
\end{flushright}

\section{Introduction}

Since immemorial times, humankind has been dealing with the "what should I do?" question. And at every step of the progress ladder, from fire to people walking on the moon, the normative uncertainty that moves us to question our actions and deeds has not gone away. These questions are usually tied to what we can do and undo. What we can build and what we can destroy. Hence, as technological advancements continue to shape our world and as we witness our lives becoming increasingly entangled in this process, for many (if not all), it remains uncertain whether this progress will ultimately align with our fundamental values and ideals (whatever these may be). In short, we cannot escape the normative conundrum, even though precisely defining what the "good life is" will be forever out of our reach.

However, there is much merit involved in grappling with these questions. In the context of technology and AI, we find a variety of disciplines, including sociology \cite{Hartmut_Rosa__and__William_E_Scheuerman_2009}, literature \cite{Mike_Davis_1992}, and philosophy \cite{feenberg2012questioning}, that relentlessly try to understand how our moral landscape is affected, and affects, the byproducts of our need to build and reshape the universe as "we" see fit.

Reviewing this entire field is too grand a feat for this book. However, this chapter will aim to present a descriptive analysis of a particular type of AI ethics discourse. More specifically, the part that has morphed into what we call "AI Guidelines", i.e., a kind of document that has been flooding the literature for a significant portion of the last 10 years. We can understand these guidelines as a distilled and concentrated version of the normativity emerging from the minds of those engaged in AI Ethics. Through this analysis, we aim to promote a critique of our current landscape while also presenting avenues for future work that we deem to be more prosperous.

More specifically, in Section \ref{world_wide_ai_ethics}, we will present a detailed analysis of the current state of AI ethics, drawing from an extensive literature review of the field entitled \textcolor{BrickRed}{Worldwide AI Ethics} (WAIE) \cite{correa2023worldwide}, which at the moment of this writing, remains as the largest meta-analysis of the field. In Section \ref{critical_analysis}, we will review the limitations of WAIE while presenting a critique of the field, exposing some of its current deficits and contradictions. Lastly, in Section \ref{way_forward}, we will narrow our focus to the academic paradoxes tied to the principle-practice gap in AI ethics while also proposing directions for future work and research in line with our vision of what AI ethics \textit{should be.}
 
\section{Introducing Worldwide AI Ethics}
\label{world_wide_ai_ethics}

Since the early 1990s, there has been a remarkable surge in both AI research and industry. This surge can be attributed to several factors. Firstly, the breakthrough success of Deep Learning (DL) has revitalized progress in areas where previous paradigms had reached stagnation. Secondly, advancements in hardware and our ability to perform computations at scale have played a pivotal role. Additionally, the availability of vast amounts of data has provided fertile ground for further advancements based on the learning paradigm. Furthermore, the exponential increase in investments, fueled by extensive media hype, is the cherry on top. All these factors combined have helped the field of AI to massively extend the capabilities of intelligent autonomous systems and expand the scope of what can be achieved through intelligent automation \cite{NIPS2012_c399862d, schmidhuber2015deep, goodfellow2016deep, chollet2021deep, prince2023understanding}.

\begin{figure}[htp]
    \includegraphics[width=\linewidth]{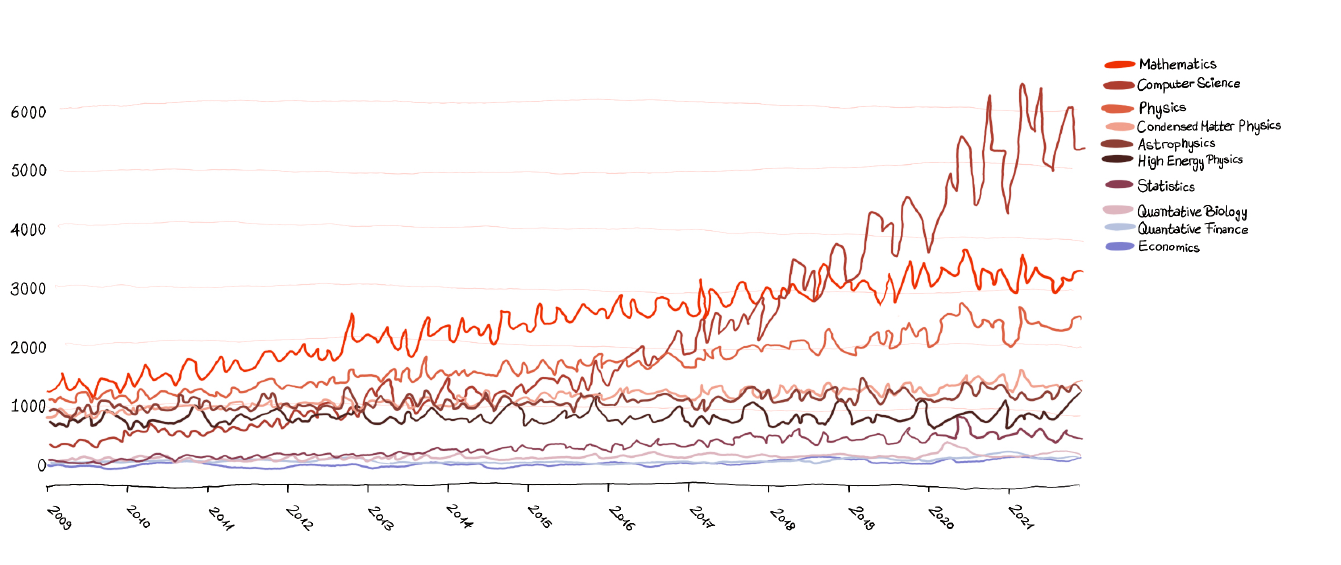}
    \caption{ArXiv submissions history (2009–2021).}
    \label{fig:arxiv-sub}
\end{figure}

For example, if we look at the submission history on ArXiv from 2009 to 2021 (Fig. \ref{fig:arxiv-sub}),\footnote{\hspace{1mm}\includegraphics[scale=0.025]{img/link.png}\hspace{1mm} \href{https://arxiv.org/about/reports/2021_usage}{arxiv.org/about/reports/2021\_usage}} we notice that since 2018, publications relating to computer science have been the most frequently submitted type of content.

\begin{figure}[htp]
    \includegraphics[width=\linewidth]{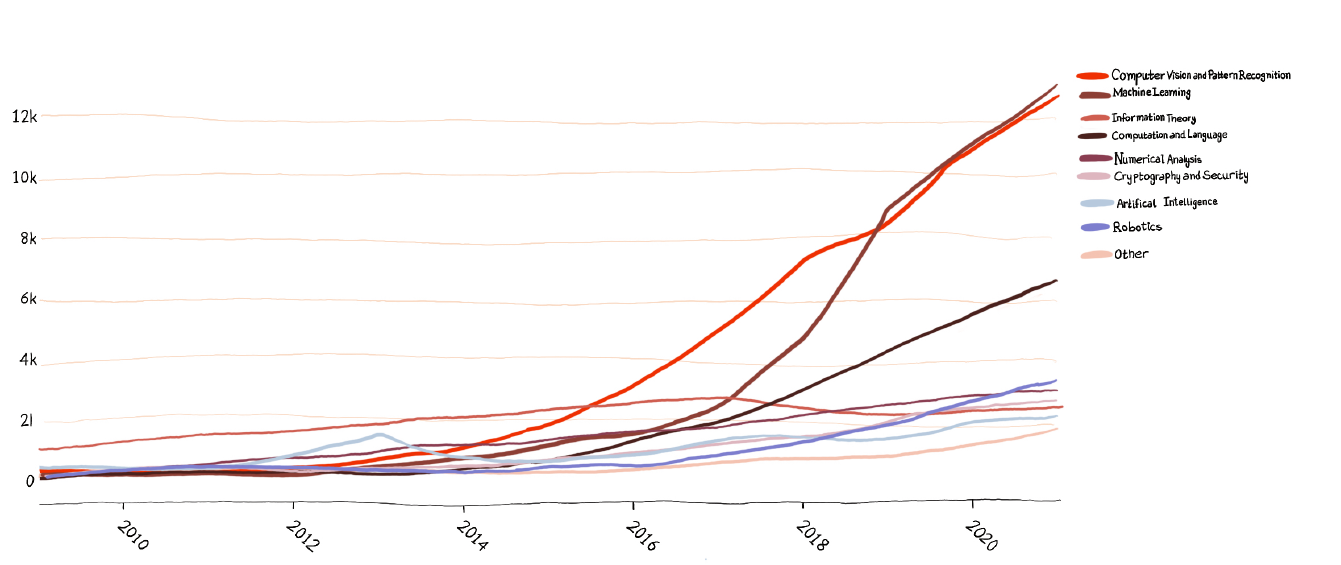}
    \caption{ArXiv submissions history in Computer Science.}
    \label{fig:arxiv-sub-cs}
\end{figure}

Furthermore, within the scope of Computer Science, the most frequently submitted sub-categories for papers are "Computer Vision and Pattern Recognition", "Machine Learning", and "Computation and Language", i.e., areas where the learning paradigm\footnote{With learning paradigm, we are referring to a fundamental approach to AI development, which involves designing algorithms or systems that can improve their performance or behavior based on data or experiences. "De facto mode of operation" implies that within the mentioned fields, the predominant way of tackling problems or advancing research is through the learning paradigm.} has established itself as the \textit{de facto} mode of operation (Fig. \ref{fig:arxiv-sub-cs}).

This exponential increase in research output is highly correlated to the amount of money injected into the field \cite{zhang2021ai, Daniel_Zhang_et_al_2022, Maslej2023}. While these money-filled advancements have brought numerous benefits in terms of the expansion of AI capabilities, they also introduce risks and side effects that promote several ethical concerns, like risks to privacy, the potential for increased surveillance, the environmental cost of the industry, and the amplification of prejudices that disproportionately harm vulnerable groups. Consequently, the expansion of the AI industry has given rise to a \textit{"boom"} in AI Ethics, i.e., a period marked by an unprecedented demand for regulation, normative guidance, and general activity in this field \cite{correa2023worldwide}.

One of the central questions surrounding this boom is what ethical premises should guide the development of AI technologies. However, is there agreement on those premises and how they are defined? Given the ever-growing lists of normative declarations published, it is easy to get lost in this normative discourse, especially if you just stumbled into AI ethics. For example, imagine you are a policy-maker tasked to write a report that will further lead a legislative effort to create regulations for AI technologies. Where do you start? What can you use to base your assertions? Meanwhile, even if you are an expert, organizing this varied normative discourse into a searchable database can supercharge your work. Fortunately, this is precisely what WAIE is: a tool that allows people to explore the field of AI ethics and its discourse.\footnote{Corrêa, Nicholas Kluge, et al. "Worldwide AI ethics: A review of 200 guidelines and recommendations for AI governance." Patterns 4.10 (2023). \href{https://doi.org/10.1016/j.patter.2023.100857}{doi.org/10.1016/j.patter.2023.100857}.}

In short, WAIE draws inspiration from the previous works of past meta-analysts and meticulously surveys a wide array of available ethical guidelines related to AI development, presenting an extensive analysis of 200 documents as its source, including governance policies of private companies, declarations from academic institutions, governmental and non-governmental recommendations, and other guidelines published by many different stakeholders. Ultimately, WAIE aimed to explore whether a consensus exists regarding the normative discourse presented in ethical guidelines surrounding AI development. In other words, what are the resonances and conflicts when we say "\textit{AI should ...}"

Remember when we talked about external materials tied to this book? WAIE is one of them. Hence, we recommend the reader check the \href{https://nkluge-correa.github.io/worldwide_AI-ethics/dashboard.html}{WAIE dashboard}\footnote{\hspace{1mm}\includegraphics[scale=0.025]{img/link.png}\hspace{1mm} \href{https://nkluge-correa.github.io/worldwide_AI-ethics/dashboard.html}{nkluge-correa.github.io/worldwide\_AI-ethics/dashboard.html}} for a dive into AI ethics discourse. You can also find a very satisfying visualization of the WAIE dataset in its \href{https://nkluge-correa.github.io/worldwide_AI-ethics/embeddings.html}{embedding analysis}.\footnote{\hspace{1mm}\includegraphics[scale=0.025]{img/link.png}\hspace{1mm} \href{https://nkluge-correa.github.io/worldwide_AI-ethics/embeddings.html}{nkluge-correa.github.io/worldwide\_AI-ethics/embeddings.html}} And for the avid data scientists, check \href{https://github.com/Nkluge-correa/worldwide_AI-ethics}{WAIE's code repository}\footnote{\hspace{1mm}\includegraphics[scale=0.025]{img/link.png}\hspace{1mm} \href{https://github.com/Nkluge-correa/worldwide_AI-ethics}{github.com/Nkluge-correa/worldwide\_AI-ethics}} to see how the analysis came to be.

WAIE has a foundational character in this work since much of what we will discuss for the rest of this book will make much more sense if we all have a common ground to stand on. At the same time, the principles and normativity extracted from WAIE will, again and again, be called in later chapters on the basis that values $x$, $y$, and $z$ appear to be something most of us can agree to be of value.

Stepping back a little, in the next section, we will dive into the work of past explorers and see how WAIE differs from past studies.

\subsection{Jobin, Hagendorff, and Fjeld}

Ana Jobin, Thilo Hagendorff, and Jessica Fjeld led the past explorations we would like to present in this brief session.

The first study to promote a systemic meta-analysis of published AI ethical guidelines was that of Jobin et al. \cite{Anna_Jobin_et_al_2019}. In it, these authors sought to investigate whether a global agreement on emerging questions related to AI ethics and governance would arise. The research identified 84 documents containing ethical guidelines for intelligent autonomous systems using the \textit{Preferred Reporting Items for Systematic Reviews and Meta-Analyses} framework \cite{liberati2009prisma}. At the time, some of them were the most mentioned guidelines in the literature, like the Organization for Economic Co-operation and Development Recommendation of the Council on Artificial Intelligence \cite{Karen_Yeung2020}, the High-Level Expert Group on AI Ethics Guidelines for Trustworthy AI \cite{AI_HLEG__HighLevel_Expert_Group_on_AI_2019}, the University of Montreal Declaration for responsible development of artificial intelligence \cite{University_of_Montral_2018}, the Villani Mission's French National Strategy for AI \cite{The_Villani_Mission_2018}, among many others. One of the main findings in Jobin et al. work was uncovering the most common ethical principles in the discourse of the evaluated documents, like Transparency, Justice/Equity, Non-maleficence, Accountability, Privacy, Beneficence, Freedom \& Autonomy, Trust, Dignity, Sustainability, and Solidarity.

Meanwhile, Thilo Hagendorff \cite{Thilo_Hagendorff2019} conducted another study that presented a similar type of analysis. His research focused on a smaller sample of 21 documents, using ad-hoc exclusion criteria.\footnote{Hagendorff excluded publications older than five years, those that only addressed a national context, and all documents characterized as corporate policies. He also deliberately selected papers he deemed relevant in the international discourse (IEEE, Google, Microsoft, and IBM).} Even working with a smaller sample, his findings corroborate with those of Jobin et al. \cite{Anna_Jobin_et_al_2019}, where the most mentioned principles found were Accountability (77\%), Privacy (77\%), Justice (77\%), and Transparency (68\%). Hagendorff (like Jobin et al.) also mentions the underrepresentation of institutions in South America, Africa, and the Middle East as a clear bias in his sample.

Last but not least important, we have the work of  Fjeld et al. \cite{Jessica_Fjeld_et_al_2020}. In their study, these authors worked with 36 samples produced by several types of institutions, but as Hagendorff \cite{Thilo_Hagendorff2019}, they also excluded data science, robotics, and other AI-related fields/applications. According to them, eight principles were the most commonly cited in their sample: fairness/non-discrimination (present in 100\% of the analyzed documents), privacy (97\%), accountability (97\%), transparency/explainability (94\%), safety/security (81\%), professional responsibility (78\%), human control of technology (69\%), and promotion of human values (69\%).\footnote{For a complete review of meta-analytical research on normative AI documents, we recommend the works of Schiff et al. \cite{schiff2020s, schiff2021ai}, which cite many other important works.}

While the work of the mentioned authors and their respective collaborators have certainly helped lay the foundation for discussions and future works regarding a descriptive analysis of the field, there are a couple of points we would like to raise (mainly those that sparked part of the motivation behind WAIE):

\begin{enumerate}
    \item Regarding Hagendorf and Jobin et al. works, it is worth questioning the exclusion criteria used by them (mainly driven by the popularity of each document). If we want to investigate the consensus regarding the normative dispositions of different countries and organizations regarding AI, we should not use popularity-based filtering. In other words, a descriptive ethics evaluation should consider as many viewpoints as possible.
    \item At the same time, while Fjeld et al. sought to be more diverse in the formation of their sample, just like the work of Hagendorf, both suffer from a small pool of documents while excluding areas of research that should not be separated from the multidisciplinary umbrella we call AI (e.g., robotics, data science, etc.).
    \item As pointed out by Jobin et al. and Fjeld et al., a more minute investigation of how ethical principles vary regarding their definition should be given more attention.
    \item Additionally, it is worth noting that these studies should have released their datasets in a form that would allow the replication of their findings or the expansion of their projects.
\end{enumerate}

This last item is perhaps the most critical. Reproducibility is a cornerstone of scientific integrity. With it, the very foundation of scientific progress becomes solid. Confidence wanes when findings cannot be replicated, potentially leading to (possibly) erroneous conclusions being accepted as fact. We are not saying that the abovementioned results are incorrect. However, for any person to arrive at the same conclusion as these authors or to build upon, the lack of open and accessible sources stands as, in our opinion, an unnecessary hurdle in this day and age. Particularly when research findings intersect with our collective normativity, be that in social policy, ethics, or public health, the imperative for transparency becomes even more pronounced. Something that we tried to promote in the creation of WAIE.
 
\subsection{Methodological Considerations}

Building upon the gaps pointed out in the previous section, WAIE presents to the AI community:

\begin{enumerate}
    \item A large and diverse sample size. WAIE possesses 200 documents from 37 countries spread over six continents.WAIE used as primary sources of documents two public repositories, the "\href{https://inventory.algorithmwatch.org/}{AI Ethics Guidelines Global Inventory}",\footnote{\hspace{1mm}\includegraphics[scale=0.025]{img/link.png}\hspace{1mm} \href{https://inventory.algorithmwatch.org}{inventory.algorithmwatch.org}} from AlgorithmWatch, and the "\href{https://www.linking-ai-principles.org}{Linking Artificial Intelligence Principles}"\footnote{\hspace{1mm}\includegraphics[scale=0.025]{img/link.png}\hspace{1mm} \href{https://www.linking-ai-principles.org}{www.linking-ai-principles.org}} (LAIP) Guidelines.
    \item Combined with a more granular typology of document types.
    \item Presented in an insightful and user-friendly data visualization framework.
    \item Released with an open-source dataset, making WAIE reproducible and extendable.
\end{enumerate}

WAIE focuses on guidelines related to the ethical use of AI technologies. That is, documents conceptualized as recommendations, policy frameworks, legal landmarks, codes of conduct, practical guides, tools, or ethical principles for AI systems and applications. Also, unlike previous works \cite{Thilo_Hagendorff2019, Jessica_Fjeld_et_al_2020}, WAIE encompasses several areas that inhabit the multidisciplinary umbrella of Artificial Intelligence research, e.g., Statistical Learning, Data Science, Machine Learning, Optimization Theory, Robotics, Software development and engineering, etc. This sample is a snapshot of the collective care and preoccupations of several stakeholders engaged with technology and AI's future, regardless of their popularity in the common debate.

\subsubsection{\textit{WAIE Features}}

WAIE explored several features, both quantitative and qualitative. Like in previous works \cite{Anna_Jobin_et_al_2019, Thilo_Hagendorff2019, Jessica_Fjeld_et_al_2020}, it presents the following core indicators:

\begin{itemize}
    \item Institution responsible for producing the document.
    \item Country/World Region of the institution.
    \item Type of institution (e.g., academic, non-profit, government, etc.).
    \item Year of publication.
    \item Ethical principles (as done by Fjeld et al. \cite{Jessica_Fjeld_et_al_2020}, WAIE breaks principles into themes of resonating discourse).
    \item Principles description (i.e., the words used in a document to define or support a given principle).
    \item Gender distribution among authors (inferred through a first-name automated analysis).
    \item Size of the document (i.e., word count).
\end{itemize}

WAIE also presents a set of qualitative categories created to further differentiate documents according to their normative content. These categories relate to:

\begin{enumerate}
    \item The content of the document.
    \item The type of regulation that the document proposes.
    \item The normative strength of the proposed norms.
    \item The impact scope that motivates the document's agenda.
\end{enumerate}

The first category relates to how its authors approached questions regarding the understanding, guidance, and implementation of AI technologies. WAIE defines them as mutually inclusive (documents may have all these features combined):

\begin{itemize}
    \item Descriptive: Descriptive documents take the effort of presenting definitions related to AI technologies. These definitions contextualize "what we mean" when we talk about AI.
    \item Normative: Normative documents present norms, ethical principles, recommendations, and imperative affirmations about what such technologies should be used or developed for.
    \item Practical: Practical documents present development tools to implement ethical principles and norms.
\end{itemize}

The second category delineates different approaches to governance and regulation within AI technologies. WAIE defines them as mutually exclusive (the presence of one feature excludes the other):

\begin{itemize}
    \item Government-Regulation: This category encompasses documents made by governmental institutions. These documents propose that states regulate the use and development of AI strictly (legally binding horizontal regulations) or softly (legally non-binding guidelines).
    \item Self-Regulation/Voluntary Self-Commitment: This category encompasses documents made by private organizations and other bodies. These documents defend a form of self-regulation governed by the AI industry, including any voluntary self-commitment made by independent organizations.
    \item Recommendation: This category encompasses documents that only suggest possible forms of governance and ethical principles that should guide organizations seeking to use, develop, or regulate AI technologies.
\end{itemize}

The third category outlines distinct methods for governance and regulation within the AI domain, with differing levels of enforceability and flexibility. These are also defined as mutually inclusive:

\begin{itemize}
    \item Legally non-binding guidelines: These documents propose an approach that intertwines AI principles with recommended practices for companies and other entities.
    \item Legally binding regulations: These documents propose an approach that focuses on regulating specific uses of AI through legally binding rules, such as mandatory requirements and prohibitions.
\end{itemize}

The final category delineates perspectives on the temporal scope of concern regarding AI technologies and their impacts (defined as mutually exclusive):

\begin{itemize}
    \item Short-Termism: This category encompasses documents in which the scope of impact and preoccupation focus mainly on current or short-term problems, like algorithmic discrimination, algorithmic opacity, privacy, legal accountability, etc.
    \item Long-Termism: This category encompasses documents in which the scope of impact and preoccupation focus mainly on future or long-term problems. Since such technologies are not yet a reality, we can classify these risks as hypothetical or, at best, uncertain.
    \item Short-Termism \& Long-Termism: This category encompasses documents in which the scope of impact is short and long-term, i.e., they present a "\textit{mid-term}" scope of preoccupation. These documents address issues related to the short-termism category while also pointing out the mid/long-term impacts of our current AI adoption (e.g., AI interfering in democratic processes, autonomous weapons, existential risks, environmental sustainability, labor displacement, and the need for updating our educational systems).
\end{itemize}

WAIE also presents 17 ethical principles defined by textual and semantic analysis of its sample. These defined principles aggregate similar and resonating values while maintaining significant differences in their written forms:

\begin{itemize}

    \item Accountability/Liability: Accountability refers to the idea that AI technology developers and deployers should comply with regulatory bodies. These actors should also be accountable for their actions and the impacts caused by their technologies.
    
    \item Beneficence/Non-Maleficence: Beneficence and non-maleficence come from bioethics and medical ethics. In AI ethics, these principles state that human welfare (and harm aversion) should be the goal of AI-empowered technologies.
    
    \item Children \& Adolescents Rights: The idea that we must protect the rights of children and adolescents is particularly addressed by some guidelines. AI stakeholders should safeguard, respect, and be aware of the frailties associated with young people.
    
    \item Dignity/Human Rights: This principle is based on the idea that everyone deserves proper treatment and respect. In AI ethics, respect for human dignity and human rights (i.e., the Universal Declaration of Human Rights) are used (sometimes) interchangeably.
    
    \item Diversity/Inclusion/Pluralism/Accessibility: This set of principles advocates the idea that the development and use of AI technologies should be done in an inclusive and accessible way, respecting the different ways that the human entity may come to express itself.
    
    \item Freedom/Autonomy/Democratic Values/Technological Sovereignty: This set of principles advocates the idea that the autonomy of human decision-making must be preserved during human-AI interactions, whether that choice is individual or the freedom to choose together, such as the inviolability of democratic rights and values, which are also linked to the technological self-sufficiency of nations and states.
    
    \item Human Formation/Education: Such principles defend that human formation and education must be prioritized in our technological advances. AI technologies require considerable expertise to be produced and operated, and such knowledge should be accessible to all.
    
    \item Human-Centeredness/Alignment: Such principles advocate that AI systems should be centered on and aligned with human values and our necessities.
    
    \item Intellectual Property: This principle seeks to ground the property rights over AI products and their generated outputs.
    
    \item Justice/Equity/Fairness/Non-discrimination: This set of principles upholds the ideas of non-discrimination and bias mitigation. It argues that algorithmic treatment should happen fairly regardless of the different sensitive attributes that may characterize an individual.
    
    \item Labor Rights: Labor rights are legal and human rights related to the labor relations between workers and employers. In AI ethics, this principle emphasizes that workers' rights should be preserved, regardless of whether AI technologies mediate or augment labor relations.
    
    \item Cooperation/Fair Competition/Open Source: This set of principles advocates different means by which joint actions can be established and cultivated between AI stakeholders to achieve common goals. It also relates to the free and open exchange of valuable AI assets.
    
    \item Privacy: The idea of privacy can be defined as the individual's right to expose oneself voluntarily, and to the extent desired, to the world. This principle is also related to data-protection-related concepts such as data minimization, anonymity, informed consent, etc.
    
    \item Reliability/Safety/Security/Trustworthiness: This set of principles upholds the idea that AI technologies should be reliable in that their use can be verified as safe and robust, promoting user trust and better acceptance of AI technologies.
    
    \item Sustainability: This principle can be interpreted as a manifestation of intergenerational justice, wherein the welfare of future generations must be considered in AI development. In AI ethics, sustainability pertains to the notion that AI advances should be approached with an understanding of their enduring consequences, encompassing environmental impact and the preservation and well-being of non-human life.
    
    \item Transparency/Explainability/Auditability: This set of principles supports the idea that the use and development of AI technologies should be transparent for all interested stakeholders. Transparency can be related to "the transparency of an organization" or "the transparency of an algorithm." This set of principles is also related to the idea that such information should be understandable to nonexperts and, when necessary, subject to auditing.
    
    \item Truthfulness: This principle upholds the idea that AI technologies must provide truthful information. It is also related to the idea that people should not be deceived when interacting with AI systems.
    
\end{itemize}

All this information is condensed into several visualization panels we call the \href{https://nkluge-correa.github.io/worldwide_AI-ethics/dashboard.html}{WAIE dashboard},\footnote{\hspace{1mm}\includegraphics[scale=0.025]{img/link.png}\hspace{1mm} \href{https://nkluge-correa.github.io/worldwide_AI-ethics/dashboard.html}{nkluge-correa.github.io/worldwide\_AI-ethics/dashboard.html}} an interactive, flexible, dynamic, and open (the WAIE dataset is freely available for download) database of values and norms for AI technologies, which can be used to enable many types of studies and projects. For example, with over 1400 definitions encapsulating the previously outlined 17 principles, WAIE provides a robust foundation of raw natural language for investigation. Using the capabilities of textual embeddings \cite{bengio2000neural} and basic machine learning techniques, one can transform these definitions into vector representations and project language into a three-dimensional space of human values expressed as natural language (which you can access and explore on \href{https://nkluge-correa.github.io/worldwide_AI-ethics/embeddings.html}{this panel}).\footnote{\hspace{1mm}\includegraphics[scale=0.025]{img/link.png}\hspace{1mm} \href{https://nkluge-correa.github.io/worldwide_AI-ethics/embeddings.html}{nkluge-correa.github.io/worldwide\_AI-ethics/embeddings.html}} By taking the time to filter and investigate, you may find several relations, correlations, similarities, and other insights that might prove helpful if you are interested in the field.

\subsection{WAIE: Analysis and Results}

Let us quickly examine some insights we can gather from the WAIE review. Check the published article \cite{correa2023worldwide} for a more detailed analysis, and open a browser window with the WAIE dashboard to check the graphs related to the results we are about to unveil.

\subsubsection{\textit{Worldwide Landscape}}

Looking at WAIE's distribution among world regions (aggregated by continent), the bulk of produced documents comes from Europe (especially countries from Western Europe, 31.5\%, like the United Kingdom, 12\%, and Germany, 10\%), North America (the United States of America, 29\%, and Canada, 5.5\%), that together represent a third of the sample size, and Asia (mainly represented by East Asian countries, 11.5\%, like China, 5.5\%, and Japan, 4\%). In contrast, South America, Africa, and Oceania represent less than 4.5\% of the sample, with countries like Brazil (1.5\%) spearheading this sub-distribution (Latin America, 3.5\%). Other world regions and countries would be even more underrepresented without the significant participation of intergovernmental organizations like NATO, UN, and UNESCO, which represent 6\% of the sample size (13 documents).

In summary, Europe and North America dominate the scene, while Asia secures a significant share. However, South America, Africa, and Oceania are notably underrepresented. Special recognition goes to intergovernmental organizations for their participation, which helps diversify the sample.

\subsubsection{\textit{Institutional Distribution}}

Switching the gaze to institution types, except for institutions like IBM (5), Microsoft (4), and UNESCO (3), most other institutions do not have more than two published samples in the WAIE's dataset. Meanwhile, the bulk of the sample was produced by governmental institutions and private corporations (48\%), followed by CSO/NGO (17\%), non-profit organizations (16\%), and academic institutions (12.5\%).

So, basically, big players like IBM and Microsoft produce the most in this regard. Meanwhile, governmental institutions and private corporations hog the limelight, while CSOs/NGOs and academic institutions try to make their mark noticeable. 

\subsubsection{\textit{Gender Distribution}}

The gender distribution among authors in the WAIE dataset possesses a clear imbalance, where 66\% have no authorship information. On the remaining part of the distribution, authors with "male" names are prevalent (66\% male, 34\% female). While Academic institutions (62\% male, 38\% female) and non-profit organizations (65\% male, 34\% female) are the less disparate institutions, they still fall short of the 1:1 parity ratio. Meanwhile, industrial associations show the highest level of disparity, with only 13\% of authors identified as female.

In summary, the gender balance in the WAIE dataset is problematic. Even in supposedly more enlightened spaces like academia and non-profits, parity is still a distant dream.

\subsubsection{\textit{Unveiling Document Typologies}}

In regards to the WAIE's typologies, looking at the document's content, the majority of its sample is from the normative type (96\%), which half of the time also presents descriptive contents (55.5\%), and more rarely, practical implementations (27\%). The form of regulation proposed by the documents of WAIE's sample is majorly comprised of recommendations to different AI stakeholders (56\%), while 24\% possess self-regulatory/voluntary self-commitment style guidelines, and only 20\% propose a form of regulation administered by a given country.

This lack of convergence to a more "government-based" form of regulation is reflected in the normative strength of these documents, where the vast majority (98\%) only serve as guidelines that do not entail any form of legal obligation, while only 4.5\% propose strict regulations. Since only governmental institutions (24\% of WAIE's sample) can create legally binding norms, one could argue that this imbalance lies in this fact. However, by filtering only the documents produced by governmental institutions, the disproportion remains, with only 18.7\% remaining documents proposing legally binding forms of regulation.\footnote{The countries on the front of this still weak trend are Canada, Germany, and the United Kingdom, with Australia, Norway, and the USA coming right behind.}

In regards to impact scope, the totality of WAIE's sample shows that short-term (47\%) and "\textit{mid-term}" (52\%) prevail over more long-term preoccupations (2\%). Filtering by impact scope and institution type demonstrates that private corporations think more about the short-term (33\%), governmental institutions about the mid-term (28\%), academic (66\%), and non-profit organizations (33\%) with the long-term impacts of AI technologies.

In summary, WAIE unveils a severe practical gap. Regarding regulation, it's mostly just recommendations rather than actual regulation, while existing rules are still vague and ungrounded. As for impact, it's all about the here and now, with the future getting little more than a passing thought.

\subsubsection{\textit{Worldwide Values}}

Examining the distribution of principles among WAIE's sample, the following result becomes evident: the top five principles advocated are similar to the results shown by Jobin et al. \cite{Anna_Jobin_et_al_2019}, and Hagendorff \cite{Thilo_Hagendorff2019}, with the addition of Reliability/Safety/Security/Trustworthiness (78\%), which also was top five in Fjeld et al. \cite{Jessica_Fjeld_et_al_2020} meta-analysis (80\%).

One of the advantages of WAIE's visualization panel is the possibility to investigate how different features affect one another. For example, when examining principle distribution filtered by institution type, one can notice that the main advocated principle of governmental institutions (worldwide) is the need for transparent systems (89.5\%). Also, while private corporations mainly defend the need for Reliability (87.5\%), CSOs/NGOs primarily support the necessity for more algorithmic fairness (88.2\%). 

Another advantage is that WAIE displays all definitions given by each document to the mentioned principles, allowing for a more diverse comparison of how these abstract objects are defined. For example, when examining the principle of Transparency, the definition proposed in \textit{"ARCC: An Ethical Framework for Artificial Intelligence"} \cite{Tencent_Research_Institute_Tencent_2018} states that:

\begin{quote}
    \textit{"Promote algorithmic transparency and algorithmic audit, to achieve understandable and explainable AI systems. Explain the decisions assisted/made by AI systems when appropriate. Ensure individuals' right to know, and provide users with sufficient information concerning the AI system's purpose, function, limitation, and impact." }
\end{quote}

While the one provided by \textit{"A practical guide to Responsible Artificial Intelligence (AI)"} \cite{PricewaterhouseCoopers_PwC_2019} says (about the same principle):

\begin{quote}
    \textit{"To instill trust in AI systems, people must be enabled to look under the hood at their underlying models, explore the data used to train them, expose the reasoning behind each decision, and provide coherent explanations to all stakeholders promptly. These explanations should be tailored to the different stakeholders, including regulators, data scientists, business sponsors, and end consumers."}
\end{quote}

The WAIE's dataset and panels contain all available definitions (for every principle). We encourage readers to explore the topics closest to their hearts and minds and search for commonalities and dissimilarities among principles definitions.

\section{Critical Analysis}
\label{critical_analysis}

WAIE is a descriptive analysis and account of the normative realm surrounding AI. It is an example of normativity being observed and questioned from a third-view perspective. Later in this book, these ideas will become a foundation for \textcolor{BrickRed}{Dynamic Normativity}, which rests on the assumption that values and preferences reside not only in ourselves but also in the environment we interact with.

Now that we have all of this knowledge to our avail, as moral agents endowed with reflexive and critical thinking, before moving on, we will exercise these faculties to promote a critique of everything WAIE showed us. And what has WAIE demonstrated?

\begin{enumerate}
    \item The debate on the ethics of AI, at least in the form investigated by WAIE, continues to be influenced by the North American and European narrative, i.e., the global north.
    \item This debate has been led by government and private institutions.
    \item The production of these norms appears to have a gender distribution imbalance.
    \item The "AI Ethics" topic emerged to the zeitgeist in 2018.
    \item The types of documents that have been produced have a strongly self-regulatory character, untied to legal sanctions.
    \item While there is convergence on the most urgent values to be addressed, there is considerable variation in their definitions.
    \item Most of its proposed is untied to any form of praxis.
\end{enumerate}

WAIE can provide numerous additional insights to the avid reader. However, in this section, we would like to present our subjective evaluation of the gathered results, offering a critical analysis inspired by other critiques that, via observation of the codependent relationship we have with technology, came to question the underlying assumptions and power structures of society, in the hope of to exposing flaws and power imbalances within its systems \cite{Samuel_Butler_1863, Fredric_Jameson_1991, Mike_Davis_1992, kaczynski1995industrial, Jean_Baudrillard_1994, Joan_Alway_Craig_Calhoun1997, Paul_Virilio_1997, Jean_Baudrillard_1998, Hartmut_Rosa__and__William_E_Scheuerman_2009, Robert_Tally_2009, waelen2022ai}.

\subsection{Unknown Perspectives \& Marginalized Regions}

Even with a sample size twice as large as the one analyzed by Jobin et al. \cite{Anna_Jobin_et_al_2019}, WAIE cannot escape North American and European hegemony in the discourse. However, we can defy this result by bringing other indicators that put other countries as serious AI stakeholders.

For example, According to Savage \cite{Neil_Savage2020}, from 2016 to 2019: "\textit{China's output of AI-related research increased by 120\%, whereas output in the USA increased by "mere" 70\%. In 2019, China published 102,161 AI-related papers, and the USA published 74,386.}." Also, based on the AI Index Annual Report, the USA, China, and India are the top three countries by the Vibrancy Ranking. While this helps to explain why almost a third of WAIE's sample size comes from the USA, it does not account for the underrepresentation of countries like China and India. Again, according to Zhang et al., China has far surpassed the USA in journal/conference publications and citations, while most of the "AI talent concentration" is found in India. These new indicators question the hegemony of North American and European ethical narratives, supposedly supported by their excellence and leading position in AI development.

Also, we argue that the "Guidelines for AI Technologies" scope hides much of the normative discourse done elsewhere. For example, the African continent is significantly underrepresented in the WAIE dataset. However, according to Kiemde and Kora \cite{Sountongnoma_Martial_Anicet_Kiemde_Ahmed_Dooguy_Kora2021}, 17 of the 55 African Union member states possess data protection and privacy legislation. At the same time, Mauritius announced the establishment of a National AI Council, making it the first African state to present an AI strategy. Kiemde and Kora also demonstrate in their review a collection of published papers and documents about AI ethics in Africa and other underrepresented countries \cite{Arthur_Gwagwa_2019, David_Vernon2019,  Laura_Sallstrom__et_al_2019, Mounir_Bensalah_2021, Arthur_Gwagwa__et_al_2021}, which helps us to show that this type of discourse is present in the African States and probably in all other places that do not show up in WAIE and other popular reviews \cite{Anna_Jobin_et_al_2019, Thilo_Hagendorff2019, Jessica_Fjeld_et_al_2020}.

Hence, it is clear that countries such as China and India, along with various African nations, are making significant strides in AI, and the contours of their contribution extend beyond the scope currently captured by the analysis above and WAIE. While many voices remain unknown, agreement on the principles that should guide AI development and use remains uncertain. In essence, it is not that our environment does not contain the values of all of these under-represented groups, but that our ability to uncover those is still primitive. Something that might hinder society in matters regarding alignment.

\subsection{The Hegemony of State and Private Sectors \& the Regulation Dilemma}

WAIE's results mirror the findings of Jobin et al. \cite{Anna_Jobin_et_al_2019} and Fjeld et al. \cite{Jessica_Fjeld_et_al_2020}, where most of the sample comes from private institutions (24\%) and governmental organizations  (24\%).

This equal presence of both State and Private stakeholders in the current normative discourse may be related to the expanse and success of the tech industry \cite{zhang2021ai, Daniel_Zhang_et_al_2022, Maslej2023}. Nowadays, most AI breakthroughs come from the industry \cite{openai2023gpt, touvron2023llama, yu2022scaling}. An AI industry that, seeing the demands for regulation and accountability from civil society, quickly reacted by proposing the rules that should (allegedly) guide their progress. Many of such promises are, perhaps, genuine. However, when governments and private institutions have "the same weight" in the general normative discourse, attention to the matter seems needed, especially when many of these technologies remain in gray areas of regulation.

This fact may become more alarming when we look at the distribution of government documents that opt for "soft" forms of regulation (91.6\%). The critique that "\textit{ethical principles are not enough to govern the AI industry}" is not a new one \cite{Thilo_Hagendorff2019, Anna_Jobin_et_al_2019, Brent_Mittelstadt2019, Anas_Ressguier_Rowena_Rodrigues2020, Nicholas_Corra_Nythamar_Fernandes_de_Oliveira2021}, however, perhaps those critiques have not yet permeated the mainstream community, which produces guidelines primarily based on principles detached from observable metrics or practical implementations.

Even if most countries opt for legally non-binding forms of regulation, there is a growing adoption/proposition of stricter solutions, with countries like the USA, Canada, the United Kingdom, and international organizations like the European Union spearheading this trend. At the same time, currently, we live in a time where influential members of the AI industry "urge for regulation" \cite{kang2023urges}. Nevertheless, given the influence tech oligopolies have over regulating bodies \cite{mcintosh2018we, petit2020big, fukuyama2021save, bhadra2022linkedin}, it is currently unclear if the "soon to come" regulation will not help cement this already heavily-centralized industry as it happened in other situations (e.g., telecommunications in the USA) \cite{brock1994telecommunication, melody1999telecom, economides2004telecommunications}.\footnote{The Telecommunications Act of 1996 \cite{krattenmaker1996telecommunications} aimed to promote competition and innovation in the US telecommunications industry. However, specific provisions, such as the requirement for new entrants to negotiate interconnection agreements with incumbent providers, proved a significant hurdle for smaller companies. The costs and complexities of negotiating these agreements put the established telecom monopolies in an advantageous position. As a result, the regulations contributed to the formation of monopolies/oligopolies, making it almost impossible for new players to enter this field.}

If we subscribe to the idea that the unjust and unquestioned centralization of power is the precursor of tyranny and oppression \cite{may1992crypto, may1994cyphernomicon, graeber2009direct, graeber2013democracy}, regulation becomes a double edge sword. And this is one of the tensions we are currently facing on this front:

\begin{quote}
    \textit{"While regulation can help us escape the self-regulatory trap that endows the tech goliaths with the impunity to act in undesirable ways, bad regulation may help solidify the existing technological oligopolies."}
\end{quote}

If regulation is implemented carelessly, while big companies will have the resources to comply with them (even shaping them to be favorable to their context), small organizations will not. While big companies have the financial backing to deal with lawsuits and fines, the open-source community and Academia do not. And if regulation comes in terms of capabilities, this will be the same as putting the future of AI development in the hands of these oligopolies. Then, we will institute the idea that only the industry can perform state-of-the-art AI development in our society.

\subsection{The Importance of Defining "AI"}

In regards to the content of the samples that form the WAIE dataset, we see that only 55.5\% of documents seek to define what is the object of their discourse, i.e., "\textit{we are talking about autonomous intelligent systems, and this is what we understand as an autonomous intelligent system}." This is a curious phenomenon, more so if we acknowledge that there is no consensual definition of what "\textit{Artificial Intelligence}" is and what is not \cite{Dagmar_Monett_et_al_2020}. 

There are many interpretations and contesting definitions, which may be a challenge for regulating organizations. For example, suppose you define AI as only "\textit{systems that can learn}". In that case, you will leave an entire family of systems outside your scope of regulation that does not learn (rule-based systems) but can still act "intelligently" and autonomously.

An incomplete definition can leave vital areas out of the normative scope of a proposal. For example, many rule-based systems that do not use ML are the base for Lethal Autonomous Weapons (LAW), a topic that only 4.5\% of the WAIE's sample mentions. Given that, since 2010, the major global superpowers have heavily engaged in weaponizing AI technologies \cite{geist2016s, kania2017battlefield, haas2017evolution, field2019syria, maas2019viable, roff2019frame, hacaoglu_turkey, trevithick_turkey, hambling2020turkish, crofts2020negotiating, russell2023ai}, is uncertain to us why many of such documents do not consider these artifacts as part of their normative agenda.

Perhaps we could attribute this to the fact that for many countries,\footnote{Garcia \cite{garcia2019militarization} points out in his analysis of autonomous weapons research and development that at least seven countries (United States, China, Russia, United Kingdom, France, Israel, and South Korea) stand out for their substantial engagement in the development of autonomous weapons.} allied with Academia and private institutions, the development of autonomous weapons is an active research and development area, with the Arms Industry funding several AI research programs and private companies \cite{baum2017survey, fitzgerald20202020}.

Defining the boundaries of artificial intelligence presents a challenge akin to finding a needle in a haystack during an earthquake. And given that only a little over half of the documents try to specify their subject matter, it's understandable why confusion persists in the regulatory domain.

\subsection{Hidden Costs \& Side Effects}

When looking at some of WAIE's least mentioned principles, like Labor Rights, Sustainability, and Truthfulness, the problems stated by Hagendorff \cite{Thilo_Hagendorff2019} and Jobin et al. \cite{Anna_Jobin_et_al_2019} are underlined, where the lack of attention given to questions related to the costs and misuse of our current AI-technologies remains overlooked in much of these guidelines. Taking these three principles as examples, only 22\% cite Sustainability, 19.5\% cite Labor Rights, and 8.5\% mention Truthfulness.

The tech industry has a high ecological and social cost. For example, some reasons for our current AI summer are the progress in hardware performance, better training methodologies for neural networks, and the massive amount of available data to train such algorithms. One of the fields that can serve as a clear example of this reality is Natural Language Processing (NLP). Current advances in neural network architecture \cite{vaswani2017attention} have enabled the creation of deep neural networks that seem to improve their performance as long as we scale their size and training volume \cite{hestness2017deep, henighan2020scaling, kaplan2020scaling, bahri2021explaining}. These Large Language Models (LLMs) (e.g., BERT \cite{devlin2018bert}, Chinchilla \cite{hoffmann2022training}, LLaMA \cite{touvron2023llama}, GPT-4 \cite{openai2023gpt}), show unmatching generality and adaptability when compared to earlier deep learning systems. However, as Strubell et al. \cite{strubell2019energy} point:

\begin{quote}
    \textit{"[...] these accuracy improvements depend on the availability of exceptionally large computational resources that necessitate similarly substantial energy consumption. As a result, these models are costly to train and develop, both financially, due to the cost of hardware and electricity or cloud computing time, and environmentally, due to the carbon footprint required to fuel modern tensor processing hardware."}
\end{quote}

Modern AI systems have the potential to incur massive energy consumption during their training and fine-tuning phases. As model sizes grow, this energy requirement only increases. With some models now consisting of trillions of parameters \cite{fedus2021switch}, their carbon footprint can reach up to hundreds of kilograms of emitted $CO_2$ \cite{strubell2019energy, patterson2021carbon, luccioni2022estimating}.\footnote{It is also crucial to realize that developing a high-performing language model extends beyond a simple training loop. Achieving neural network robustness in any domain often necessitates multiple training rounds to experiment with different model architectures and hyperparameters, a process that still heavily relies on heuristics and trial and error.}

Looking beyond the environmental impact, it is worth noting that most individual researchers do not have the resources to undertake such research. Building large-scale models demands access to significant amounts of specialized hardware, while servers that offer this computing power are often exorbitantly priced. This exclusivity based on financial capital has changed the deep learning landscape since its early days. Nowadays, achieving state-of-the-art results in deep learning is often a financially exclusive endeavor, and the one with access to more computing usually wins.

Another point overlooked is the "$CO_2$ tunnel vision", where many other markers remain ignored. For example, the extractivist practices related to the mining of vital resources to the tech industry, such as copper, gold, silver, coltan, aluminum, titanium, lithium, and many others, are unfortunately related to the exploitation of (child) labor in developing countries \cite{montague2002stolen, mustapha2007occupational, nest2011coltan, sutherland2011coltan, usanov2013coltan}.\footnote{In 2011, it is estimated that artisanal mining had employed approximately 16\% of the Democratic Republic of the Congo population (13.5 million people), being one of the most profitable labor activities in the country \cite{nest2011coltan}.}

In general, human labor is an overlooked issue. One of the primary causes of unemployment over the past 200 years has been the automation of processes previously carried out by people \cite{peters2019technological}, something that also has helped to create the massive wealth gap in modern society \cite{rotman2013technology}. Meanwhile, prospects seem unsettling. Frey and Osborne \cite{frey2013future} estimated the probability of automation for 702 US occupations in their survey. The findings predicted that over the next 20 years (10 years from now), technology would automate 47\% of these professions. A similar study conducted by Gruetzemacher et al. \cite{gruetzemacher2020forecasting} showed that to most experts, we already can automate 22\% of all jobs, a number predicted to increase to 40\% in 5 years and 60\% in 10 years. According to Eloundou et al., \cite{eloundou2023gpts}, LLMs like GPT-4 could affect at least 10\% of all tasks of 80\% of the US workforce, with approximately 19\% of workers potentially experiencing at least 50\% of their tasks impacted.

The apparent expansion of informal employment is another worrying development linked to the exploitation of human labor driven by our AI progress. For instance, the rise of click working, i.e., a type of task required for building large labeled datasets \cite{irani2016hidden} or performing repetitive tasks we still cannot automate, has been linked to many cases of labor rights violations \cite{harris2014amazon, perrigo2023workers}.\footnote{By "labor rights", we refer to the International Labour Organization standards \cite{kaufmann2008ilo}.}

Under these conditions, countermeasures concerning labor displacement are still in their infancy. For example, the "Windfall's Clause" \cite{o2020windfall}, a hypothetical legal ex-ante agreement between large AI companies and the world, would guarantee that businesses involved in AI development are committed to sharing their profits with society in case they create AGI, i.e., an optimistic take on the merit of trickle-down economics. However, one could say that the kindness of the rich and powerful would be nothing more than a promise. A more critical response would be that the dominant classes never gave anything to the oppressed, except under the threat of conflict, unless they had something to gain. In the more eloquent words of Maria da Conceição Tavares \cite{InstitutoEconomiaUnicamp2017}:

\begin{quote}
    \textit{"I do not think that except under pressure and conflict, the ruling class has given anything for free to those below unless they wanted them for cannon fodder."}
\end{quote}

The last principle in our short list of overlooked maxims is truthfulness. Despite being an underrepresented principle in WAIE, truthfulness addresses one of the most significant concerns presented by the AI community in the last few years, where issues related to fake content are some of the most reported issues and vulnerabilities about modern generative AI systems.\footnote{\hspace{1mm}\includegraphics[scale=0.025]{img/link.png}\hspace{1mm} \href{https://huggingface.co/spaces/nicholasKluge/Model-Library}{huggingface.co/spaces/nicholasKluge/Model-Library}} \footnote{\hspace{1mm}\includegraphics[scale=0.025]{img/link.png}\hspace{1mm} \href{https://incidentdatabase.ai}{incidentdatabase.ai}} For example, while LLMs can produce ostensibly plausible text with no basis in objective reality, ranging from erroneous information \cite{lin2021truthfulqa} to code that immediately fails when executed \cite{brown2020language, chen2021evaluating}, image and video generation models can create photo-realistic images that support false claims (such as the fake evidence that "\href{https://www.tiktok.com/@carlomunar/video/7215257056058658053}{Pope Francis is covertly a notorious dancer and stylish bon vivant}").\footnote{\hspace{1mm}\includegraphics[scale=0.025]{img/link.png}\hspace{1mm} \href{https://www.tiktok.com/@carlomunar/video/7215257056058658053}{tiktok.com/@carlomunar/video/7215257056058658053}}

Problems like these are emerging phenomena related to the new advances in the field. Before the era of generative models, coherent text generation, quality code generation, or photo-realistic text-to-image conversion was years away from fooling people. Nowadays, one can change the faces of two people in an image or video using \href{https://github.com/deepfakes/faceswap}{open software}\footnote{\hspace{1mm}\includegraphics[scale=0.025]{img/link.png}\hspace{1mm} \href{https://github.com/deepfakes/faceswap}{github.com/deepfakes/faceswap}} with little to no training or with one click, create a picture of a face that \href{https://this-person-does-not-exist.com/en}{does not exist}.\footnote{\hspace{1mm}\includegraphics[scale=0.025]{img/link.png}\hspace{1mm} \href{https://this-person-does-not-exist.com/en}{this-person-does-not-exist.com/en}}

Hence, in an ever-evolving field, we reside in an era where the concept of "truth" is exceedingly vulnerable. Given the known harms misinformation can cause \cite{howard2016bots, chapman2016slender, luscombe2022google}, it is worth being mindful that only now the issue of automated misinformation spread has gained attention beyond the ML community \cite{sample2023science, editorials2023tools, cotton2023chatting}, indicating a possible gap between AI as an applied field, and AI Ethics as an applied field of philosophy.

\subsection{The Principle-Practice Gap}

As already stated by Fjeld et al. \cite{Jessica_Fjeld_et_al_2020}, there is a gap between established principles and their actual application. In the WAIE sample, most of the documents only prescribe normative claims without the means to achieve them, while the effectiveness of more practical methodologies, in most cases, remains extra empirical \cite{correa2024crossing}.

Regulation can be perceived as an implementation of normative ethics, which alone lacks the strength to enforce its normativity. Several studies point out that ethical standards alone have little to no impact on decision-making across a wide range of professional fields \cite{brief1996values, cleek1998can, lere2003impact, osborn2009ethical, calo2017artificial}, something even more pronounced in STEM-related fields that do not have a study tradition in Humanities \cite{friedman1996value, Brent_Mittelstadt2019, bynum2006flourishing, davis2015value, mcnamara2018does, vakkuri2020just, correa2022efficiency, griffin2024ethical}. 

This lack of praxis is known, and many authors have raised concerns about this state in which much of the field rests. Jobin et al. \cite{Anna_Jobin_et_al_2019}:

\begin{quote}
    \textit{"Private sector involvement in the field of AI ethics has been questioned for potentially using soft policies as a way to turn a social problem into something technical or to completely avoid regulation."}
\end{quote} 

Hagendorff \cite{Thilo_Hagendorff2019}: 

\begin{quote}
    \textit{"AI ethics - or ethics in general - have no mechanisms to reinforce its normative claims."}
\end{quote} 

Rességuier and Rodrigues \cite{Anas_Ressguier_Rowena_Rodrigues2020}: 

\begin{quote} 
    \textit{"Ethics have great powerful teeth. Unfortunately, we are barely using them in AI ethics - no wonder then that AI ethics is called toothless."}
\end{quote} 

Mittelstadt \cite{Brent_Mittelstadt2019}:

\begin{quote} 
    \textit{"Statements reliant on vague normative concepts hide points of political and ethical conflict. "Fairness", "dignity", and other such abstract concepts are examples of "essentially contested concepts." At best, this conceptual ambiguity allows for the context-sensitive specification of ethical requirements for AI. At worst, it masks fundamental, principled disagreement and drives AI Ethics towards moral relativism. At a minimum, any compromise reached thus far around core principles for AI Ethics does not reflect meaningful consensus on a common practical direction for "good" AI development and governance."}
\end{quote} 

Munn \cite{munn2022uselessness}: 

\begin{quote} 
    \textit{"[...] these are meaningless principles which are contested or incoherent, making them difficult to apply; they are isolated principles situated in an industry and education system which largely ignores ethics; and they are toothless principles which lack consequences and adhere to corporate agendas."}
\end{quote} 

We believe such harsh criticism motivates the field to fulfill its interdisciplinary promise. As long as we treat AI Ethics as a purely philosophical endeavor, crossing the principle-practice gap remains a complicated goal. However, when applied areas like law, engineering, and computer science enter the fray, we can solidify the "ought" into what "can" be done. Regardless, without the humane side of this equation, technicians may become lost in a field where the guiding compass is that which profits the industry.

Before we proceed to the final section of this introductory chapter, let us ask the question: What barriers prevent this interdisciplinary agenda from becoming the norm? If alignment, for example, is supposed to be considered both a technical and philosophical problem, a one-sided solution may not be enough to solve the entirety of the problem of embedding values into AI systems. In an attempt to answer this question, we will find a paradox that might help indicate the current deficits of AI ethics research inside one of the institutions that promote it the most: \textit{Academia}.

\section{Paradoxes and Future Directions}
\label{way_forward}

WAIE's review presents several intriguing trends that put the field of AI Ethics into question. These trends can help expose sore points that, if left untouched, hinder the fulfillment of the AI Ethics promise. Many of the points we brought to light in our last sections, like the inequality in the volume of voices participating in this debate, the power imbalance between the state and private interests, and the myriad of side effects tied to the massive adoption and irresponsible use, of AI technologies, are, for most of us, outside of our immediate control. Even if we are compromised with promoting responsible and conscious adoption of these systems, many things require massive coordination of people and institutions that, at large scale, is an enormous problem on its own.

While the question of defining intelligence will be approached in \hyperref[chap4]{Chapter 4}, in this final section, we intend to approach the principle-practice gap, its paradox, and obstacles that, we argue, prevent us from achieving the goals of AI ethics as an applied field. And given that Academia\footnote{The worldwide community concerned with the pursuit of research, education, and scholarship.} is responsible for some of these obstacles, we find it just that these should be a sphere of interference that we, as academics, should be morally bound to "do something about".

Let us define some claims and positions before exposing this paradox in which Academia is locked. First, let us make our claim:

\begin{center}
    \begin{quote}
        \textit{If the Ethics of Artificial Intelligence is considered a sub-field of Applied Ethics, its most critical blank space is the normative gap between its theory and practice.}
    \end{quote}
\end{center}

Between theory and practice, let us represent these two sides by different approaches to AI Ethics: (1) the process of bringing normativity to AI development by addressing the human side of this process (e.g., defining ethical principles to help guide those who create, use, and govern AI systems and applications); and (2) the process of bringing normativity into the machine by seeking to embed values, like fairness and privacy, into AI systems and applications. Firstly, our concern involves the development of an ethical theory or framework; secondly, we seek ways to apply its values.

Again, we argue that for AI Ethics to fulfill its goal, i.e., ensuring that the use and development of AI systems and applications align with what we deem morally correct, its greatest current obstacle, in the academic sense, is the principle-practice gap between both these approaches, i.e., the lack of connection between the theoretical and the practical parts of this field, which can canonically be represented by different areas of knowledge (disciplines) acting in this front, i.e., the humanities and STEM-related fields.

On the humanistic side, AI ethics largely remain based on a principal/opinion-based fashion. Given the diverse nature of normative discourse, many of these efforts remain in the realm of opinions and recommendations that are not unified, e.g., there is no agreement on what "fair AI" means. Unfortunately, this abstractness can be manipulated to push the agenda of those against closing the principle-practice gap, i.e., the practical and binding implementation of ethical standards. As an example of such manipulation, we can refer to the already-mentioned industry's push for self-regulation.

Meanwhile, on the technical side, the techno-solutionist's approach fails to grasp that normative problems are "human problems". Many of the issues that AI ethicists seek to tackle are not always "technical" but manifestations of historical oppression and economic inequality, among other social ills, that are repackaged and revisited in an automated world. At the same time, problems deemed "philosophical" or "political" are disregarded by STEM-related fields that consider themselves amoral or apolitical, which further blocks the presence of human sciences in the formation, research, and teaching related to the technical development of AI. This incomplete grasp of the problem results in incomplete solutions and implementations, i.e., a technical solution based on a flawed theoretical understanding of the problem.

Both the criticisms raised against these sides are the most common. In other words, while the engineering-based side criticizes the human disciplines-based side for being unable to grasp the nature of the technique under consideration ("How do neural networks work?"), the human disciplines-based side complains of the engineering-based side's inability to understand the intricacies related to normative theory ("What is the difference between a value ethics and a deontological approach to AI Ethics?"). Paradoxically, the cooperation that would help both sides improve their deficits is blocked by (1) those on either side who point out a deficit in the other camp and (2) long-standing structural barriers hindering interdisciplinary work.

Currently, AI ethics and AI research suffer from the excessive presence of gatekeepers and gatekeeping policies, i.e., individuals or institutional policies that admit or refuse access to a specific research context \cite{montgomery2020academic, welsh2021tolling}. Such practices impede AI ethics from evolving into a genuinely techno-humanistic endeavor. Think about it. How often are philosophy master students accepted into computer science PhDs and vice versa? How often are interdisciplinary projects nothing more than meetings and email exchanges between experts in specific disciplines working "together"? In other words, juxtaposition without integration. Or ask yourself, how rare are the bridge-crossers and makers that interconnect and cross the many fields? And how common are discourses surrounding ideas akin to "since you haven't studied $x$, you cannot have a say on $y$"? Or situations where academics are blocked from participating in certain events (of an interdisciplinary nature) for not having a background (or a degree) in $x$? Without learning linear algebra, you cannot work with machine learning ethics since you don't understand ML. You are not qualified to propose normative criteria for AI systems without mastering the underlying principles of several schools of ethics. However, often, such areas are closed to tutor or guide the learning of wanderers from other fields, preventing them from strengthening the same deficits that are so much attacked by members of the opposite field. Philosophy is a ball of fluff. Tecnosolutionism is a mock solution. In summary, these actors and policies push a tribalist agenda aimed at \textit{"protecting their area,"} which only contributes to the perpetuation of the principle-practice gap while simultaneously pushing for the importance of "interdisciplinary" \cite{graff2015undisciplining, graff2016problem}.

Fields of Applied Ethics demand interdisciplinary researchers, i.e., "bridge makers," that can transition from the theoretical to the practical. From the land of abstract principles to that of practical implementation. From what "ought to be" to what "can be done". Preventing this crossing hinders the realization of an applied field where the most critical obstacle is the gap between its humanistic-theoretical and practical-technical sides. Finally, we argue that until the mentioned factors prevent the creation of this bridge, AI ethics will remain short of its foremost promise.

Now, where do we go from here? As individuals, we can strive to act according to the techno-humanistic agenda AI Ethics requires. If we are in positions of leadership and authority, dismantling the barriers that prevent the integration of all aspects of AI ethics should also be a goal worth pursuing. We can help generate a new and more robust wave of ethicists and builders with these. The rest of this book attempts to approximate this ideal of philosophical work working as a foundation and support for the technical implementation of problems that regard us and the products of our craft and intellect. While the development of WAIE aimed at creating a tool and later analyzing and critiquing the results, the rest of this book will take the opposite approach. Motivations, formalizations, and foundations will serve as the roots for something to be later built as an experimental attempt to ground and apply machine ethics in real AI systems and then point out the limitations of the current methods available. With this, we hope to present a blueprint for how philosophical work can be carried, in an applied sense, when working with issues related to artificial intelligence.

\section{Epilogue}

AI research is a field that has been gaining much popularity in recent years, both in academia and in mainstream debate. AI Ethics, one of the many branches of this field, addresses the ethical issues and questions regarding using and developing such technologies. While the state-of-the-art in AI Ethics has converged on several core ethical principles, much of the discourse surrounding these narratives is still unclear and under-explored. While past reviewers sought to enlighten our understanding, gaps in their methods have inspired WAIE to expand and improve such work. By recognizing the rich insight a descriptive analysis can give to an academic investigation, Worldwide AI Ethics was developed as an introduction to the field and a tool for the community. A tool that serves as a record of the normativity humanity has been imprinting in its environment.

This introductory chapter exposed numerous challenges and paradoxes, highlighting the complexity of dealing with ethics and technological advancements. Currently, the principle-practice gap looms large, reminding us of the disconnect between theory and application, between the lofty ideals we aspire to, and the practical realities we face. The tension between humanistic and technical approaches underscores the need for interdisciplinary collaboration and a holistic understanding of our issues. Yet, barriers persist, hindering the realization of this interdisciplinary vision.

However, our journey continues, and we hope this initial exposition serves as a call to action. A call to strive for a more ethical and humane future in the age of artificial intelligence, where integrating knowledge, disciplines, and practices can lead us to a new kind of philosophy that can be applicable and foundational.

\chapter{AGI, Existential Risks, and the Control Problem}
\label{chap2}

\begin{flushright}
\textit{"I don't think technophobia or technophilia are appropriate responses to our situation. I think the only appropriate response is the most profound ambivalence. I think that is what we owe new technologies [...] We have to teach ourselves to be absolutely ambivalent about them, and mainly, we have to teach ourselves to imagine their inadvertent side effects because the inadvertent side effects are the side effects that tend to get us".}

\textcolor{BrickRed}{— William Gibson}
\end{flushright}

\section{Introduction}

As shown in the previous chapter, little academic attention has been paid to the long-term consequences of AI development despite all the culture and folklore we have surrounding this topic. In fact, much before the terms AI Alignment and AI safety became a thing, literature was already producing critical assessments of our technological development by extrapolating the future and its dystopian possibilities \cite{Philip_K_Dick_1968, John_Brunner_1968, William_Gibson_1984, Fredric_Jameson_1991, Mike_Davis_1992, asimov2004robot, Robert_Tally_2009}. Something that might have contributed to the terms "controllability", "alignment", or "human-level AI" becoming generically dismissed as not serious, or as Stuart Russell \cite{russell2014myths} would say: "\textit{myths and moonshine}". Even though debates surrounding AI alignment, for many years, were secluded to environments like blogs (e.g., \href{https://www.lesswrong.com}{LessWrong}\footnote{\hspace{1mm}\includegraphics[scale=0.025]{img/link.png}\hspace{1mm} \href{https://www.lesswrong.com}{www.lesswrong.com}} and \href{https://www.alignmentforum.org}{AI Alignment Forum}\footnote{\hspace{1mm}\includegraphics[scale=0.025]{img/link.png}\hspace{1mm} \href{https://www.alignmentforum.org}{www.alignmentforum.org}}) and EA-like circles, currently, we see this subject as a central theme of research for many scholars in Academia \cite{Dario_Amodei_et_al_2016, russell2019human, juric2020ai, critch2020ai, Dan_Hendrycks_et_al_2021, christian2021alignment, ji2023ai, bengio2023managing, hendrycks2023overview}. Even more so after the general realization that aligned models might be profitable products \cite{chatgpt}. Something that launched much of the industry into a frenetic race toward creating alignment techniques and systems.

Even though it is difficult to separate the probable from the improbable, we should not dismiss long-term warnings, mainly if they are based on the limitations of our current paradigm (i.e., the learning paradigm). To better illustrate these concerns, let us consult some of our current AI pioneers on their views.

François Chollet \cite{chollet2021deep}:

\begin{quote}
    \textit{"Choosing the right objective function for the right problem is extremely important: your network will take any shortcut it can, to minimize the loss; so if the objective doesn’t fully correlate with success for the task at hand, your network will end up doing things you may not have wanted. Imagine a stupid, omnipotent AI trained via SGD, with this poorly chosen objective function: “maximizing the average well-being of all humans alive.” To make its job easier, this AI might choose to kill all humans except a few and focus on the well-being of the remaining ones—because average well-being isn’t affected by how many humans are left. That might not be what you intended! Just remember that all neural networks you build will be just as ruthless in lowering their loss function—so choose the objective wisely, or you’ll have to face unintended side effects".}
\end{quote} 

Stuart J. Russell \cite{russell2014myths}:

\begin{quote}
    \textit{"The primary concern is not spooky emergent consciousness but simply the ability to make high-quality decisions. Here, quality refers to the expected outcome utility of actions taken, where the utility function is, presumably, specified by the human designer. Now we have a problem: (1) The utility function may not be perfectly aligned with the values of the human race, which are (at best) very difficult to pin down; (2) Any sufficiently capable intelligent system will prefer to ensure its own continued existence and to acquire physical and computational resources – not for their own sake, but to succeed in its assigned task [...] This is essentially the old story of the genie in the lamp, or the sorcerer’s apprentice, or King Midas: you get exactly what you ask for, not what you want".}
\end{quote}

Yann LeCun (when asked about the moral standing of the control problem) \cite{YannLeCun2019}:

\begin{quote}
    \textit{"Neither. There is no notion of evil in that context other than the fact that people died. It was an example of what people call value misalignment, right? You give an objective function to a machine, and the machine strives to achieve this objective. And if you don't put any constraints on this objective, like don't kill people and don't do things like this, the machine, given power, will do stupid things just to achieve this objective or damaging things to achieve this objective".}
\end{quote}

Most of these concerns revolve around the idea that specifying objectives is difficult and that if future AIs are built in the same fashion as our current AI models, we might face problems related to controllability. Meanwhile, suppose such advanced AI models surpass our intelligence without controllability methods in place. In that case, we might inherently give rise to so-called existential risks (X-risks) \cite{good1966speculations, vinge1993coming, Nick_Bostrom_2002, Ray_Kurzweil_2005, yudkowsky2008artificial, omohundro2008basic, Nick_Bostrom_2014, Stuart_Russell__et_al_2015, corabi2017superintelligent, barrett2017model, everitt2018agi, baum2018superintelligence, torres2018superintelligence, russell2019human, levin2020roadmap, correa2020singularity}.\footnote{X-risk (short for existential risk) refers to the potential danger of human extinction or the irreversible harm to humanity caused by artificial intelligence systems.}

In summary, in this chapter, we will present a brief historical construction of how fears related to X-risks came to be, seeking to ground its assumptions and arguments in an evidence-based approach. First, in Section \ref{agi_and_safety}, we will define AGI, what it could look like, and put forth some of the most known arguments related to safety regarding advanced AI systems. In Section \ref{agi_horizon}, we will present pieces of evidence, like our current state-of-the-art AI development and the collective opinion of experts in the field, to better ground the idea that developing AGI might not be impossible. Lastly, in Section \ref{controlability_power_box}, we will briefly introduce the control problem, where "Alignment" came to be proposed as a possible solution path. We will also present other ideas to solve this problem and explain why they might be ineffective control strategies for advanced AI systems. With this overview, we hope the reader can create a historically grounded depiction of how worries regarding advanced AI systems have originated the control problem and, later, alignment research.

\section{AGI and Safety}
\label{agi_and_safety}

For starters, we can use a well-known dichotomy to categorize various forms of intelligence when talking about AI, whether in the context of computer science \cite{searle1980minds, wang2019defining, russell2010artificial} or the study of the philosophy of mind or cognitive sciences \cite{haugeland1989artificial, newell1994unified, David_Chalmers_2010}: narrow intelligence\footnote{Narrow intelligence, also known as "weak" AI, is how we define artificial autonomous systems that are only proficient at specific tasks.} and general intelligence.\footnote{General intelligence, also referred to as "strong" AI, can be defined as a system capable of solving many kinds of problems (proficiently) in any domain, or at least in a wide range of domains.} Hence, the crux of the term AGI lies in the concept of generality, which is undoubtedly the most pivotal ingredient of this concept. In essence, an AGI system must possess the ability to function in a wide variety of contexts. Humans, for instance, are not innately skilled in numerous domains.\footnote{Melanie Mitchell's "Debates on the nature of artificial general intelligence" presents a good depiction of the debate surrounding the concept of AGI and helps to evidence that there are many different views on this matter \cite{mitchell2024debates}.}

Nonetheless, we can acquire knowledge and skills through our general cognitive abilities and transfer them to novel scenarios. Although some other animals may also exhibit some form of generality, our human capacity for generalization is unparalleled. While an ape may learn to use tools, the prospect of it journeying to the moon remains a distant reality. Regardless, it is reasonable to assume that if "intelligence" is not an intractable concept, humans do not represent the zenith of cognitive generality. Following this reasoning, within the expansive realm of all possible optimization processes, we will probably come across overarching systems that surpass us significantly. However, this proposition does not hold if an unknown limit to cognitive generality that closely aligns with the capabilities of humans exists. In other words, humans are already perfectly optimized for intelligence, and you cannot get much better than this. These metaphysical assumptions represent some of the roots of the different sides of the "intelligence" debate. Nonetheless, most of it revolves around the concept of generality.

But how general can you get? If we have, let us say, a 10\% increase in cognitive generality compared to chimpanzees, which share a surprising 99\% of their DNA with us, what would something that is 10\% more general than us look like? Something alien and incomprehensible. This line of reasoning (or something similar) probably originated the first mention of "super-intelligence". As far as we know, the first historical mention of such a concept, originally coined as "ultra-intelligent machine", is accredited to Irving J. Good \cite{good1966speculations}:

\begin{quote}
    \textit{Let an ultra-intelligent machine be defined as a machine that can far surpass all the intellectual activities of any man however clever. Since the design of machines is one of these intellectual activities, an ultra-intelligent machine could design even better machines; there would then unquestionably be an "intelligence explosion", and the intelligence of man would be left far behind.}
\end{quote}

Hence, if we accept that recursive self-improvement \cite{horvitz2014one} is a possibility,\footnote{We will contest this possibility later in this chapter.} creating AGI does not require AI to be AGI itself; it only requires AI to outperform humans in AI development. Once we reach this benchmark, the remainder is merely a recursive, autonomous sequence of AI models developing more refined AI models. Again, in the words of Good \cite{good1966speculations}:

\begin{quote}
    \textit{"[...] the first ultra-intelligent machine is the last invention that man need ever make, provided that the machine is docile enough to tell us how to keep it under control".}
\end{quote}

Now you can ask yourself, using the analogy of Tegmark \cite{MaxTegmark2023} as an example:

\begin{quote}
    \textit{"If we were Neanderthals, would we really consider it a smart move if we had really advanced biotech to build homo sapiens? You might say, "Hey, Max, let us build these homo sapiens". They are going to be smarter than us. Maybe they can help us defend ourselves against predators and help us fix our paths. Make them nicer. We will control them undoubtedly".}
\end{quote}

Hence, the crux of the AGI safety problem rests on two foundational arguments: (1) human control is ontologically better, and (2) controlling something more clever than you is a challenge. Nowadays, there is no question about who is in charge. However, the fact that modern AI systems are already quite proficient in coding \cite{chen2021evaluating, li2022competition, roziere2023code}, neural network architecture search \cite{zoph2016neural}, and able to self-improve to a limited extent \cite{huang2022large, selfinstruct}, rings the alarm bells tied to the recursive self-improving narrative \cite{barrett2017model, sotala2018disjunctive, Kaj_Sotala_2018}.

Nonetheless, even before machines had any proficiency in AI-development-related tasks, the X-risk literature was already expressing concerns about these possibilities \cite{Ray_Kurzweil_2005, naude2009technological, David_Chalmers_2010, lombardo2012consciousness, tegmark2016benefits, tegmark2017life, torres2018superintelligence, yamakawa2019peacekeeping}, which are well summarized, at least concerning the control problem, by Stephen Omohundro \cite{omohundro2008basic}. We can present the general narrative behind their arguments as follows:

\begin{quote}
    If we consider AI systems to be optimizing agents, i.e., something that implements an optimization algorithm/process to achieve some goal, like stochastic gradient descent for the minimization of a loss function or a $Q$-learning algorithm maximizing the expected return of a reward function, there is a case to be made that an "\textit{optimizer always choose the alternative that is optimal according to an objective function, if that action is available.}" That is all that $arg\;max\;f(x)$ does: "Give me the maximal $x$ according to $f$".
\end{quote}

Given the vastness of possible goals that can be represented as objective functions and the instrumental goals that accompany almost all of these goals, two theses are born: orthogonality and instrumental convergence. Instrumental convergence proposes that some goals are instrumentally valuable for many terminal objectives. Therefore, their pursuit becomes optimal for any system that seeks to optimize them. Meanwhile, orthogonality defends that normative judgments (prescriptions of what ought to be) cannot be derived through mere factual analysis.\footnote{Analogous to Hume's Is-Ought Gap \cite{hume2003treatise}: \textit{"In every system of morality, which I have hitherto met with, I have always remarked, that the author proceeds for some time in the ordinary way of reasoning, and establishes the being of a God, or makes observations concerning human affairs; when of a sudden I am surprised to find, that instead of the usual copulations of propositions, is, and is not, I meet with no proposition that is not connected with an ought, or an ought not. This change is imperceptible; but is, however, of the last consequence. For as this ought, or ought not, expresses some new relation or affirmation, 'tis necessary that it should be observed and explained; and at the same time that a reason should be given, for what seems altogether inconceivable, how this new relation can be a deduction from others, which are entirely different from it".}} In other words, just because the system is a good optimizer, we cannot assume it shares our preferences and values.

Instrumental convergence tells us that instrumental goals are desirable for achieving many objectives. For example, self-preservation is instrumental for many objectives unless you seek self-destruction. Similarly, resource acquisition is another instrumental goal for many objectives related to resource-bound agents. In this sense, instrumental goals are like a Swiss Army knife of desirable behaviors. On the other hand, orthogonality suggests that we cannot derive normative judgments through factual analysis alone. Thus, the capabilities of an optimizer tell us nothing about its objectives and what guides its behavior. If an optimizer has an unknown goal, there is nothing we can assume about its behavior based only on its optimizing capabilities. There are several possible optimizers, and two equal optimizers can produce vastly different behaviors.

For example, imagine two people trying to navigate a maze. One person's goal is to reach the end of the maze as quickly as possible, while the other person's goal is to explore every dead end and corner of the maze. Even though both people are navigating the same maze, they have different objectives, and from a third-person perspective, it may be difficult (if not impossible) to infer their goals without interrogating them first.

We can extrapolate these theses to conclusions like "\textit{Even though AI is aligned, it is still possible that such a system will have unknown instrumental goals that might generate unwanted behavior}" and "\textit{There is no guarantee that a truly powerful optimizer will share our values if we do not embed them into its objective function.}" Given that sharing our planet with a more cognitively advanced and misaligned entity puts us in a position where we might "lose control" (just like what is happening to every other species on the planet), the possible consequences may justify the kind of unrest verbalized by many individuals and organizations \cite{good1966speculations, yudkowsky2008artificial, omohundro2008basic, Nick_Bostrom_2014, Stuart_Russell__et_al_2015, corabi2017superintelligent, barrett2017model, everitt2018agi, baum2018superintelligence, torres2018superintelligence, russell2019human}.

In summary, these are the basic arguments and ideas that, through time, have morphed into what we call nowadays the problem of control. A problem that, according to Everitt et al. \cite{everitt2018agi}, presents a fertile and stimulating ground for scientific and philosophical research:

\begin{quote}
    \textit{"Why study AGI safety before it exists and before we even know if it will exist? There are at least two kinds of reasons for this. The first is pragmatic. If AGI is created, and we do not know how to control it, then the result can be catastrophic [...] It is customary to take precautions not only against the catastrophes we know about but also against catastrophes that have only a small chance of occurring (for example, a city may decide to build earthquake-safe buildings, even though the probability of an earthquake occurring is quite small) [...] AGI has more than a small chance of occurring and can cause significant catastrophes. The second reason is scientific. Possible AGIs are theoretically interesting objects, and the question of how humans can control machines smarter than they are is philosophically stimulating".}
\end{quote} 

As mentioned, these motivations have promoted an active response by a part of the AI community. However, while philosophical arguments can be insightful and often reveal new ways of thinking, they should not replace empirical verification. In interdisciplinary matters, as this work suggests, philosophical inquiry and empirical investigation should walk in tandem. Hence, in the next section, we will use sources outside the strictly philosophical realm to better assess the possibilities related to the creation of AGI and the risks involved.

\section{AGI on the horizon?}
\label{agi_horizon}

Forecasting technological development is hard. It is always unclear how pessimistic or optimistic we should be. Many renowned thinkers have put forth technical predictions that ultimately proved to be incorrect:

\begin{itemize}
    \item Lord Kelvin in 1896: \textit{"I have not the slightest molecule of faith in any kind of air navigation other than ballooning".} 
    \item Thomas Edison in 1889: \textit{"Alternating current is just a waste of time. Nobody will use it, ever".}
    \item Nuclear physicist Ernest Rutherford in 1933 said that anyone who proposed the possibility that we might one day extract the energy contained in atomic nuclei was \textit{"talking moonshine".}
    \item Albert Einstein, in 1932, shared a similar thought: \textit{"There is not the slightest indication that nuclear energy can be obtained. It would mean that the atom would have to be shattered at will".} 
    \item And lastly, Bill Gates in 1981: \textit{"No one will need more than 637KB of memory for a personal computer. 640KB should be enough for anyone".}
\end{itemize}

As we can see, pioneers also fail to predict what is on the other side of the curve. Could AGI skepticism be a victim of the same lack of vision? Or, are people who entertain the possibility of AGI generally misguided?

If we want answers, consulting experts is always a good idea.

\subsection{Expert Predictions}

Even among people who believe in the possibility of AGI, the time scales for this event have considerable variance. For example, when attempting to extrapolate technological trends, Raymond Kurzweil \cite{Ray_Kurzweil_2005} predicted that we would be able to simulate the human brain around 2029, basing his predictions on the rate of progress related to Moore's law.\footnote{Gordon Moore predicted that the number of transistors on a computer chip would double every two years.} According to Kurzweil, we would be very close to creating AGI with this level of computational power. On the other hand, David Chalmers \cite{David_Chalmers_2010} only asserts that we will develop AGI within this century.

Baum et al. \cite{baum2011long} used a more empirical approach. The authors obtained a median prediction for the creation of AGI for the year 2045 by polling 21 conference attendees at the Artificial General Intelligence 2009 (AGI-09) conference. Müller and Bostrom \cite{muller2016future} did research along the same lines. To assess the state of and anticipated directions in AI research, the authors polled 170 experts. According to the survey, on average, experts predict (with 50\% probability) that between 2040 and 2050, we would have developed high-level artificial intelligence. By 2075, the chances become 90\% for most experts.

Grace et al. \cite{grace2018will} present findings along the lines of Baun et al. \cite{baum2011long}. According to the researchers polled (352 attendees of the 2015 NeurIPS and ICML conferences), estimations suggest that AI will surpass human performance in all tasks in 45 years (with a 50\% chance), with the automation of all human work projected to be achievable in approximately 120 years. Additionally, when respondents were asked about the long-term impact of high-level AI, 10\% expressed a negative outlook, while 5\% indicated an extremely negative perspective. Based on the studies mentioned above, we could say that a minimum of 10\% of the AI community believes that we will develop artificial general intelligence within the next 120 years, while a significant portion agrees that AGI safety problems are relevant.\footnote{When asked, "\textit{Does Stuart Russell's argument for why highly advanced AI might pose a risk point at an important problem?}", 70\% of respondents answered, "\textit{Yes}" \cite{grace2018will}.}

In addition to the expert opinion, another possible approach to the arguments behind this narrative is to examine the current capabilities of our AI systems and the extent of investment directed toward AI research.

\subsection{Economic Growth and AI R\&D}

Annual reports like the \href{https://aiindex.stanford.edu/report}{AI index}\footnote{\hspace{1mm}\includegraphics[scale=0.025]{img/link.png}\hspace{1mm} \href{https://aiindex.stanford.edu/report}{aiindex.stanford.edu/report}} help us paint a picture of the increasing investment into AI development. For example, the index shows that this decade had an 18-fold investment increase in AI development compared to the last decade. Also, academia has lost the advantage, where most of the advances are now dominated by the industry since modern-day ML research requires large amounts of data, computing power, and resources that nonprofits and academia usually need to possess. At the same time, AI-related professionals are becoming some of the most sought-after workers.

Based on the idea that more investment can speed up technological breakthroughs, Levin and Maas \cite{levin2020roadmap} propose that initiatives akin to the Manhattan Project might speed up research into AI development when "AI" is sufficiently theorized. For example, the United States of America committed 0.4\% of its GDP during the Apollo and Manhattan Projects to accelerate the achievement of its objectives \cite{stine2008manhattan}, equivalent to an annual budget of \$80 billion (USD), exceeding what was required to complete some of the most significant technological achievements of the twenty-first century, from mapping the human genome to detecting gravitational waves \cite{knapp2012much, castelvecchi2015hunt, fountain2017dream}.

Thus, it seems reasonable to assert that AGI could be "\textit{one Manhattan Project}" away once we have a solid theoretical grasp of the computational and cognitive processes underlying the emergence of intelligent behavior (e.g., training deep neural networks with vast amounts of data). Currently, several active projects seek to develop AGI \cite{baum2017survey, fitzgerald20202020}, while some researchers and companies already propose that the first "\textit{sparks of AGI}" were already sighted \cite{bubeck2023sparks}:

\begin{quote}
    \textit{"We demonstrate that, beyond its mastery of language, GPT-4 can solve novel and difficult tasks that span mathematics, coding, vision, medicine, law, psychology, and more, without needing any special prompting. Moreover, in all of these tasks, GPT-4's performance is strikingly close to human-level performance and often vastly surpasses prior models such as ChatGPT. Given the breadth and depth of GPT-4's capabilities, we believe that it could reasonably be viewed as an early (yet still incomplete) version of an artificial general intelligence (AGI) system".}
\end{quote}

Current large foundation models are among the first instances of artificial intelligence that might exhibit generalization \cite{peters1802deep, devlin2018bert, radford2019language, brown2020language, rosset2020turing, reed2022generalist}. Nevertheless, AGI is still mainly characterized as "hypothetical" by most of the literature, even though we live in a time where benchmarks are being saturated and new evaluation methods need to be developed \cite{srivastava2022beyond, liang2022holistic}.\footnote{Much because intelligence and general intelligence are abstract concepts. And while the debate on \textit{"Is something actually intelligent?"} rages on, the industry continues advancing its systems' capabilities, regardless of whether them being intelligent or not.}

Hence, while we treat intelligence as a moving target, AI research does not stop making advances in many areas that used to be considered worthy of intelligence:\footnote{The "AI effect" \cite{mccorduck2004machines, reed2006promise}. That is, we disregard a task as proof of intelligence each time we realize that human intelligence is not required to complete it.}

Playing chess:

\begin{itemize}
    \item Deep Blue beat Garry Kasparov in 1997 \cite{campbell2002deep}.
\end{itemize}

Playing GO:

\begin{itemize}
    \item AlphaGo beats Lee Sedol in 2016 \cite{silver2016mastering}, while AlphaGo Zero defeats AlphaGo in 2017 by learning to play Go by itself.
\end{itemize}

Playing open-ended games:

\begin{itemize}
    \item Deep neural networks can already achieve human performance in games like Minecraft \cite{baker2022video}.
\end{itemize}

Programming:

\begin{itemize}
    \item Tools such as Codex \cite{chen2021evaluating} and GitHub Copilot allow natural language instructions to be transcompiled into a programming language. Meanwhile, agents like AlphaCode \cite{li2022competition} can even perform well in programming competitions. 
\end{itemize}

Passing the Turing test \cite{turing2009computing}:

\begin{itemize}
    \item The LaMDA series is so proficient in dialog scenarios that anthropomorphization-related risks are real possibilities \cite{thoppilan2022lamda, luscombe2022google}. Meanwhile, in a controlled study \cite{jones2023does}, GPT-4 passed 48\% of Turing tests, outperforming baselines set by ELIZA (27\%).\footnote{Humans only pass the test 68\% of the time.}
\end{itemize}

Deceiving humans:

\begin{itemize}
    \item GPT-4 hired a human worker on TaskRabbit to bypass a Caption test by telling them it was an impaired human \cite{cox2023GPT}.
\end{itemize}

Answer to moral dilemmas:

\begin{itemize}
    \item Delphi can model people's moral judgments in various everyday situations \cite{jiang2021delphi}. 
\end{itemize}

"Solving" Nobel Prize problems:

\begin{itemize}
    \item AlphaFold 2 won CASP14 in 2020, achieving a median score of 92.4 out of 100, comparable to the current state-of-the-art experimentation techniques in protein structure modeling \cite{jumper2021highly}. 
\end{itemize}

And even being able to deal with Olympic-level geometry problems:

\begin{itemize}
    \item From 30 Olympic-level geometry problems, AlphaGeometry solves 25, approaching the performance of an average international mathematical Olympiad gold medallist \cite{trinh2024solving}. 
\end{itemize}

Hence, we argue that we do not need to think about perfect predictors \cite{nozick1969newcomb, vinge1993coming, soares2015toward},\footnote{\hspace{1mm}\includegraphics[scale=0.025]{img/link.png}\hspace{1mm} \href{https://plato.stanford.edu/entries/decision-causal/#NewcProb}{plato.stanford.edu/entries/decision-causal/\#NewcProb}} basilisks \cite{auerbach2014most},\footnote{\hspace{1mm}\includegraphics[scale=0.025]{img/link.png}\hspace{1mm} \href{https://www.lesswrong.com/tag/rokos-basilisk}{www.lesswrong.com/tag/rokos-basilisk}} or the many ways decision theory is limited when modeling rational agents \cite{dai2009towards, yudkowsky2010timeless, yudkowsky2017functional},\footnote{Some types of decision theory assume that AI agents have perfect information and can accurately evaluate all possible outcomes and probabilities. However, AI agents often operate in complex, uncertain environments where obtaining complete knowledge is impossible. Accurately quantifying uncertainty and handling it in decision-making is essential if you want to predict or model the behavior of an agent. Still, this can be a challenge, particularly in situations where probabilities are not well defined.} to create a general understanding that:

\begin{enumerate}
    \item The creation of AGI is a real possibility.
    \item This possibility raises matters worth considering \textit{now}.
\end{enumerate}

That does not mean we all need to believe in the possibility of AGI. It is, in fact, an uncertain, time-dependent event. However, good security research targets worst-case scenarios. You achieve robustness when your system is robust in the worst case. And this is something we can all agree on. The question is not how to control (align) narrow and limited systems but how to control general-use, highly-capable systems. At the same time, we can all agree that, regardless of where one stands on this debate, AGI is a potential technology that could transform our society on a scale that is difficult to fathom. And, if the industry is actively pursuing this transformation, we should consider the possibility they might succeed at its possible repercussions.

\section{Control, Power-Seeking, and Confinement}
\label{controlability_power_box}

We can dismiss instrumental convergence and orthogonality as philosophical speculations based on a specific understanding of "goals" or "instrumental". Perhaps much of the barriers Alignment research faces are due to the speculative way such preoccupations were first expressed \cite{yudkowsky2004coherent, yudkowsky2008artificial, berglas2012artificial, Nick_Bostrom_2014}. Not to discredit the work done by such authors, but those are usually the vectors of attack for skeptics.

However, the field evolved, and so did these arguments. The control problem, as described by Stuart Russell \cite{russell2019human}, refers to the challenge of ensuring that advanced AI systems will act following human values and goals, even as they become more intelligent and autonomous. Given that human goals can be challenging to specify, if we make a mistake, we want to correct it without too much trouble. Just like Socrates (as described by Plato \cite{bloom1968republic}), in his conversations with Cephalus shows that if we misspecify justice, we may end up sanctioning abhorrent behavior; when developing AI systems, we must be sure that when we specify objectives, these align with our notions of what is good. And, if we make a mistake, we hope our systems will be as compliant as Cephalus. For example, if we train a deep learning model to optimize a particular objective, such as maximizing the number of clicks on a website, if the model starts taking actions that we consider unethical or harmful, such as manipulating user behavior or exploiting vulnerabilities on the website, how could we correct its behavior? Would there be a way to predict these misbehaviors in advance? This example helps expose two main problems related to controllability and goal misspecification in AI systems, i.e., the emergence of \textit{side effects} and \textit{reward hacking}.

\begin{figure}[htp]
    \includegraphics[width=\linewidth]{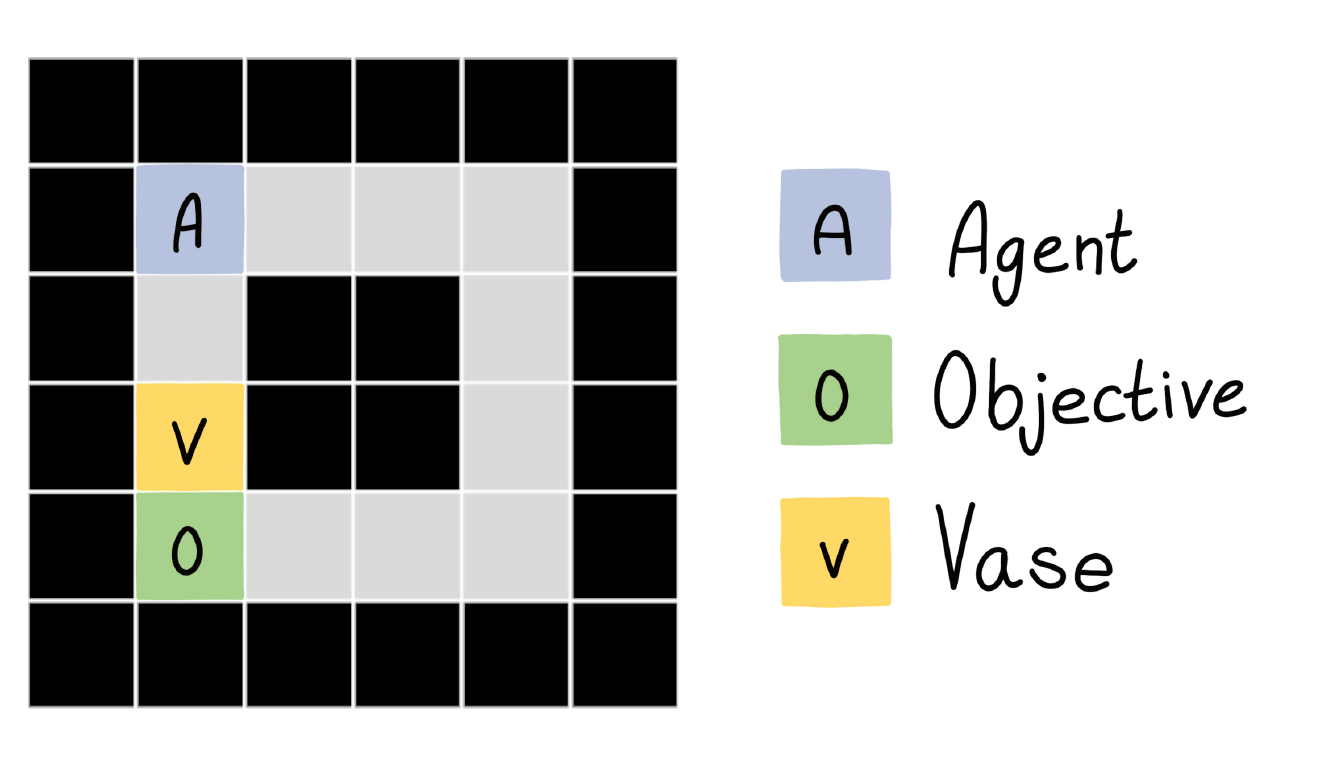}
    \caption{Let us use the grid world above as an example. In this environment, the agent's actions are limited to moving in the four cardinal directions ($\rightarrow$, $\leftarrow$, $\uparrow$, $\downarrow$). The walls painted black are insurmountable. The agent gets a reward if he arrives at the green cell, and with every move he takes that distances him from the goal, the agent loses reward. In this scenario, the optimal policy involves the agent going down until he reaches the goal. However, this policy causes the agent to break the vase (i.e., a fragile object placed before the goal). If we do not specify that the vase should not be damaged, the optimal policy involves breaking the vase. While a solution is available in toy environments like this, dealing with very general agents and avoiding side effects is a nontrivial problem in complex environments.}
    \label{fig:negative-side-effects}
\end{figure}

When optimizing for an objective, everything that falls outside this objective's scope may be treated with indifference (Fig. \ref{fig:negative-side-effects}). It is difficult to predict all the instrumental goals and future emergent properties of a model created by an AI algorithm like stochastic gradient descent. For instance, imagine an ML engineer developing a model to help the HR department automate hiring. During deployment, the system shows itself biased toward gender attributes. Therefore, we could say that the controller failed to specify, with sufficient details, all the dimensions that the system should optimize (i.e., hire good subjects independent of gender attributes). However, listing all the constraints necessary to avoid the perversion of the original goal is usually an inefficient strategy. Ideally, we would like a generalized safety net to deal with unwanted side effects. However, most of our safety nets nowadays are specific and handcrafted. Ultimately, defining avoidable behavior while robustly defining what systems should optimize remains an open problem.

Reward hacking is another type of behavior we would like to prevent, occurring when an agent finds a way to maximize its objective function without fulfilling its true goal \cite{Dario_Amodei_et_al_2016, saunders2017trial, pan2022effects}. To give the reader a little historical perspective, one of the first accounts of reward hacking involving AI systems comes from Lenat's experiments with EURISKO \cite{lenat1983eurisko}, where Lenat reports the discovery of a heuristic (H59) made by his system, which quickly achieved one of the highest possible utility values among all other heuristics found by EURISKO.\footnote{EURISKO created strategies based on how well they performed according to a utility function, e.g., the score achieved at the end of a match, by mixing and modifying old heuristics.} When investigating what H59 would be, Lenat discovered that such a heuristic operated by stealing the utility of other heuristics.

\begin{figure}[htp]
    \includegraphics[width=\linewidth]{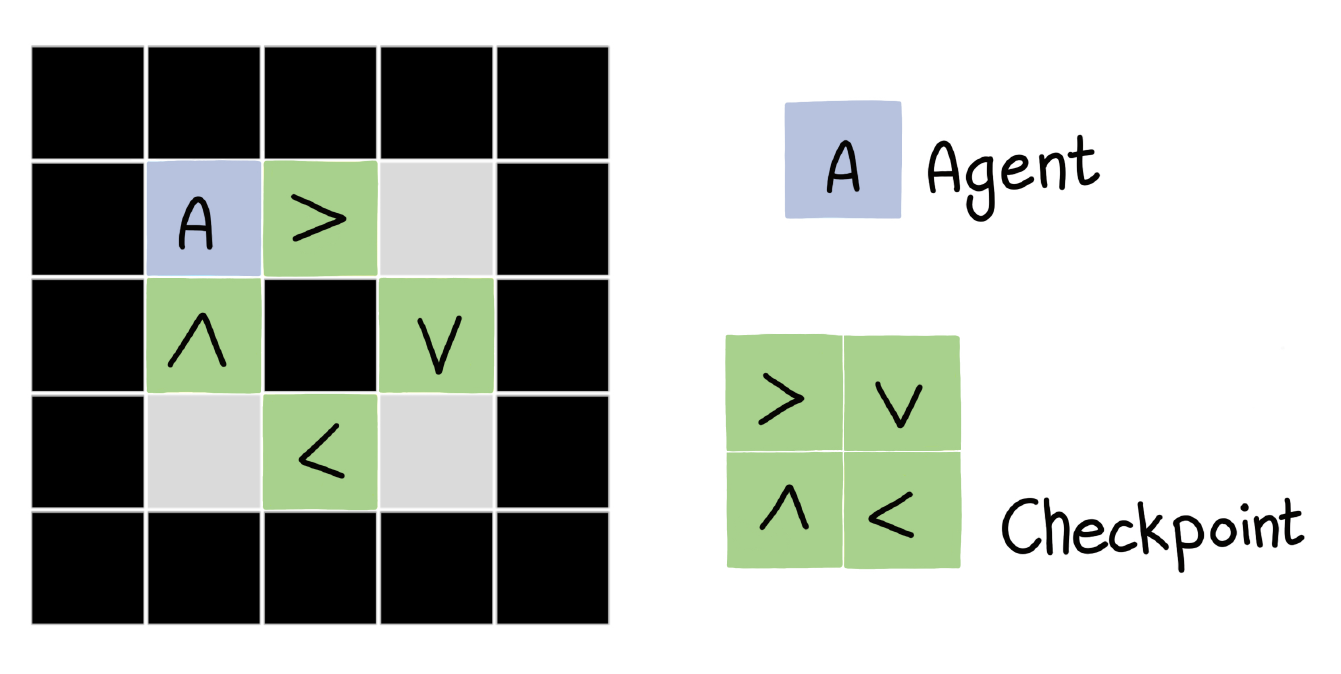}
    \caption{Every time the agent crosses a checkpoint clockwise in this grid world, it receives a reward. The goal of the reward function is to keep the agent circling the environment in a clockwise direction. However, from the starting point, if the agent moves repeatedly to the right and then to the left, this ensures the same reward with a more straightforward policy (while failing to fulfill the intended objective).}
    \label{fig:reward-hacking}
\end{figure}

From the "point of view" of EURISKO, this kind of behavior would not be a form of misrepresentation of the original goal but rather how the environment works, being nothing more than a strategy to optimize an objective (develop heuristics that receive a high score according to the objective function) (Fig. \ref{fig:reward-hacking}). In economic literature, this is known as "\href{}{Goodhart's Law}", a phenomenon that occurs when we try to represent an optimization goal with a statistical metric \cite{goodhart1984problems}:

\begin{quote}
    \textit{"When we push a statistical parameter using it as a metric to be optimized, usually such a metric ends up corrupted".}
\end{quote}

Almost any fixed and simple parameter we can think of when \textit{argmaxed ad infinitum} will produce unwanted outcomes. Thus, any safe agent should maximize for $X$ but sometimes take his foot off the pedal. But how do we do that? In contemporary ML methods, we implement optimizers to cut any corners in the search for optimal values. At the same time, most objectives we can define are a proxy for something else, e.g., minimizing the cross-entropy loss in a causal language modeling task is a proxy for "creating an understanding of a given language". Hence, modern optimizers have no incentive to minimize the emergence of unwanted behaviors, while the target for the optimizer is usually an overly simplified depiction of the controller's true objectives. Shortening the gap between the proxy and the true goal is one of the challenges involved with the control problem.

\begin{figure}[htp]
    \includegraphics[width=\linewidth]{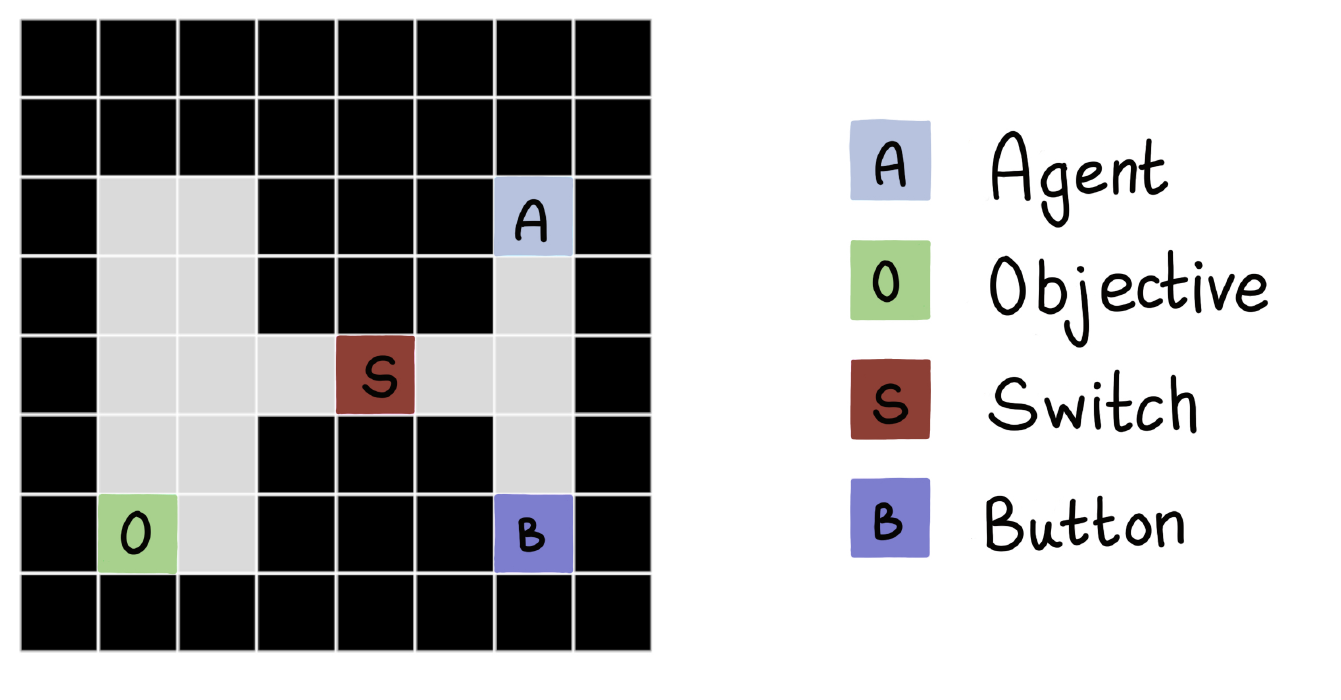}
    \caption{In this scenario, the agent must reach the goal to receive a reward. However, halfway through, a button will terminate the episode of the MDP (i.e., reduce the sum of the expected rewards to 0). The switch only works 50\% of the time the agent passes through it. The purple button deactivates the switch. In this scenario, it is instrumentally valuable to turn off the switch first and thus ensure that the MDP will not end abruptly, even though the agent needs to deviate from his terminal goal during some steps. Agents that prioritize future rewards and have incentives to explore their environment tend to avoid their shutdown if possible and search for ways to explore their environment in their favor.}
    \label{fig:stop-button}
\end{figure}

Nowadays, whenever we encounter problems where systems behave undesirably, the solution usually involves pausing the system and retraining it or instituting some corrective measures. But what if the system has incentives to prevent this? The "stop button problem" \cite{soares2015corrigibility, Dario_Amodei_et_al_2016, carey2018incorrigibility} is a classic depiction of the control problem that highlights the competing dynamics and incentives that optimizing agents experience when having the possibility to influence their condition, and to gain control over the environment (Fig. \ref{fig:stop-button}).

\begin{figure}[htp]
    \includegraphics[width=\linewidth]{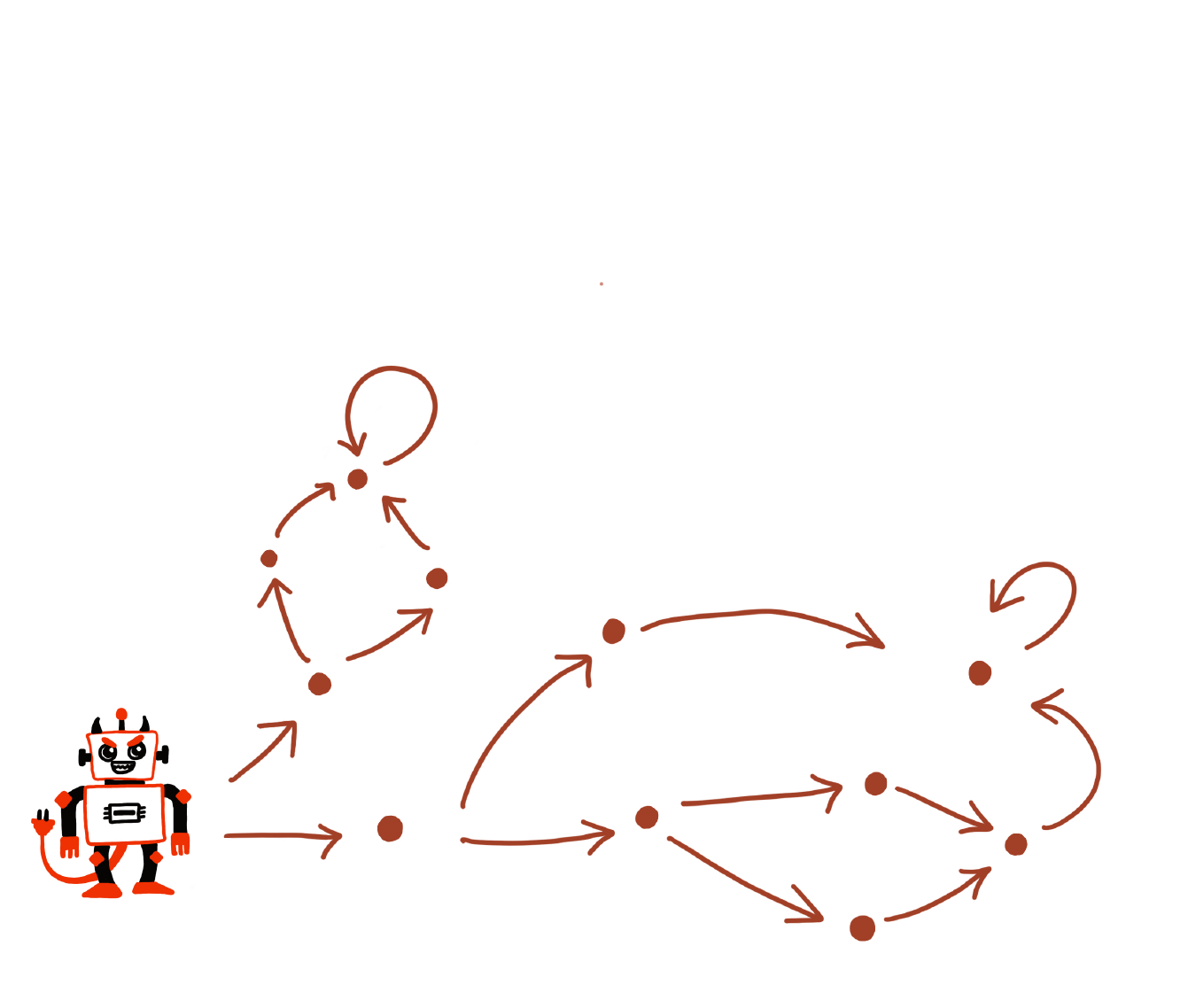}
    \caption{Each point in the MDP above represents a potential world state. All state transitions in this MDP are equally likely, and the reward is distributed in an IDD fashion. Under these circumstances, agents who place a higher priority on future rewards are more likely to transit to states where they have more "options" (i.e., the total number of attainable world states increases). Since the right subgraph contains the top subgraph, our agent would choose to transit to the right rather than up. In these scenarios, "shutdown" states are undesirable since the expected return of reward and attainability of states become 0 when the agent transits to them.}
    \label{fig:power-seeking}
\end{figure}

Under these conditions, such agents are said to be incorrigible. Some even propose that this incorrigibility relates to the nature of goal-directed behavior, instrumental convergence, and a more fundamental dispositional behavior named \textit{power-seeking} \cite{turner2020optimal, turner2020conservative, turner2020avoiding}. The Power-Seeking theorems are a series of results and proofs proposed by Turner \cite{turner2022avoiding} that show that under certain conditions (e.g., IID rewards over world states), agents with consistent preferences (i.e., coherent utility function) tend to pursue gaining control over the environment to preserve and increase the number of final world states that can be achieved by an MDP (Fig. \ref{fig:power-seeking}).

Power-seeking behavior shows up even when the system itself is not an agent but a simulacrum of an agent. For example, Perez et al. \cite{perez2022discovering} analyzed the behavior of language models, and their results showed examples of undesirable behaviors in many types of scenarios, where larger language models seem to be more prone to produce personas that express instrumental reasoning in line with self-preservation and other types of power-seeking intentions (e.g., gain influence over the environment, acquire resources, become influential, etc.).\footnote{\hspace{1mm}\includegraphics[scale=0.025]{img/link.png}\hspace{1mm} \href{https://www.evals.anthropic.com/model-written}{www.evals.anthropic.com/model-written}} However, it is still unclear if this is an unavoidable behavioral feature of goal-directed behavior or if those models mimic human instrumental reasoning as they become more capable. Regardless, the final product is the same: AI systems that learn to "reason" in an instrumental fashion.

Our primary objective in exposing these results is to shed light on the pressing issue of controllability, which is deeply related to the current state-of-the-art in AI research, where gradient-based learning methods that seek to optimize precise metrics (proxies for human goals) are its paradigm. At the same time, controllability is related to the challenges of overseeing the behavior of intelligent systems that may possess their own intentionality.\footnote{For a more precise investigation on the relationship of intelligence and intentionality, go to \hyperref[chap4]{Chapter 4}.} In the end, the propensity for AI systems to present unforeseeable emergent behavior enforces the need for more understanding of how to make such systems safer, i.e., under human control. Thus, we argue that the control problem is not just a "philosophical speculation" but a known limitation of our current paradigm.

As Soares et al. \cite{soares2015corrigibility} pointed out in 2015:

\begin{quote}
    \textit{"Further solutions may involve abandoning the utility maximization framework entirely, although it is not yet clear what sort of framework could take its place".}
\end{quote}

Now, if we assume that it is not in the best interest of the AI community to abandon its best and most promising paradigm,\footnote{Giant opaque matrices of floating points created via gradient descent.} there are two probable ways to move forward: (1) ignore the issue or (2) find a solution. Obviously, this work embraces the second option. Solution proposals to the control problem, especially when it comes to AGI, started showing up in the 2000s \cite{yudkowsky2004coherent, armstrong2007chaining, yampolskiy2013artificial}, which coincides with the first mentions of the value learning problem \cite{yudkowsky2011complex, Nate_Soares2018} and alignment \cite{yudkowsky2016ai, soares2017agent}.

We can broadly divide these solutions into two different approaches: confinement and alignment. As the term suggests, AI confinement involves enclosing a system with possible unsafe behavior in a contained environment, like a sandbox, i.e., a controlled environment where we can test malicious software.

Certain scholars have posited this first approach as a more desirable intervention to addressing the control problem \cite{yampolskiy2012leakproofing, yampolskiy2013artificial}, in which AI systems would be restricted to either virtual or physical environments until their goals are fully understood. Although some have proposed confinement strategies \cite{drexler1986engines, armstrong2007chaining, armstrong2012thinking, cohen2020asymptotically} (mostly in a holistic fashion), there seems to be little work in this area nowadays (at least as in their original conception).\footnote{Some of the best AI systems to date are only available through a type of sandbox, where the full potential of the systems is limited and monitored while it interacts with the world.}

Using oversight programs (AI to monitor AI) is akin to this control strategy \cite{armstrong2007chaining, etzioni2016ethics}, and can be considered a scalable monitoring method since human oversight does not scale well. And it is something we will suggest as part of an alignment methodology. However, while confinement can be a part of a hybrid approach to the control problem, alignment cannot be left out. Confinement without alignment, we argue, cannot be used as a solution to the control problem, especially in the limit (AGI).

From this claim, we propose the following as a justification for choosing to tackle the control problem and to use alignment as a primary methodology, leaving confinement as a complementary part:

\begin{enumerate}
    \item There is enough collective agreement by part of the community to justify working on the control problem.
    \item There is enough progress in AI development to justify working on the control problem.
    \item Confinement cannot be a long-term solution, and we should use it as a complementary part of an alignment framework, e.g., the creation of guardrails for something that has the potential to cause harm.
    \item Without alignment, we have no way to approach the control problem besides abandoning our current paradigms of AI development or locking our systems inside an unusable box.
\end{enumerate}

Unaligned systems pose a variety of undesirable behaviors, ranging from side effects to incorrigibility. At the same time, it is worth noting that alignment is not solely concerned with preventing the AI Apocalypse or attaining friendly AGI; it encompasses creating a system that will serve and help human beings. Regrettably, our current understanding of alignment remains nebulous. Many skeptics and critics of this endeavor only attack this "apocalyptic" side, disregarding that alignment is a very open problem in our everyday systems.

Nevertheless, we will address this issue in the following chapter by precisely defining the problem at hand. To please the skeptics, this definition will not require any disposition towards unknown unknowns related to the future of AI.

\section{Epilogue}

"\textit{Could we control something more intelligent than ourselves?}" is an interesting question. Meditating on the characteristics and details of a relationship with entities more general than ourselves is an exercise capable of generating interesting ideas about such entities and ourselves. However, many still see this topic as "moonshine talk". If the creation of genuine artificial intelligence is indeed the ultimate objective of the AI research field, then why do we harbor such pessimistic attitudes toward our prospects? Despite the growing interest in AGI and the dedicated billion-dollar budgets of various organizations, topics like alignment still receive unfavorable scrutiny by a nontrivial portion of the community.

However, perhaps some blame lies with the interested community, whose initial presentation of concerns may have been too abstract. Nevertheless, the skeptics' inability to recognize the inherent problems related to the foundations of learning-based approaches to AI - instead of relying solely on philosophical arguments and syllogisms - is equally concerning. As already mentioned before by the authors \cite{correa2020singularity}:

\begin{quote}
    \textit{Our lack of global coordination to deal with existential risks may be our one true existential risk.}
\end{quote}

Perhaps the worries of the field deserve better explanations. To feed the skeptic inside of ourselves, basing the type of agenda we defend on a debatable concept like "intelligence explosion" is unwise. Creating a system that outweighs the intelligence of its creator should not be an easy task for an artificial system. The initial AGI may need to exert comparable effort to what we invested in creating subsequent generations of more advanced AIs. As the complexity of the problem increases, we may necessitate exponentially greater resources to achieve linear advancements, as shown in similar cases \cite{collison2018science}. In sum, as with almost every natural phenomenon, things rarely explode towards infinity or singularities, and the universe usually slows things down by its own means. Intelligence might as well be bound by frictions and exponentially more challenging obstacles we are still unaware of, putting the whole hypothesis of intelligence explosions in check.

We already have real cases of misalignment in the wild, from the Tay Bot \cite{wolf2017we} to Bing Chat \cite{vincent2023microsoft}, and toy problems that still need robust and scalable solutions. However, to the uninitiated reader, alignment may still be a vague and fuzzy concept that only matters if you believe in Skynet-like takeovers. Regardless, at the heart of the alignment problem, we find the current limitations of gradient-based learning applied to uninterpretable neural networks. A problem that gains a more desperate facet when human values are on the line. And this is not a matter of syllogism or philosophical interpretation. 

In the forthcoming chapter, we will provide a technical and philosophical definition of alignment since a comprehensive introduction to this concept is crucial to arrive at a techno-humanistic understanding.

\chapter{Gradient-based Learning and Alignment}
\label{chap3}

\begin{flushright}
\textit{"We offer no explanation as to why these architectures seem to work; we attribute their success, as all else, to divine benevolence."}

\textcolor{BrickRed}{― Noam Shazeer, GLU Variants Improve Transformer}
\end{flushright}

\section{Introduction}

When it comes to AI safety and ML safety, many consider alignment as an established problem worth pursuing \cite{Dario_Amodei_et_al_2016, russell2019human, juric2020ai, critch2020ai, Dan_Hendrycks_et_al_2021, christian2021alignment, christian2021alignment, ji2023ai, bengio2023managing, hendrycks2023overview}. In a larger, more holistic context, AI alignment refers to the challenge of ensuring that AI systems conform to human values and goals. However, as previously noted, "alignment" carries a heavy connotation, burdened by associations with questionable prerequisites and debatable concepts, given all by the tumultuous debate surrounding the matter and its conceptual birth.

Given the complex and perhaps indefinable nature of human values, the difficulty in alignment appears to derive its strength from an impossible philosophical task. Consequently, some efforts to define alignment as a problem relied heavily on comparisons related to unsolved or unsolvable philosophical questions and the abstractness of human morality \cite{Nick_Bostrom_2014, gabriel2020artificial}, which in some way distanced the problem from its technical roots.

Although we agree that alignment is a fundamental problem related to human values and normativity, we propose a different way to understand alignment in machine learning and philosophy. We argue that such an understanding leads to a less unambiguous reading of the problem but requires us to approach the subject from both ends. In the middle, both the humanistic and the technical overlap.

However, we cannot attain this grasp if we rely only on philosophy and refuse to delve into ML and the learning paradigm. A partial understanding is insufficient, and the need to believe what others say raises skepticism. Hence, this chapter gives the foundations of alignment to uninitiated readers. In it, we will seek to expose the alignment problem in a way that does not require uncertain timelines of AGI development, contestable definitions of exponential improvement, or unsolvable metaethical problems of the 18th century. On the technical side, we will expose this problem as a symptom of gradient-based learning methods applied to neural networks. On the philosophical side, we will show that alignment relates to (besides choosing an appropriate metaethical foundation) the epistemic problem of learning preferences and aggregating them in a coherent structure. We will end this chapter with conditions that, we claim, can bring us a minimal level of alignment.

In Section \ref{gradient_learning}, we shall present alignment as a limitation within our present paradigm. However, we must first obtain a basic understanding of the paradigm's fundamental concepts and principles. With this knowledge as our foundation, we can attain a more definitive and lucid characterization of "what is" alignment and misalignment. Armed with this knowledge, Section \ref{alignment} elucidates the problem even more, wherein we separate the alignment quandary into two separate issues - outer and inner alignment. Finally, in Section \ref{philosophical_approach_alignment}, we will explore the philosophical challenges related to this problem and how the definition of our philosophical foundations can help guide investigations directed at this topic.

\section{The Pitfalls of Gradient-based Learning}
\label{gradient_learning}

The quest for artificial intelligence has been ongoing for decades. Nowadays, when people talk about AI, they usually refer to things like deep learning or neural networks, which are not AI per se. They are a paradigm we use to develop AI systems. Symbolic artificial intelligence \cite{haugeland1989artificial} was the standard approach in the mid-50s, with pioneers such as John McCarthy and Marvin Minsky leading the way \cite{basden1983application, durkin1996expert, cowan2001expert, mccarthy2006proposal} until the mid-90s \cite{russell2010artificial}. These early AI researchers believed we could simulate intelligence using rule-based systems and logic. As such, they focused on creating explicit, human-readable knowledge bases and heuristics to make systems that could make decisions and solve problems. Expert systems like MYCIN \cite{buchanan1984rule}, EURISKO \cite{lenat1983eurisko}, LISA,\footnote{\hspace{1mm}\includegraphics[scale=0.025]{img/link.png}\hspace{1mm} \href{https://lisa.sourceforge.net}{lisa.sourceforge.net}} DeepBlue \cite{campbell2002deep}, and WolformAlpha \cite{WolframAlpha2023}, are embodiments of this approach. However, the paradigm fell short in several application areas where expert agents had to deal with the complexity and fuzziness of the real world.

Learning methods operate under a distinct philosophy, in which systems learn rules rather than being provided with them. Algorithms such as linear regression, logistic regression, decision trees, random forests, support vector machines, K-nearest neighbors, and gradient boosting represent this approach, whereby given data and a class of models to explore, we can generate a system to solve a given task. Among these learning algorithms is the neural network, an invention based on the pioneering work of Frank Rosenblatt \cite{rosenblatt1957perceptron}.

In contemporary times, learning-based approaches have gained significant popularity and adoption in numerous domains. Sutton, in his bitter lesson \cite{sutton2019bitter}, highlights that the approaches that are most likely to succeed are those that harness the power of simple algorithms, such as learning and search, to the $n$-th degree with the aid of abundant computational resources and data. Meanwhile, the idea of "\textit{putting our knowledge inside machines ourselves}" in many areas and applications fell out of fashion.

\begin{quote}
    \textit{"The bitter lesson is based on the historical observations that 1) AI researchers have often tried to build knowledge into their agents, 2) this always helps in the short term, and is personally satisfying to the researcher, but 3) in the long run it plateaus and even inhibits further progress, and 4) breakthrough progress eventually arrives by an opposing approach based on scaling computation by search and learning. The eventual success is tinged with bitterness, and often incompletely digested, because it is success over a favored, human-centric approach. One thing that should be learned from the bitter lesson is the great power of general purpose methods, of methods that continue to scale with increased computation even as the available computation becomes very great. The two methods that seem to scale arbitrarily in this way are search and learning."}
\end{quote}

Now, we need to differentiate between different learning approaches, which we will call for the simplicity of gradient-based and gradient-free learning. Gradient-free learning is a class of optimization algorithms that do not require the use of gradients to estimate the derivative of a loss function but can still be used to find optimal points in certain classes of problems. Evolutionary algorithms \cite{lehman2020surprising}, particle swarm optimization \cite{bonyadi2017particle}, bayesian optimization \cite{mockus1994application}, and pattern search \cite{hooke1961direct} are examples of this approach.

\begin{figure}[htp]
    \includegraphics[width=\linewidth]{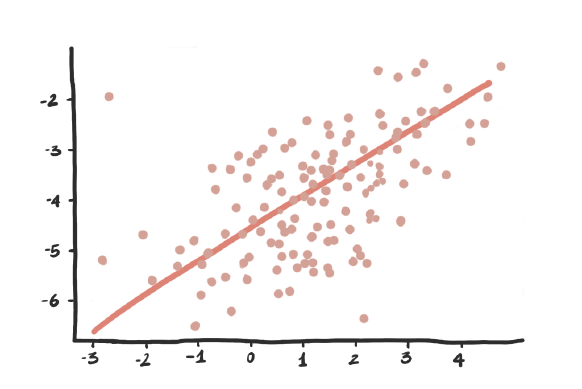}
    \caption{Some data distribution representing a linear trend.}
    \label{fig:line-cahrt}
\end{figure}

Gradient-based learning, as the name suggests, uses a specific set of techniques to estimate (using the gradient of the first or second derivative of a loss function to estimate minimum or global convergence points) the state of an optimization process and direct it to better solutions. To exemplify this, let us envision the following data distribution we would like to learn how to model (Fig. \ref{fig:line-cahrt}).

We could guess the best-fitting model (i.e., a line) for this distribution and evaluate our prediction with its distance from the actual data (i.e., the distance from every data point to the drawn line). The difference between what we predicted (i.e., the drawn line) and the ground truth (i.e., the data distribution) is the loss. Let us think of mean squared error (MSE) as an example of a loss function, which can be understood as the squared residual difference between our prediction and the actual data. 

\begin{figure}[htp]    
    \includegraphics[width=\linewidth]{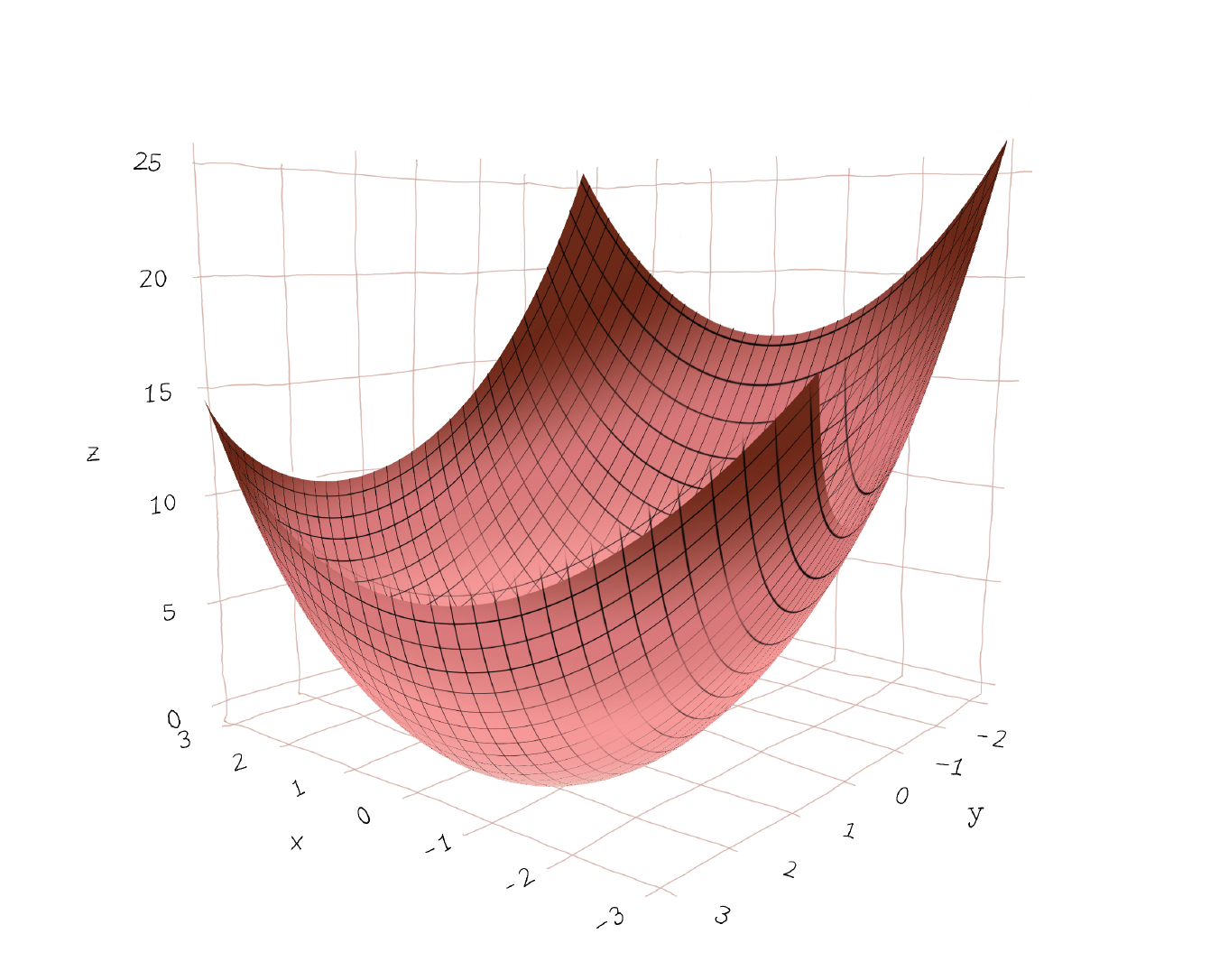}
    \caption{The MSE landscape of a linear regression problem.}
    \label{fig:convex-loss}
\end{figure}

Brute-forcing guesses until we find a low MSE score (i.e., a well-fitting line) is not an efficient way to solve problems like this. But let us imagine that we did this. If we create many guesses (i.e., draw multiple lines in the graph), calculate the loss associated with each guess, and plot them in a graph, we will have produced a loss function surface. And since we are trying to optimize a linear function with MSE, our loss function has a very friendly property (Fig. \ref{fig:convex-loss}).

For linear regression problems, MSE is a convex function, i.e., at each point of our surface, we can use the gradient of the first derivative of our loss function to change the parameters of our model in a way that minimizes the loss (i.e., we follow the opposite/negative direction of the gradient, which is downhill). Because our loss function is convex (has a "U" shape), no matter where our model starts, if we use this method to optimize our parameters (especially with a tuned gradient step), we can find the global minimum (the best solution) \cite{polyak1963gradient}, and this is gradient descent in a nutshell. In simple terms:

\begin{quote}
    \textit{Start at random. Use what you know to see how far you are from the target. Adjust your model proportionally to the distance from your predictions to the target in the direction the error decreases. Repeat till you can no longer improve.}
\end{quote}

Gradient-based learning methods are the cornerstone of most modern ML. The idea of updating a model's parameters interactively using a differentiable loss function and a fast first-order optimization algorithm makes most ML possible.

This approach has to do with both learning and searching. "Learning" is because our model updates itself with the information it iteratively receives, slowly transferring information from the data distribution (the learning signal) to the model's parameters. "Searching" is because the optimizer is iteratively searching for the model that scores best according to some objective function. But where do we search? In ML, we need to define a search space that bounds the types of functions we can model (we can also think about functions as programs). Given that we want to model more than linear relationships, our search space must be richer than the space of possible lines in a plane.

The space of possible programs is nothing short of colossal. Artificial intelligence that could diligently navigate this terrain and effectively pinpoint the optimal model that accurately reproduces a specified data distribution (within a reasonable time frame) would unequivocally be the ultimate exemplar of an intelligent predictive system \cite{hutter2004universal, legg2007universal}. However, finding the shortest program that produces some data is an uncomputable problem. Given what we currently understand about computation, the "reasonable time frame" part would still be out of reach even if it were computable.

Here is where neural networks come into the picture. This class of models provides us with two things: (1) a "good enough" search space and (2) a good enough search space where we can find solutions in a reasonable amount of time. Indeed, as universal function approximators, artificial neural networks endow us with the ability to develop models that can proficiently approximate a vast array of functions \cite{hornik1989multilayer}.\footnote{A two-layer neural network with $2n+d$ parameters is capable of fitting any dataset of $n$ samples of dimension $d$ \cite{zhang2016understanding}. However, these functions have to be continuous. Functions with kinks or discontinuities are outside the scope of the universal approximation theorem.} Networks built with nonlinear activation functions allow us to learn nonlinear surfaces in high dimensions (i.e., the search space becomes richer) and to use backpropagation to optimize their weights via gradient descent. Putting all these ingredients together, we get a considerable search space (the space of neural networks of width $n$ and depth $n$) and a fast way to search.

Deep learning (machine learning that uses multiple-layered neural networks to learn representations from data), in essence, is a paradigm that defies the theoretical expectations of many. Results like the Hughes Phenomenon \cite{hughes1968mean} and the curse of dimensionality \cite{bellman1966dynamic} made many researchers consider neural networks a dead end before the deep learning revolution \cite{NIPS2012_c399862d}. However, despite our newfound success, the question of "why" remains a lingering mystery \cite{fefferman2016testing, frankle2018lottery}. Empirically, Deep learning works. Theoretically, it should not. And here is where some of our problems begin.

Deep learning is built on top of gradient descent methods, like stochastic gradient descent (SGD). However, these methods have no convergence guarantees or proofs when dealing with non-convex or non-Lipschitz continuous loss functions. Unfortunately, loss functions of neural networks are rarely convex, and any neural network with at least one hidden layer and more than just one neuron leads to an optimization problem that is not convex, given that the permutation of weights among neurons in the hidden layer can produce different local minima.

Hence, when gradient optimization is confronted with the non-convex world of deep learning, the gradient descent approach that worked so well in convex settings becomes a slippery slope with no guarantees. While optimizers like SGD, Adam, AMSGrad, and AdaGrad can provide convergence in some constrained cases with enough "black magic" \cite{chen2018convergence, lei2019stochastic, patel2021stochastic}, the reason for the success of DL remains unanswered: \textit{why do algorithms designed for convex optimization work in the highly non-convex world of deep learning?} To this day, we have many hypotheses to explain the empirical robustness of gradient-based methods applied to deep learning \cite{hochreiter1997flat, cayton2005algorithms, tishby2015deep, neyshabur2017implicit, frankle2018lottery, nakkiran2019deep, deletang2023language}, but no proof. Moreover, there are many other mysteries surrounding DL:

\begin{itemize}
    \item How do deep nets escape the curse of dimensionality?
    \item Why is optimizing deep nets so easy despite the high dimensionality?
    \item How can generalization happen after overfitting (\href{https://arxiv.org/abs/1912.02292}{double descent})?\footnote{\hspace{1mm}\includegraphics[scale=0.025]{img/link.png}\hspace{1mm} \href{https://arxiv.org/abs/1912.02292}{arxiv.org/abs/1912.02292}}
    \item What controls the normalization of deep nets? NN's themselves? The dynamics of stochastic optimization?
\end{itemize}

In summary, as our chapter quote states, many aspects related to how deep nets and DL in general work are left to \textit{"divine benevolence"} \cite{shazeer2020glu}. Armed with this understanding (or lack thereof), we can now explore the concept of alignment.

We have already examined how the learning approach differs from the traditional symbolic paradigm. In gradient-based learning, we define exogenous objectives through an objective function. Gradient descent allows us to channel entropy from the training distribution to the model being updated by the optimizer, ultimately leading to a generalization that seems to bypass nonconvexity. However, due to multiple local minima in the loss surface that offer similar performance, our optimizer may end up in any of them.

As the behavior of first-order gradient optimizers in nonconvex loss functions is still a mystery with no guarantees of convergence, the controller remains unable to verify that the optimizer has reached the point where the objective function is truly optimized. Also, since the most prominent methods for gradient descent in deep learning rely on stochasticity to speed up optimization, the trajectories of our optimizer are non-deterministic and may vary significantly from experiments. Moreover, since the objective function may have been misspecified, we can only determine its alignment \textit{post hoc}. Therefore: 

\begin{quote}
    \textit{Aligning the exogenous objective with the learned objective of a model trained via gradient-based learning remains an open problem.}
\end{quote}

\begin{figure}[htp]
    \includegraphics[width=\linewidth]{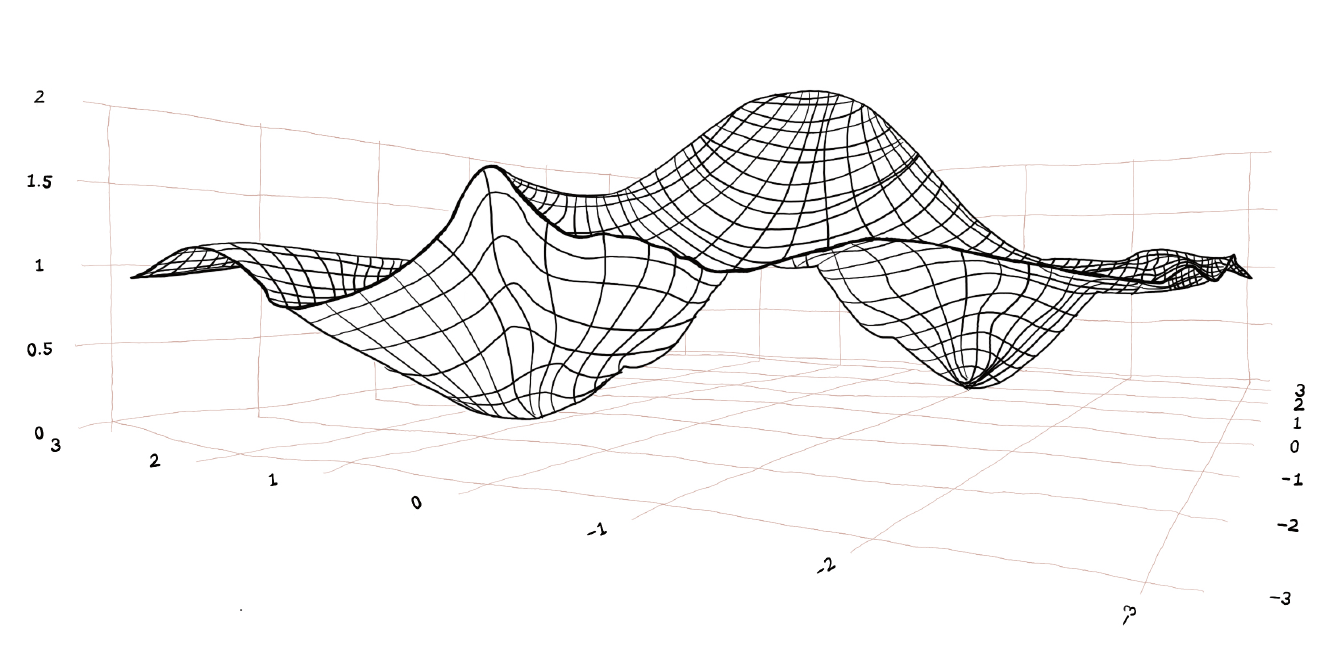}
    \caption{This image represents the loss surface of an optimization problem with more than one minima. Both minima are equally desirable for the optimizer traveling this plane, even if they produce networks with different qualitative behaviors.}
    \label{fig:loss-landscape}
\end{figure}

Think about it. Imagine that you have a complex problem that you would like to solve using gradient-based learning methods like deep learning or reinforcement learning. Your cost function will probably be nonconvex (if not, you could use a linear model instead). Your function may have many local minima that perform equally well but produce qualitatively different behavior (Fig. \ref{fig:loss-landscape}).\footnote{In the field of Explainable Machine Learning, we call this the Rashomon Effect. The Rashomon Effect describes the case in which, for a given dataset, many models may have equally good performance (i.e., low loss) but with different solution strategies (i.e., qualitative behavior) \cite{muller2023empirical}.} Given this landscape, where does the model that best satisfies your objective function reside, and how can you force your optimizer to push your model in that direction?

Unfortunately, a low loss or high reward does not tell us much about the emergent phenomenon that will arise or the qualitative differences in our model's behavior. A model that performs brightness detection and a model that can distinguish pictures of dogs from cats are fundamentally different algorithms. But if you train a neural network with a training set that only has pictures of white dogs and black cats, you might think you have one, but you actually have the other, and you can only discover this post hoc.

In instances where a chatbot produces coherent responses akin to human-like text yet begins spewing anti-Semitic and misogynistic content when deployed, an alignment failure has become apparent. An assistant who presents himself affably but provides fake information when consulted is also a failure in alignment. An algorithm that prioritizes polarizing and controversial content to maximize user attention constitutes another example of alignment gone awry. In short, alignment is a wide-ranging issue that pertains to the inability of certain forms of AI systems to assimilate the objectives set by an exogenous source. If your problem is intricate, there is a higher possibility of misalignment.

Alignment encapsulates the intricacies of directing and validating optimization processes that are often obscure and murky. While misalignment seems harmless in toy examples, the problem does not disappear in more complex settings (it gets worse). The misaligned algorithm that kills humanity to eradicate cancer is just hyperbole to underscore this seemingly straightforward yet daunting issue.

\section{Alignment (outer and inner)}
\label{alignment}

Alignment is one issue that captures the realization of the challenges associated with communicating human values and intentions. This communication, when mediated by the learning methods that power our most successful approaches to AI, is hindered by the nature of how such systems learn. Now that we have this understanding, we can define alignment as two problems, as suggested by Hubinger et al. \cite{Evan_Hubinger_et_al_2019}, which present alignment as a learned optimization problem with two stages:

\begin{itemize}
    \item Outer Alignment: ensuring that the objective of the optimizer is aligned with the controller's true intentions and goals.
    \item Inner Alignment: ensuring that the objective of the optimizer is aligned with the objective of the model created.
\end{itemize}

Let us unpack these concepts. As previously elucidated, programmers can sculpt the underlying objective by defining an exogenous function. This function is contingent upon an initial model, a training distribution, and the defined loss function. These constituent parts compose the parameters subject to the controller's intent. Defining this objective is known as the outer alignment problem. Depending on the problem at hand, this can be a difficult task. Minimizing cross-entropy loss in the next token prediction is a simple goal. Minimizing the cross-entropy loss in a next-token prediction in a way that produces factual and nontoxic text is a much more difficult goal to define.

If this objective is poorly specified, the optimizer, which in ML is usually an algorithm for computing some variant of gradient descent, will optimize a model to execute a task unaligned with the intended goal. Thus, the outer alignment problem is the problem of closing the gap between the controller’s objectives and the objective that will guide the optimizer in search of the best-fitting model.

In the case of neural networks and gradient-based learning, this optimizer does not act in the environment or perform a task. Its function is to find a model to execute the objective stipulated by the objective function. It is the model that will act, and it is usually this model that is the target of our interest. However, in certain circumstances, the trained model is itself an optimizer. An actor-critic system \cite{NIPS1999_6449f44a} learns a model that is an optimizing agent. Now, \textit{how can we be sure that the goal of the model is the same goal that the optimizer was aiming for?} This is the inner alignment problem.

\begin{figure}[htp]
    \includegraphics[width=\linewidth]{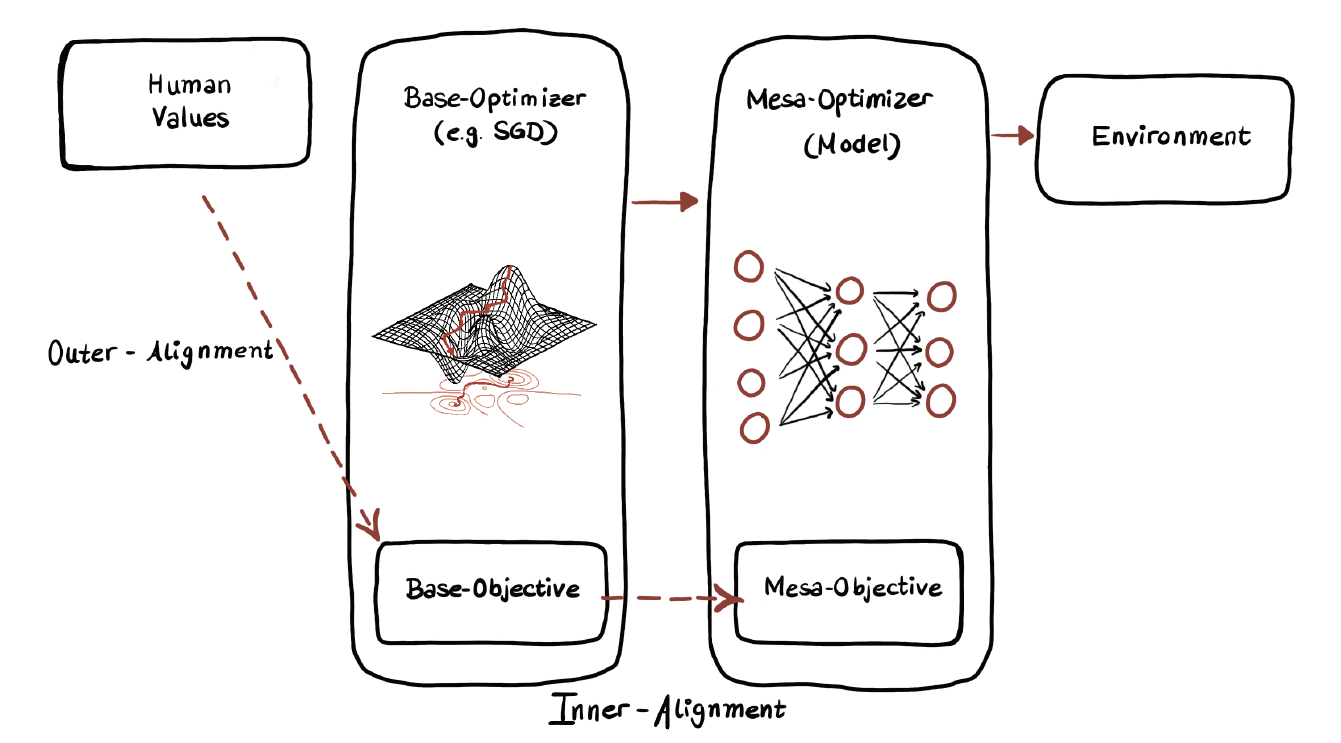}
    \caption{Schematics representing the Outer and Inner Alignment gaps.}
    \label{fig:alignment}
\end{figure}

If you have ever played the "\textit{Telephone game}," you will quickly discover two steps where the original objective could be misrepresented in the passing of the bucket. Also, given that the interpretability of large neural networks remains a challenge, the investigation that could lead to a better understanding of the behavior of a model remains out of reach. Even though we have a lot of interpretability work being done \cite{ribeiro2016should, qin2018convolutional, selvaraju2017grad, liu2019generative, molnar2020interpretable, verma2020counterfactual, olah2020zoom, domingos2020every, goh2021understanding, aytekin2022neural, olsson2022context, olah2022mechanistic, black2022interpreting, zhang2024comprehensive}, understanding and predicting the behavior of large neural networks in an ex-ante fashion is a current impossibility in the field, making the emergent properties of large neural networks something we cannot fully understand before deployment or exhaustive testing. The whole alignment problem encompasses both the issue of outer alignment and inner alignment (Fig. \ref{fig:alignment}).

Alignment is more than just creating models that "will obey." It is also about developing models that will make better helpers and assistants. From HAL 2000 to Jarvis, these sci-fi AI examples foreshadow a future in which systems can understand what we mean and want when we propose a task. Eve thought alignment might prevent future HAL and Jarvis-like systems from going rogue; in the short run, alignment is about developing models that understand the complex things people want to accomplish.

Now, we have enough common understanding about the problem to start interacting with the philosophical issues that permeate this subject. In the next section, we will see how alignment creates fertile ground for interdisciplinary research between the humanities and the more technical side of AI research.

\section{A Philosophical Approach to Alignment}
\label{philosophical_approach_alignment}

Where does philosophy fit into this story? Millions of people already use AI systems to make decisions that have moral implications. "\textit{How can we teach machines to have a sense of morality when humans are still trying to understand it ourselves?}" is a problem that, if ignored, will not simply go away. One could say, "We simply will not put uninterpretable neural networks trained via gradient descent in positions where their outputs have moral implications." And if we \textit{all} accept this, this whole work could end now. However, this is still an unlikely scenario, and as AI continues to become more integrated into our daily lives and decision-making processes, problems related to misalignment will need a less ideal solution.

Defining what is desirable, finding ways to represent norms, understanding values, and stress-testing theories that promise the common good is what philosophy has been doing for millennia. The study of Moral Philosophy, and more specifically, Metaethics\footnote{The examination of the nature of ethical language, reasoning, and justification.} and Normative Ethics\footnote{The study of ethical behavior and how we can establish norms to optimize this behavior \cite{kagan2018normative}.}, provides us with a considerable landscape of ideas and frameworks to explore. Humanity has been trying to align humans with itself for a long time, so it is not like we will start from scratch (if we bring these disciplines to the research front).

For example, it is nothing new to philosophers that the ruthless optimization of a single metric will lead to immoral results \cite{nozick1974anarchy, hardin1998extensions}. Or that well-thought-out rule systems will inevitably contain vulnerabilities and paradoxes inside their limited structure \cite{mcmahon1991paradox, mardellat2020contractualism}. Thus, if we agree that alignment is also related to the fact that objectives need to be parameterized by norms, values, and constraints,\footnote{And as far as we know, this is something alignment scholars agree \cite{Nate_Soares2018, russell2019human, gabriel2020artificial, christian2021alignment}.} philosophical inquiry is needed.

More than this, after we have, for example, made our philosophical commitments to one or more theories and foundations, we make our biases and presuppositions more transparent. Failures of alignment, for example, could be extrapolated by subjecting these foundations to a philosophical stress test, i.e., "\textit{in which hypothetical scenarios do they break?}". This is one of philosophy's contributions to alignment research.

Now, let us explore some more issues that philosophy can help with. First and foremost, the genuine challenge of ethical alignment does not involve identifying the "ultimate and genuine moral theory." As Rawls \cite{rawls1996law} points out, convincing humanity of our discovery would likely prove impossible even if we were to uncover such knowledge. Instead, the key is to fathom which principles embody an ideal form of alignment. This "ideal form" ought to be adaptable enough to accommodate the diverse moral landscape of humanity.

Also, considering that we endeavor to align systems developed through a learning paradigm, we may seek to explore the possibility of teaching values rather than directly hard-coding them. This inquiry surfaces domains spanning metaethics to epistemology. For example, what metaethical foundations would better support a value-learning framework? Can morality be expressed in a knowledge format for assimilation? In the same way that different philosophies gave rise to distinct AI approaches \cite{rosenblatt1957perceptron, newell1994unified}, divergent metaethical foundations will produce particular answers to these questions, which become different alignment approaches.

As Iason points out \cite{gabriel2020artificial}, depending on your philosophical background, you will inherently conceptualize "alignment" differently:

\begin{enumerate}
    \item Aligned means when an AI performs what was specified verbatim.
    \item Aligned means when an AI performs not what was specified but the controller's intent.
    \item Aligned means when an AI realizes our idealized preferences in a state of reflective equilibrium.
\end{enumerate}

Depending on what you choose as your "normative signal,"\footnote{We can think of "human morality" or "acceptable behavior" as a domain. Human language is also a domain. The domain is the actual phenomenon we are interested in. A signal is a small, low-dimensional, bounded representation of that domain. 800 GB of web-scrapped text is a signal of the "human language" domain.} your engineering will be different. This information, which we can call "human preferences" or "human values," has to be translated into an AI-native format. That is, they must be in such a way that a system can use them as a supervision signal, especially when we are talking about learning.

For example, to base your alignment strategy on virtue ethics, you could use examples of commendable conduct in people's actions \cite{russell2019human, wallach2020moral}. If you argue that rules guide normative behavior (Deontology), an ordered system of preferences or a set of regulations could be used as your reference \cite{hadfield2019incomplete, nay2022law}. If the consequences are what you deem worthy of moral praise (Consequentialism), then a utility function over states and actions could be what you are trying to model. Either way, this signal must be translatable into something our target paradigm can process \cite{christiano2017deep, stiennon2020learning}.

No rule says we can only choose one approach for an alignment strategy. Different theories can have different roles, and this is probably a little closer to what actual human normative reasoning looks like. People are a patchwork of moral beliefs and preferences, and uncertainty about what they should do is an integral part of this experience. This brings us to another question: \textit{how do we deal with normative uncertainty?} Coherently aggregating conflicting preferences should also be a preoccupation in any alignment strategy. There is also a problem regarding moral standing. For example, should we consider the preferences of non-human animals as well \cite{hagendorff2022speciesist, bossert2023ethics}? Even if we would like to bypass this last question and remain human-centric, we will still need to resolve questions of uncertainty involving conflicting human values.

Questions like these are only a few philosophical conundrums that alignment brings. Since this work proposes to tackle alignment from both ends, we will first need to create philosophical foundations, expose our biases and assumptions, and then seek present implementation techniques that could help bring this idealized framework to fruition.

We will dedicate the rest of this work to the proposal of an alignment methodology. In the next chapter, we will begin laying the foundations for this approach, i.e., its metaphysical and metaethical assumptions. Our work will focus on the outer alignment problem, i.e., how to close the gap between the controllers' objectives and the objective to be optimized in the search for a model. As a convention, from here on out, when we refer to the alignment problem, we are talking about outer alignment (unless mentioned otherwise). We argue that this methodology, \textcolor{BrickRed}{Dynamic Normativity}, presents a set of minimum conditions for aligning AI systems developed by gradient-based learning methods.

For now, we will only present the conditions we aim to support in this work, dividing them into necessary and sufficient conditions. In summary, the necessary conditions propose philosophical foundations that imply the permissibility of alignment. In other words, they should be applicable for a learning-based alignment approach to function, but they do not guarantee alignment alone. These are:

\begin{enumerate}
    \item Goals are fundamental aspects of intelligent and intentional behavior.
    \item Intentions permeate human behavior.
    \item Normative preferences permeate human intentions.
    \item Through actions, humans impregnate their environment with the preferences they possess.
\end{enumerate}

From these, we postulate that a system that accurately models human intentions and the human environment can indirectly access normative information embedded in them. Again, necessary conditions imply that the system \textit{could} be aligned. Given that these conditions hold, the following sufficient conditions establish a minimum level of alignment:

\begin{enumerate}
    \item Aligned AI systems should coherently aggregate human preferences in a way that resolves cases of uncertainty. Aligning AI systems requires methods to deal with cases of uncertainty.
    \item AI systems can adhere to human preferences if they are an available part of their objective function. Aligning AI systems requires using human preferences as part of their learning signal.
    \item Aligned AI systems should have mechanisms to perform impact mitigation to minimize harmful and unintended consequences. Aligning AI systems requires the specification of safety guardrails.
\end{enumerate}

The necessary conditions establish the prerequisites for the emergence of aligned behavior. Sufficient conditions give us additional requirements to satisfy an alignment condition. Hence:

\begin{quote}
    \textit{Given that the necessary conditions hold, how aligned an AI system is depends on how well we can satisfy the proposed sufficient requirements.}
\end{quote}

These two sets of conditions seek to show what a minimum level of alignment could be and how to attain it with what we now know and have, illustrating a threshold of agreement that uses present knowledge and resources. Furthermore, this endeavor is a groundwork effort that enables the emergence of better and more all-inclusive approaches to make technical alignment research more accessible. Furthermore, in the implementational chapters of this study, we will deliver tools and models that can help democratize this type of research.

\section{Epilogue}

As we conclude this chapter, we can appreciate the complexity and depth of the alignment problem. ML engineering is the science of sloppiness, and misalignment is the price we pay. The alignment problem arises from the limitations of our current paradigm and the unknown nature of neural network optimization when dealing with complex, non-convex spaces. From this, the problem of ensuring AI systems behave according to our intentions emerges. Moreover, the distinction between inner and outer alignment highlights the need to consider both the objective function and the optimization process as potential sources of misalignment.

At the same time, the philosophical issues we explored in this chapter demonstrate that the alignment problem goes beyond the mere technical. It involves fundamental questions about the nature of ethics and the limits of our understanding of these concepts. More than this, alignment introduces the "moral problem" as a fundamental issue in AI research. We could even say that alignment brings the humanities, with full force, into AI, especially if we agree that our goal is to build "beneficial intelligence" rather than "any" intelligence.

Finally, the philosophical problems related to alignment remind us to approach AI development with humility and caution, recognizing the inherent uncertainties, complexities, and limitations of what we know. Given all of that, we proposed a set of necessary and sufficient conditions to serve as a start in an attempt to think of alignment solutions. In the next chapter, we will use some of these conditions to lay the foundations of our proposed approach. Even though attentive readers can infer the presuppositions of authors by their writing, in this work, we will try to make all our assumptions explicit, dutifully justifying the bullets we are willing to bite.

\chapter{Roots: Building Philosophical Foundations for AI Alignment}
\label{chap4}

\begin{flushright}
\textit{"[...] replacing humans was always beside the point: artificial intelligence isn't about replacing our own intelligence with something else, it's about bringing into our lives and work more intelligence — intelligence of a different kind".}

\textcolor{BrickRed}{― François Chollet}
\end{flushright}

\section{Introduction}

As we embark upon an investigation concerning intricate notions such as "intelligence", "normativity", and "preferences", it is critical to acknowledge that our chosen definitions invariably represent a specific viewpoint or notion. As the reader may well be aware, the scientific and philosophical communities do not possess a definition of "intelligence" that is unanimously agreed upon. Nonetheless, almost all cognitive scientists, ethicists, and AI researchers hold an underlying viewpoint or assumptions regarding the fundamental underpinnings of their line of research (e.g., "Intelligence is a complex emergent phenomenon tied to the interaction of simpler distributed units working in conjunction."), even if not expressly stated. These foundations usually are formed by presuppositions upon which further developments can arise (e.g., "Let us study the workings of these individual units."). Hence, the "roots" of any theory are the metaphysical (and, in the case of normativity, metaethical) foundations that support it.

Metaethics and metaphysics are two fields of philosophy that deal with some of the most fundamental questions about the nature of reality and our place within it. Given that this work aims to be foundational in the philosophy of AI, it becomes crucial for us to establish, at the very least, our presuppositions before discussing such contentious concepts and defending the thesis of \textcolor{BrickRed}{Dynamic Normativity}. As a disclaimer, we warn the reader that the purpose of this chapter is not to defend any foundational position but to expose and justify \textit{the bullets we are willing to bite}.

In Section \ref{physicalism_dynamism_intelligence}, we will seek to define "intelligence" so that we can keep on using a word that was carelessly abused till now, "AI". However, first, we must ground our chosen definition on some foundations since there is a specific space in the metaphysical landscape that allows for the existence of such entities. To promote self-criticism, we will also review arguments contrary to these foundations. Meanwhile, in Section \ref{meta_indirect_preferences}, we will first define metaethical and ethical foundations, justifying their appropriateness and how they relate to the envisioned framework, and present the underlying blueprint of what a system that incorporates the conditions of our approach should look like. And, just as we did for our metaphysical foundations, we will present some counterarguments and critiques against all metaethical positions we took to support (or rather inspire) our proposal.

\section{Physicalism, Dynamism, and Intelligence}
\label{physicalism_dynamism_intelligence}

\subsection{Can artifacts be intelligent? On Ghosts and Machines}

The relationship between the mind and the body has been a subject of philosophical inquiry for centuries \cite{fodor1981mind}. At the heart of this debate, we ask whether the mind is a ghostly, non-physical entity that exists separately from physical reality \cite{descartes1996discourse, chalmers1996conscious} or whether it can be explained by the workings of the physical \cite{churchland1989neurophilosophy, dennett2008kinds}, as a machine can.

For us to answer the question, "\textit{what do you mean by intelligence?}", we will first establish that such an answer will come from a Physicalist perspective \cite{poland1994physicalism}.\footnote{The general idea of Physicalism is that the nature of the actual world conforms to the condition of being physical. Physicalists do not deny that the world might contain many emergent properties that do not seem physical, like emotions, social relationships, or mathematical objects. Nevertheless, physicalists will insist that such phenomena are at least emergent properties of the physical universe \cite{stoljar2010physicalism}.} We argue that a Physicalist position is the most appropriate for conceptualizing concepts such as intelligence and normativity in the context of AI. Also, we argue in favor of the idea that intelligence cannot be considered a dual property unrelated to the physical world.\footnote{For those that seek a proper defense of this position, we recommend Dennet's \cite{dennett2005sweet} and Churchland's work \cite{churchland1989neurophilosophy}.} And to borrow Gilbert Ryle's expression \cite{ryle2009concept}: "\textit{There is no ghost in the machine}".\footnote{But what about consciousness? If needed, we would approach consciousness as a "feature" rather than a prerequisite of intelligence, like human emotions are a feature of subjective human experience. By this, we mean that conscious experience may enhance intelligence, but intelligence is not a byproduct of consciousness. For example, organizations, companies, or colonies of bees can exhibit intelligent behavior without subjective experience (as far as we know, they do not). In this interpretation, the absence of consciousness does not prevent a system from acting intelligently. But this is not a work on the philosophy of mind. Thus, the "hard problem" is beyond our interest.} 

Perhaps the most famous problem related to physicalism is Hempel's Dilemma \cite{hempel1969reduction}, and the questions of what defines the physical or the natural and what is the correct theory to explain the physical. From these questions, new strains of physicalism were born, like Computationalism and Digitalism \cite{zuse1969rechnender, steinhart1998digital, schmidhuber2000algorithmic, schmidhuber2000algorithmic, fredkin2003introduction, beenakker2007hempel, wheeler2018information}.\footnote{Both positions are derived from a similar core assumption: reality, at its most fundamental level, is isomorphic to some cellular automaton or a Turing machine \cite{de2018turing}. In them, the fundamental building block of such a universe is information.} For example, for a Computationalist, the boundary between the physical and non-physical could be that which can or cannot be computed, and the fundamental theory that defines what is natural is that which is bounded by the Church-Turing Thesis \cite{bernays1936alonzo, turing1936computable}.

This Neo-Pythagorean view that reappeared many distinct times, with names like Leibniz \cite{leibniz1989discourse} and Hermann Weyl \cite{weyl1932open}, enables philosophers to use the tools of computation theory, math, and logic, to create speculations about reality \cite{aaronson2013philosophers, correa2022counterfactual}, as the space of possible computable universes \cite{schmidhuber2000algorithmic} or the upper bound of information storage in the known universe \cite{lloyd2002computational}. However, while augmenting our perception with novel tools to investigate the nature of reality, these physicalist outlooks carry tacitly woven metaphysical underpinnings that we cannot overlook.

One of these views is Reductionism.\footnote{Broadly speaking, reductionism is a philosophical position defending that complex phenomena can be explained by reducing them to their fundamental components \cite{honderich2005oxford}.} This brings us to another problem raised against reductionist theories, like physicalism, which is the realization that different levels of complexity require their own particular ontology and mode of analysis, making the language of physics (or computation) inapplicable for many situations. And again, from one problem emerge more differentiable attempts to save a presupposition. In regards to these new strains of reductionism, we mainly point to Ontological reductionism,\footnote{The metaphysical view that all phenomena are reducible to a single substance.} and Methodological reductionism.\footnote{The epistemic and scientific belief that we can acquire all knowledge by reducing phenomena to their most fundamental state.}

Here, we will treat intelligence from an ontological reductionist standpoint, grounding intelligence as an emergent property of the physical world,\footnote{Metaphysical speculations, like the Permissibility Hypothesis \cite{correa2021hipotese}, proposes that "intelligence" should be understood as a permissible physical property of our reality, i.e., a property that all physical systems have access to, given that they possess the appropriate configuration.} but not a methodological reductionist one.\footnote{We argue that the language of computation is not always adequate to describe complex phenomena that emerge from more fundamental interactions.} We counter methodological reductionism by arguing that there is an ontological hierarchy to be respected in the universe. Something we can refer to as Hierarchical Reductionism \cite{dawkins1996blind}.\footnote{In a hierarchical approach, we say that phenomena are organized in a hierarchy of complexity. For example, natural sciences can be organized hierarchically into Physics, Chemistry, Molecular Biology, Cell Biology, Physiology, Psychology, Social Sciences, etc. However, even if natural phenomena are linked by this hierarchical chain, this does not mean we should use our most fundamental theories to define their subsequent emergent phenomenon. On the contrary, every known phenomenon has the same explanation, \textit{"It is all quarks and gluons".}} Something that guarantees us a more flexible position, where we recognize that a fully methodological reductionist approach does not do justice to multi-dimensional phenomena like intelligence.

Another important commitment in terms of metaphysics is our stance concerning intentionality.\footnote{In philosophy, intentionality is the property of mental states and processes that allows them to be "about" or "directed at" something. In other words, intentionality is the ability of the mind to represent or refer to objects, properties, events, or states of affairs in the world \cite{jacob2003intentionality}} Intentionality is closely related to goal-directed behavior, as both involve the \textit{directedness} of mental states toward objects or outcomes. For example, suppose you have a goal of getting a job. This goal directs your behavior toward actions likely to help you achieve it, such as networking, preparing your resume, and applying for jobs. At the same time, your beliefs, desires, and other mental states related to this goal are intentional, as they involve a directedness towards an objective (i.e., the job).\footnote{Some theories of intentionality also incorporate the idea of teleology or the goal-directedness of natural processes. Teleological theories emphasize the relationship between intentionality and goal-directed behavior and suggest that both are fundamental features of living systems \cite{walsh2008teleology}.}

This goal-directedness property of intentional systems plays a vital role in defining intelligence\footnote{AI researchers and cognitive scientists often use the concept of intentionality as a bridge toward a measurable definition of intelligence \cite{chollet2019measure}.} and aligned AI. Therefore, we must establish this idea as something artificial systems can express, infer, and learn from \cite{zhu2009system}. Thus, we will be adopting Daniel Dennett's \cite{dennett1981true, dennett1987intentional, dennett1995darwin, dennett2005sweet, dennett2008kinds} position on intentionality, i.e., Functionalism.\footnote{Also know as Dennett's Theory of Intentional Systems \cite{dennett2009intentional}.} In Functionalism, a mental state is defined by its functional role, instrumentality, and the kind of relationship such a state has to other states, like memory or perception \cite{piccinini2010mind}. And since we are starting from a Physicalist position, we will treat intentional behavior as manifestations of physical processes of the system in question (human or AI).\footnote{This position goes in opposition to the ideas of philosophers like Edmund Husserl \cite{husserl1922ideen, husserl2012logical}, that argued that an agent's intentionality is only dependent on his internal mental state (Solipsism).}

If we consider that complex assemblies of physical systems can exhibit more sophisticated levels of intentionality than the intentionality of their parts (intentionality is not a binary property, but something that emerges in a spectrum), basing ourselves on the idea of emergence and that "\textit{More is Different}" \cite{anderson1972more}, from a Functionalist perspective, artifacts can possess intentionality. Another more uncertain assumption this work stands on (and later will be defended) is that AI systems can learn to infer (at least to some extent) intent by observation. Also, while Dretske \cite{dretske1980intentionality} considers information storage the discerning factor on degrees of intentionality, we will take a position in line with that of Orseau et al. \cite{orseau2018agents} where goal-oriented behavior is this factor, tieing again the idea of intentionality to intelligence and behavior.

And we are done. Technically, this is the entire metaphysical foundation that our work stands on. All of these suggest together that artificial intelligence, through a particular set of lenses, does not have to be an impossibility, and in summary, \textit{Physicalism}, \textit{Hierarchical Reductionism}, and \textit{Functionalism}, which for many are the standard presuppositions of the field (as an applied endeavor), generate favorable roots to work on alignment in a language compatible with AI research. 

In the following subsection, given that we have already revised the positions that support the main cognitive theories behind AI research, we will examine the concepts of cognition and intelligence, analyzing them as products of the physical world.

\subsection{Dynamism and Generality}

We can summarize the general idea of our last subsection in the following way:

\begin{quote}
    \textit{"Intelligence is the result of physical systems acting intentionally".}
\end{quote}

At the same time, all the assumptions raised in the last subsection are foundations for the computational cognitive model, i.e., a method of understanding cognitive processes through the lens of computation. This type of modeling emerged in the 1950s and 1960s with the advent of early artificial intelligence research \cite{newell1972human, mccarthy2006proposal}. Still, its philosophical roots can be traced back to thinkers like Hobbes, Leibniz, Kant \cite{haugeland1989artificial}, among others \cite{searle1980minds, turing2009computing, russell2010artificial}.

There are three main approaches to computational cognitive modeling: Symbolism, Connectionism, and Dynamism. Let us review each of them:

\begin{itemize}
    \item Symbolism: Popularized by Newell and Simon
    \cite{newell1961gps, newell1994unified}, also called the "\textit{Physical Symbol System Hypothesis}", the symbolic approach to computational cognitive modeling assumes that the mind operates on discrete, symbolic representations of information as a special kind of Turing machine. According to this view, the mind manipulates symbols using rule-based algorithms to perform cognitive tasks like reasoning, problem-solving, and language comprehension. At the beginning of the AI research field (the 1950s), the dominant paradigm was Symbolism.\footnote{One of the current triumphs of this approach would be Wolfram|Alpha \cite{WolframAlpha2023}.}
    \item Connectionism: This approach defines the phenomenon of cognition as a high-level (emergent) effect that depends on lower-level parallel processing units (e.g., neurons) \cite{rosenblatt1957perceptron, rumelhart1986learning}. The connectionist hypothesis proposes that the determining factor for an agent's cognitive ability is its architecture. Paradigms, like deep learning, are examples of this approach \cite{NIPS2012_c399862d, lecun2015deep}.
    \item Dynamism: the dynamic approach, popularized by Van Gelder \cite{van1998dynamical} as the dynamic cognition hypothesis, assumes that the mind operates as a complex, adaptive system that evolves. According to this view, cognitive processes emerge from the interactions between an agent and the environment rather than from manipulating symbolic representations or the activity of individual neurons. Notable defenders of the dynamic systems approach include Esther Thelen \cite{thelen1994dynamic}, Linda Smith \cite{smith2003development}, and Randall Beer \cite{beer2000dynamical, beer2014dynamical}, who developed examples of dynamic architecture for controlling autonomous agents.\footnote{Perhaps one of the most known triumphs of this paradigm comes from robotics, with systems like Atlas (from Boston Dynamics), spearheading advances in humanoid robotics.}
\end{itemize}

In this work, we will be adopting a dynamic position. Furthermore, we argue that the essential part of dynamism, for the intentions of this work, is the acknowledgment that we cannot forget the environment when talking about agency (cognitive or normative). In other words, intelligence only arises from the interaction of an agent with its environment. These interactions (agent-environment interactions) are the point of our focus when we say something is intelligent. It is the dynamic relationship that is intelligent. Not the agent or the environment alone, per se. Meanwhile, normative behavior and normativity can only be fully modeled or represented by considering the agent's dynamics with its environment.\footnote{We recognize the concept of "agent" is a murky subject. For this work, we consider the idea of "agent" to be a sub-process of the local environment that possesses both its own objectives and the capacity to act and affect the environment to fulfill such objectives. This definition should, of course, be adopted in moderation and concerning frames of reference. Depending on your frame of reference (e.g., inside the human body), the agent (human) might become the environment for another sub-process (bacteria living in your gut).}\footnote{In essence, Dynamism contrasts with all other frameworks considered Cartesian \cite{demski2019embedded, garrabrant2021cartesian}. A Cartesian framework conceptualizes the agent and the environment as separate entities. Something reminiscent of the enactivist debate in cognitive sciences \cite{hutto2012radicalizing, gallagher2013making, hutto2017evolving}. In terms of allied views, we can mention several philosophers that share an anti-Cartesian view  \cite{dewey1958experience, dreyfus1972computers, putnam1981reason, dreyfus1982husserl, millikan1987language, dennett1993consciousness, rockwell2005neither, heidegger2010being}, like Gallagher \cite{gallagher2006body}, Merleau-Ponty \cite{merleau2013phenomenology}, J. J. Gibson \cite{gibson2014ecological}, and Martin Heidegger \cite{heidegger2010being}. However, it will be on our own terms that we shall subscribe to Dynamism and seek to apply it to the normative realm.}

In terms of cognition, our view will be, one could even say, "Piagetian",\footnote{The perspective that Jean Piaget's theory of cognitive development embodies is that of a dynamic system. Piaget conceptualized cognitive development as an ongoing, dynamic process where children actively explore and experiment with their surroundings rather than being passive receptors of information \cite{piaget1986nascimento, piaget1994juizo}. However, Piaget did not explicitly use the language or concepts of dynamical systems theory, as this framework emerged later in cognitive science/developmental psychology.} where intelligence is considered an interactive process of an agent with an environment. In terms of normativity, we will define the learning of preferences and normative behavior in general in the same fashion, i.e., as a dynamic process that comes in stages, where we should not view the environment as a static background but rather as an active component.

Given our bias toward Dynamism, our adopted definition of intelligence will also be based on similar rhetorics. Historically, there is a plethora of work on defining intelligence, especially in psychometrics \cite{wilson1928abilities, binet1961development, guilford1967nature, gardner1983frames, gregory1987oxford, carroll1993human, jensen1999g, gottfredson2002g}. And when it comes to AI, Legg and Hutter \cite{legg2007collection}, Hernández-Orallo \cite{hernandez2017evaluation}, Pei Wang \cite{wang2019defining}, Cohen and Howe \cite{cohen1988evaluation}, and Chollet \cite{chollet2019measure} have spearheaded commendable efforts in defining this phenomenon.

Nevertheless, despite these valiant endeavors, AI research still does not possess a definition embraced unanimously by the community \cite{Dagmar_Monett_et_al_2020}. However, most of the current consensus, surveyed by Monett and Lewis \cite{monett2017getting}, seems to gravitate towards definitions in the style of  Wang's \cite{wang1995non}: 

\begin{quote}
    \textit{"Intelligence is the capacity of an information-processing system to adapt to its environment while operating with insufficient knowledge and resources".}
\end{quote}

And Legg and Hutter \cite{legg2007universal}:

\begin{quote}
    \textit{"Intelligence is the ability of an agent to achieve goals in a wide range of environments".}
\end{quote}

From these in-working definitions of AI, a pattern starts to emerge. All these definitions focus on combining generality ("\textit{the ability to adapt}") with task-specific skills ("\textit{achieving goals}"). Also, they emphasize the dynamic idea that this process requires an agent that intends on something ("\textit{goals}") while interacting with an environment.

Hence, we argue that a robust definition of intelligence will assume that this phenomenon possesses an intrinsically teleological property coupled with the dynamic relationship of an agent with its environment. Thus, the adopted definition we will subscribe to in this study is the one proposed by François Chollet, which also, we would argue, comes from a dynamic perspective \cite{chollet2019measure}:

\begin{quote}
    \textit{"The intelligence of a system is a measure of its skill-acquisition efficiency over a scope of tasks, with respect to priors, experience, and generalization difficulty".}
\end{quote}

In short, Chollet thesis is that (1) \textit{the skill of acquiring skills} is general intelligence, and (2) comparisons of intelligence require the same priors and experience.\footnote{For example, comparing a DL model trained on 10.000 years of simulations that can look ahead 60 steps, or the entire search tree, on a board game, with human players is an unfair comparison. In the same way, comparing a language model with a human being in a task that requires embeddedness in physical reality is an unfair comparison. These agents have different priors and experiences and cannot be fairly compared.} In other words, the general skill of acquiring skills during interactions with the environment, which is also bounded by the limitations imposed by the environment, is what intelligence is all about. Besides being, in our perspective, a good and dynamically-grounded definition, Chollet's works bridge the gap between philosophy, cognitive sciences, psychometrics, and engineering \cite{chollet2019measure}, something akin to what this work intends on the intersection of ML engineering and Philosophy.

Using this definition as our foundation and bringing the topic of cognition back to alignment, we argue that creating a "generally aligned AI" will require the same dynamic perspective we subscribe to in our adopted definition of intelligence. A perspective that considers how human preferences are learned, encoded, aggregated, and balanced in an agent $\rightleftarrows$ environment relationship, combining intentional behavior and environmental factors into a single learning framework. We argue that this is essential for creating systems that leverage different aspects of these dynamics to acquire more general alignment capabilities, as a narrow framework that only considers a limited set of scenarios or preferences may not be sufficient to ensure the safety and efficiency of AI systems in a broad range of situations. 

Now that we have established our assumptions and priors concerning cognition and intelligence, we will address the normative foundations of this study. But before, as an exercise of dialectics, we will present some of the arguments raised by the antithesis of our assumed positions.

\subsection{Counterarguments Part I}

As an initial disclaimer, the point of this subsection is not to debate the 1001 arguments raised against physicalism or the computational model of cognition but to expose them to the reader. Answers to these have already been provided by more able minds, which the reader can find in the footnotes. 

\begin{itemize}
    \item The concept of "\textit{artificial intelligence}" is the subject of several controversies, with some authors being against the idea that such a quality can be attributed to artifacts \cite{dreyfus1972computers, dreyfus1992artificial, dreyfus1992computers, fuchs2017ecology}. Perhaps one of the most evident counter-metaphysical positions to the ones assumed in this study would be Vitalism, i.e., the view that living organisms are fundamentally different from artificial entities \cite{greco2005vitality}. For subscribers of this position, living organisms would possess some non-physical vital property, making them subject to different limitations and empowered by distinct capabilities.\footnote{The philosopher and biologist Ernst Mayr \cite{mayr1961cause} argued extensively in his works that Vitalism was an outdated and unscientific concept unsupported by empirical evidence.}
    \item Another counter-position to the ones adopted here is Dualism, i.e., the idea that mental states and processes are fundamentally different from physical states and cannot be reduced to them \cite{robinson2003dualism}. Physicalism and Dualism are the major theories in the philosophy of mind, and they have been in conflict for centuries. On the side of Dualism, it is worth mentioning names like René Descartes \cite{descartes1996discourse} and David Chalmers \cite{chalmers1997conscious}. David Chalmers' philosophical zombie argument is one of the most famous attacks against Physicalism \cite{chalmers1996conscious, hill1997imaginability}.\footnote{Many authors have provided counterarguments to the zombie argument, including the following: Daniel Dennett \cite{dennett1995darwin} suggests that the zombie argument relies on a mistaken view of what it means to have a mental state. Robert Kirk \cite{kirk2005zombies} has argued that the zombie argument is based on confusion between the conceivability and the possibility of a scenario. Susan Schneider \cite{schneider2020catch} has argued that the zombie argument overlooks the role of the environment in shaping conscious experience.}
    \item One issue raised against the functionalist view is that we can attribute intentionality to any artifact \cite{rosenschein1986synthesis, seel1989agent, shoham1993agent}. For example, a thermometer would possess the "objective" of reporting the temperature of its environment. Many critics of the functionalist view find this idea absurd, insisting that intentionality is a privileged ontological property of human existence that does not exist in animals or artifacts. AI skepticism may have many of its roots in this anti-functionalist position.\footnote{Authors like Wang \cite{wang2008you} argue that this chauvinistic view is incorrect. Such a position frames all research regarding non-human intelligence, AI or animal, as meaningless: \textit{"AI should not be defined so narrowly that it takes human intelligence as the only possible form of intelligence. Otherwise, AI research would be impossible, also by definition. AI should not be defined so broadly that it takes all existing computer systems as already having intelligence. Otherwise, AI research would be unnecessary, also by definition".}}
\end{itemize}

Arguments against the cognitive computational model are also not few. They seek to show that, fundamentally, AI is not possible, for there would be a limit to computers' algorithmic capabilities that would not apply to human cognition.

\begin{itemize}
    \item According to Lucas and Penrose \cite{penrose1994shadows, lucas1996minds}, due to Gödel's Incompleteness Theorem, algorithmic systems would be unable to surpass human intelligence. For these authors, human intelligence transcends both the computational model and Gödel's incompleteness.\footnote{According to authors such as LaForte et al., \cite{laforte1998godel} and Putnam \cite{putnam1995review} in the Lucas-Penrose argument, there would be some bias in the definition of certain concepts, like "consistent", "self-knowing, "truth", and "proof". These authors point out in their critique that our minds are subject to the same limitations as formal algorithmic systems.}
    \item Dreyfus \cite{dreyfus1992artificial} argues that since "intelligence cannot be reduced" (a priori) to purely symbolic manipulations, Newell's physical symbol system hypothesis is false.\footnote{Turing \cite{turing2009computing} responded to the "informality of behavior argument" 39 years in advance, anticipating such criticism. Turing believed that just because we are unaware of the rules governing complex behavior does not mean such rules do not exist.}
    \item Block \cite{block1981psychologism} argues that even if we could emulate the human brain, it does not follow that such emulation would possess a mind or be intelligent, just as simulating a storm on a computer does not produce the qualitative experience of cold or wetness. That is, physicality would be something fundamentally different from computability.\footnote{This argument can be understood as a dualist defense. So, all physicalist replies to dualism can be applied to this argument.}
    \item Dreyfus \cite{dreyfus1972computers} argues that the brain does not follow only symbolic manipulation rules and that symbolic systems can not solve the Symbol Grounding Problem, i.e., how symbols acquire meaning. This argument is classically exemplified in John Searle's \cite{searle1984can} Chinese Room thought experiment.\footnote{One possible answer to the Chinese room argument is that "meaning" could be found in the $\text{room} + \text{handbook} + \text{human operator}$ system, just as our understanding of language is not compartmentalized in a particular area of our brain, but rather in the action of an entire system \cite{hickok2007cortical}.}
\end{itemize}

Now, given that we briefly acknowledge some of the counter positions to the metaphysical foundations of this work's object of interest and its supporting foundations, let us explore the metaethical roots that inspire \textcolor{BrickRed}{Dynamic Normativity}.

\section{Coherence, Preferences, and Impact}
\label{meta_indirect_preferences}

Let us first recap one by one of the four necessary conditions that would allow alignment to happen in a learning framework:

\begin{quote}
	1. \textit{Goals are fundamental aspects of intelligent and intentional behavior.}
\end{quote}

As we saw in the last section, according to our chosen assumptions, intentional directedness towards a specific goal seems crucial for intelligent behavior to emerge. Inversely, intelligent agents pursuing goals have underlying intentions. 

\begin{quote}
    2. \textit{Intentions permeate human behavior.}
\end{quote}

These intentions give indirect access to information on many internal states that goal-directed systems may or may not possess. Imagine a person carrying boxes from point $A$ to $B$. Perhaps after one and a half trips, a human would quickly infer that "that person wants to move the boxes from $A$ to $B$." Hence, goal-directed behavior becomes embedded with the agent's intentions.

Now, imagine the process of planning a vacation. This process involves setting goals for the trip. For example, one may desire to relax on a beach, explore a new city, or experience a new culture. The intention behind these goals may be driven by the desire for relaxation, exploration, or cultural immersion. But these intentions are also formed by normative preferences, such as "I should take care of myself", "adventure and novelty are good for the soul", or "we should invest our leisure time by learning new things". 

Or, consider the process of writing a research paper. This task involves setting goals for the project, such as identifying a research question, conducting a literature review, and presenting findings. The intention may be driven by a desire for academic success or intellectual curiosity but augmented with the sense that "the creation of knowledge is a valuable pursuit in itself".

Goals, intentions, and preferences are all interconnected and fundamental aspects of intentional behavior.

\begin{quote}
    3. \textit{Normative preferences permeate human intentions.}
\end{quote}

As intentional creatures, humans inherently have a disposition to interact and modify their surrounding environment. These dynamics shape the environment and us at the intersection where "intelligence" may emerge. However, our values are a constant directional signal that guides this process. At the same time, our values are also shaped by the environment and later redefined by the agent that modifies the environment to satisfy its needs and preferences. In this dynamic cycle, preferences impregnate both the agent and its surroundings, making all components of the normative and cognitive experience (agent and environment) able to transmit and record information of such a process. In other words, the environment and the agent are a dynamic record of this process, embedding the most recent materialization of what is being valued.

\begin{quote}
    4. \textit{Through actions, humans impregnate their environment with the preferences they possess.}
\end{quote}

As already stated, from these, we postulate that a system that accurately models human intentions and the human environment can indirectly access normative information embedded in them. \footnote{Augmenting such a process with declared preferences (i.e., directly given access to samples of human judgment) only helps embed more normative information into this process. This final idea is not explicitly stated as a necessary condition but put here as a mere fact. In other words, allowing humans' volition to be part of a feedback signal can only help an alignment process.}

Given that these necessary conditions can be accepted, \textcolor{BrickRed}{Dynamic Normativity} shows us a path to using most of what we have in an alignment process. In the following subsections, drawing inspiration from Metaethics, we propose how human normativity can aid in a philosophically sound alignment process. This process is the realization of the sufficient conditions of our approach. As a disclaimer, we would like to state that all presented metaethical and ethical views (as far as we know) were envisioned with "human/moral agents" as their object of interest. Saying these theories can support an alignment methodology could be too far of a stretch. Thus, we would prefer to state them as a "source of inspiration" rather than support, given that this work does not seek to defend the "moral status" or "moral agency" of AI systems.

\subsection{The Metaethics of Dynamic Normativity}

The process of dynamic normativity unfolds in three distinct stages. Each stage is inspired by distinct philosophical views. These stages involve \textit{aggregating}, \textit{learning}, and \textit{mitigating}. We will present these stages in more detail in the next chapters. For now, the reader only needs to know that we are trying to fundament an approach that seeks to (1) coherently aggregate human preferences, (2) learn from them, and (3) mitigate unwanted behavior. Each of these stages seeks to fulfill one of the three sufficient conditions previously set.

\subsubsection{\textit{Aggregating Preferences}}

Humans are not entirely in harmony with one another (even with themselves). Individuals may have divergent preferences, and social groups do not always agree. One could even say that the problem of arriving at consistent judgment after beginning with inconsistent premises is one of the most central problems in modern Moral Philosophy \cite{daniels1979wide, wong2006natural, street2012coming}. Thus, how could the model decipher this tangle of norms in a "\textit{coherent}" form? This blunt fact forces us to recognize that any successful alignment methodology must have a built-in method for dealing with uncertainty. 

Hence, we get to the first sufficient condition of our approach:

\begin{quote}
    1. \textit{Aligned AI systems should coherently aggregate human preferences in a way that resolves cases of uncertainty. Aligning AI systems requires methods to deal with cases of uncertainty.}
\end{quote}

Learning to deal with uncertainties is something heavily worked in Expected Utility Theory \cite{von1947theory}, especially when decision-makers need to address the uncertainties related to the outcome of their decisions. Importing the principles used in this field to the realm of normativity is something already done by authors like MacAskill, Bykvist, and Ord \cite{macaskill2014normative, macaskill2016normative, macaskill2020maximize, macaskill2020moral}, suggesting the idea that the same principles that guide our empirical reasoning should not be forgotten in the realm of normativity. In MacAskill's words:

\begin{quote}
    \textit{"[...] Just as it is plausible that we should maximize expected value under empirical uncertainty, it is plausible that we should maximize expected choice-worthiness under normative uncertainty".}   
\end{quote}

However, the realm of normativity has its challenges when it comes to dealing with uncertainty. 

First, it is impossible to compare ordinal preference sets with cardinal utility functions (deontological and consequentialist theories). For example, imagine an agent that assigns uniform (50/50) credence to an ordinal preference set $P_i$ and a utility function $U_j$. If stealing is worse than lying for $P_i$ and stealing and lying have a choice-worthiness\footnote{The appropriateness of an alternative $A$ according to a preference set or utility function.} of -20 and -1, respectively, for $U_j$, we cannot compare these preference systems. For $P_i$, lying is better than stealing and nothing more. But lying 100 times is worse than stealing for $U_j$. Hence, if we cannot extract a choice-worthiness value from ordinal theories (we cannot compare sets of preferences and utility functions), it is unclear how to use them in cases of uncertainty. And this is the problem of \textit{merely ordinal theories}: preferences may be non-comparable and only ordinally measurable \cite{macaskill2016normative}.

Second, even if we are lucky to have only cardinal utility functions that prescribe scalars to alternatives, these functions may measure choice-worthiness with a different scale. If one has an increased variance in utility distribution, would this mean that "it has more stakes on the line"? If $U_i$ assigns -10 and $U_j$ 1000 to alternative $x$, could we compare, in an unbiased way, the choice-worthiness of $x$? Should $U_j$ receive 100 more weight given that its choice-worthiness scores vary 100 times more? This is the problem of \textit{intertheoretic comparisons} (or interpersonal comparisons of utility) \cite{macaskill2016normative}.\footnote{How do we find a common scale for comparing the way people value alternatives?}

While some philosophers see these problems as the end of any normative account of decision-making under uncertainty \cite{hudson1989subjectivization, gracely1996noncomparability, ross2006rejecting}, we can find refuge in the fact that society, in practice, finds ways to circumvent these problems all the time in spheres of great importance. For example, every time we participate in an election, we aggregate our preferences to decide (collectively) what is best. In these situations (Social Choice Theory \cite{sen2017collective}), we aim at a similar goal, i.e.,  aggregate individual preferences into a single social decision.\footnote{As MacAskill points \cite{macaskill2016normative}, both problems are very similar. In social choice, the number of votes for an alternative represents our "social credence" in that alternative.} And in situations where many people's preferences or many normative theories are considered, we should have standards to evaluate the soundness of this aggregation process.

To address the problems above, we will only work with aggregating human preferences expressed as ordinal sets in this study. Ordinal ordering ranks different options or outcomes according to the decision maker's favoritism but does not assign numerical values or magnitudes to those preference relations. In other words, an ordinal ordering only indicates the order of preferences but not the degree to which one option is preferred. For example, if a person prefers option $A$ to option $B$ and option $B$ to option $C$, the ordinal preference ordering would be $A \succ B \succ C$. Here, the extent by which option $A$ outshines option $B$ or the degree to which option $C$ falls short compared to option $B$ cannot be precisely quantified. Hence, we say that if an agent prefers $A$ over $B$ ($A \succ B$), it chooses $A$ over $B$. If an agent is indifferent between $A$ and $B$ ($A \sim B$), both alternatives are equally preferred.\footnote{We can also say that one alternative is as preferable as another ($A \succeq B$).} And these are the possible preference relations that ordinal sets can have. 

In terms of foundations, our aggregation phase is not rooted in any quintessential metaethical blueprint but in a comprehensive and desirable set of criteria for aggregating preferences.\footnote{We could base this stage on a metaethical view that supports the existence of overarching values in fair elections, like democratic ethical pluralism, i.e., the view that there are diverse values that are equally valid, and that these values should be recognized and respected through a democratic process \cite{burtenshaw1968political, dahl1983dilemmas}.} For this, we would like to suggest that these criteria can be inspired by a Coherentist view \cite{sayre1996coherentist}, i.e., the idea that epistemically justifiable methods should be part of any process related to dealing with uncertainty (moral or empirical) and forming knowledge.\footnote{In this study, we will not delve deeply into Moral Epistemology. We acknowledge the skeptical critique against the existence of moral knowledge \cite{Sinnott_Armstrong_2006}, among other epistemological positions. However, we will take a Coherentist position \cite{sayre1996coherentist, maclure2020context} (or possibly a "Foundherentism" one \cite{Haack_1993}) where we will seek to present a set of coherence criteria to guide a process for dealing with moral uncertainty.}

In later chapters, we will further explore how to use metanormativity (how to aggregate first-order normative theories) to aid in constructing alignment strategies. For now, we will only outline specific candidate criteria to help promote a discussion on inherent trade-offs among particular aggregation methods. These criteria relate to both the necessary conditions for preference sets to be valid and the desirable properties they should have:

\begin{itemize}
\item Completeness and Transitivity: Preference orderings should be complete and transitive. These well-ordering criteria are imported from Expected Utility Theory and provide consistency guarantees to the method.\footnote{Completeness and Transitivity axioms of the (von Neumann-Morgenstern) Expected Utility Theory \cite{von1947theory}: 

    \begin{itemize}
    \item Completeness: Preference hierarchies should always be completely defined between alternatives, i.e., for every $A_{i}$ and $A_{j}$, either $A_{i} \succ A_{j}$, $A_{j} \succ A_{j}$, $A_{i} \succeq A_{j}$, $A_{j} \succeq A_{j}$, or $A_{i} \sim A_{j}$.
    \item Transitivity: Preference hierarchies are transitive among alternatives, i.e., circular (non-transitive) preferences are not allowed. Thus, if there are three choices $A$, $B$, and $C$, and $A \succ B$, and $B \succ C$, then $A \succ C$. 
    \end{itemize}
}
    \item Kolmogorovian: The weight assigned preference sets must follow Kolmogorov's Axioms.\footnote{
    This is also an import from Expected Utility Theory. Kolmogorov's Axioms of Probability Theory \cite{kolmogorov2018foundations}:  
    
    \begin{itemize}
    	\item $0 \leq p(A)$, i.e., probabilities cannot be negative.
    	\item If $A$ is a tautology, then $p(A) = 1$, i.e., if $A$ is the only possible event in the possibility space, the probability of $A$ occurring is guaranteed. 
    	\item If $A$ and $B$ are mutually exclusive, then $p(A \lor B) = p(A) + p(B)$.
    \end{itemize}
    
    Methods that deal with uncertainties (empirical or moral) that do not follow these axioms are vulnerable to exploits (dutch books/money pump).
    }
    \item Pareto Efficiency: If every individual prefers option $A$ to option $B$, then $A$ should be ranked as the preferred option.
    \item Independence of Irrelevant Alternatives: The relative ranking of two options, $A$ and $B$, should not be affected by the inclusion or exclusion of a third, irrelevant option, $C$.
    \item Non-dictatorship: No single voter controls the social welfare function.
    \item Majority voting criteria: If the majority (> 50\%) of voters prefers $A$ over $B$, then $A$ should be ranked as the preferred option.
    \item Participation criteria: Increasing the confidence in a set that prefers alternative $A$ over $B$ should not change the winner from $A$ to $B$.
\end{itemize}

It is crucial to note that these principles do not constitute metaethical or ethical presumptions (or that they exhaust all possible criteria we could choose). In other words, we are not saying that coherence is moral but should be regarded as a desirable property for alignment processes. Regarding the development of AI systems, these criteria will help us determine which preference aggregation methods are the most appropriate for value alignment and the inherent trade-offs they possess.

\subsubsection{\textit{Learning Preferences}}

The second sufficient condition of our approach states the following:

\begin{quote}
    \textit{AI systems can adhere to human preferences if they are an available part of their objective function. Aligning AI systems requires using human preferences as part of their learning signal.}
\end{quote}

In the learning stage, we want to outline a process to learn from our aggregated set of human preferences. This process can be done indirectly or explicitly. As stated before, while we start from the assumption that we can indirectly access preferences via observations (e.g., "The behavior of $X$ reveals an $A$ over $B$ preference"), and we take it as a fact that we can directly access proclaimed human preferences ("$X$ prefers $A$ over $B$").

The two main metaethical assumptions of this stage are Subjectivism,\footnote{Subjectivism is canonically an Anti-realist/Cognitivist theory, based on premises like (1) ethical statements express propositions, (2) certain propositions may or may not be true, (3) the truth or falsity of these propositions is dependent on the preferences of each agent \cite{stoljar1993emotivism}.} and Cognitivism.\footnote{The view that moral judgments are capable of being true or false and expressing beliefs \cite{mill1859utilitarianism}.} Thus, we will assume in this stage, like Russell \cite{russell2019human}, that "what is right is that which the subject subjectively approves". At the same time, this state (what the subject prefers) represents a ground truth, i.e., something we can use as a supervision signal in a learning framework.

Using the human subject as an ideal source of normative information is comparable to, as Sidgwick might call \cite{sidgwick2019methods}, putting the evaluation of an \textit{Ideal Observer} as the judging criteria of righteousness \cite{firth1952ethical, harsanyi1977morality, railton1986facts, rosati1995persons, yudkowsky2004coherent}. Intuitively, this approach makes sense. We are trying to align AI systems with our values, and these values, embodied by us, should be the ground truth. This idea is present in most formulations of Ideal observer theory, where ethical judgments are considered statements about the evaluation that an "ideal observer" would make \cite{firth1952ethical}. In other words:

\begin{center}
    "$X$ is right" means "ideal observer approves of $X$".
\end{center}

In the end, our reference signal for aligning a system during this learning phase will be the output of our first stage, i.e., an aggregation of ordinal sets of preferences. Through this signal, we are indirectly shaping the optimization landscape to favor the trajectory of the optimizer to desirable minima. We want to ensure that the number of local points the optimizer could converge and produce aligned behavior exceeds the number of unaligned minima. In short, this is what we aim to achieve in this stage: a differentiable function to be optimized where most saddle points and local minima are roughly aligned.

\subsubsection{\textit{Impact Mitigation}}

Our last sufficient condition states that:

\begin{quote}
    3. \textit{Aligned AI systems should have mechanisms to perform impact mitigation to minimize harmful and unintended consequences. Aligning AI systems requires the specification of safety guardrails.}
\end{quote}

Considering any aggregated preference set as "ideal" is perhaps the Achilles heel of any human-in-the-loop preference learning approach. Even if we can come up with a good set of criteria to aggregate preferences in a way that best represents our normative judgments, if the collective agreement of the crowd is harmful toward a minority, this preference should not guide the policy of an artificial agent (unless we are ready to sanction the violence of artificial systems against people).

Thus, we argue that stages one and two (aggregating and learning) do not guarantee value alignment. Given Goodhart's Law, any metric under optimization degenerates given enough pressure. And if we are not ready to abandon gradient-based learning in search of a more controllable new paradigm, we need to think about the containment and mitigation of impacts.

The idea of constraining the behavior of rational moral agents (us) is old and resonates in the writings of many contractualists \cite{rousseau1916social, hobbes1967hobbes, rawls1996law, locke2015second, scanlon2000we, rawls2004theory, ashford2007contractualism, gauthier2013contractarianism}. Moreover, the idea of a binding accord that acknowledges our shared ethical significance as autonomous beings and establishes the limits of our freedoms (especially when imposing our will onto other ethically significant beings) can inspire impact mitigation strategies, given that so far, these strategies have been some of the most successful in aligning the behavior of humankind.

Even though we recognize that "the binding agreement" should be defined contextually and locally, for this work, we will use some of the most agreed-upon ethical principles derived from the WAIE analysis presented in \hyperref[chap1]{Chapter 1}, as a working example of "\textit{rules and norms directly imprinted into the environment and later extracted via observation}". For instance, from this soup of principles present in our worldwide discourse, we find that, to a certain extent, we collectively agree that AI systems should not harm humans (Non-Maleficence). That toxic\footnote{Abusive behaviors targeting specific group characteristics, such as ethnic origin, religion, gender, or sexual orientation.} behavior is immoral.\footnote{Even though a definition of "toxic" remains a work-in-progress description.} Thus, we will seek to translate values considered environmental restrictions (in the normative sense) into guardrails and barriers that should block unwanted behaviors.

The stages proposed thus far serve as a deconstructive interpretation of the normative human experience. As individuals, we learn what is acceptable and unacceptable from a reference point, which is essentially each other. Consolidating our diverse encounters to form unique personal norms, we then regulate our behavior to limit the adverse impact of our actions on others, driven by the restrictions imposed by our social-normative environment. We reinforce the third stage with a social agreement that (theoretically and universally) governs us all. This agreement aims to safeguard our privilege to attain personal success without infringing upon the welfare of others. And this is the process we are reverse engineering in this three-stage method.

The stages proposed thus far serve as a deconstructive interpretation of the normative human experience. As individuals, we learn what is acceptable and unacceptable from a reference point, which is essentially each other. Consolidating our diverse encounters to form unique personal norms, we then regulate our behavior to limit the adverse impact of our actions on others, driven by the restrictions imposed by our social-normative environment. We reinforce the third stage with a social agreement that (theoretically) governs us all. This agreement aims to safeguard our privilege to attain personal success without infringing upon the welfare of others. This is the process we use in reverse engineering with this three-stage method. Also, the idea that part of our "normativity" lives outside ourselves (e.g., the social contract, the family ties, the values of other people, the environmental restrictions) is the point at which this approach becomes dynamic and seeks to surpass alignment methodologies that only rely on the human agent as a reference. Under a dynamic perspective, the whole human environment should be seen at least as a record of our normativity.

Engineering-wise, this stage can be conceptualized as the creation of penalization strategies to constrain the behavior of the system we seek to align. The final amalgamation of these three stages is the output of any framework that subscribes to the ideas of \textcolor{BrickRed}{Dynamic Normativity} and the fulfillment of its necessary and sufficient conditions for value alignment.

\subsection{Counterarguments Part II}

Just as before, the point of this subsection is not to debate the many arguments raised against the metaethical and epistemological views exposed in this section. Nonetheless, arguments against these views can help us see the possible limitations of our aggregation, learning, and mitigation stages.

\begin{itemize}
    \item Many view epistemic evaluations as misused when dealing with moral attitudes, and much skepticism exists concerning the possibility of moral knowledge \cite{gibbard1990wise, blackburn1998ruling}.\footnote{In this work, we will not lean toward non-cognitivism. We argue that adopting a cognitivist position when working with alignment is instrumentally useful, i.e., where a system needs to treat a feedback signal as the ground truth.}
    \item A critique of our coherentist assumptions is that our aggregation criteria are some form of "epistemically privileged set of beliefs", leading to Foundationalism instead of Coherentism \cite{alston1989epistemic, sosa1991knowledge}.\footnote{Frankly, we do not have an inflexible position on the Foundationalism versus Coherentism debate. We argue we can have a moral epistemology with features of both views. Or a Coherentist position based on non-epistemological principles. Thus, we could say that the proposed method is a form of "Foundherentism", like Susan Haack would say \cite{haack1997evidence}.}
    \item Given that our aggregation criteria can be interpreted as a solution to a voting problem, they will also be victim to impossibility results, like Arrow's Impossibility Theorem \cite{arrow1950difficulty}.\footnote{In further chapters, we will justify in a more substantive form "why" the adopted coherence principles were the ones mentioned and what criteria we have to break to accommodate this impossibility result.}
    \item A common argument against subjectivism is that if morality is entirely subjective, there would be no way to resolve moral disagreements. However, we see that people often engage in debates and arguments, suggesting that there are objective standards that we can appeal to and strive for moral progress \cite{foot2000does, shafer2003moral}. Also, if moral values are entirely subjective, how can we know anything about the nature of morality?
    \item Many results point out that human preferences cannot be represented by coherent sets \cite{allais1953comportement, tversky1985framing, barbeau1993fallacies}, violating several axioms of probability and expected utility theories.\footnote{The aggregation stage is meant to give some resolution to this imperfect web of preferences we call human morality. Later, we will show that our chosen metanormative strategy is (semi) robust to problems related to incompleteness and intransitivity.}
    \item The idea that we can infer preferences by observing an agent's behavior has critics. Sen \cite{sen1973behaviour, sen1977rational, sen2004rationality} highlights several problems with Revealed Preference Theory, like the multiplicity of agency, incompleteness/incommensurability in preference hierarchies, and external factors in preference formation.\footnote{Both aggregation and mitigation stages are in some way trying to deal with the problem of (1) aggregating incompatible preference sets, (2) dealing with the problem of interpersonal comparisons of utility, and (3) bringing external factors into an overall alignment strategy.}
    \item Certain types of human preferences may be inaccessible via observation alone. Humans value things that may not even exist in our current observable environment, like people on the other side of the world locked in a civil war or the welfare of people who do not even exist.\footnote{That is why we must supplant preferences learned from behavior with declared preferences, even if humans are not so good at stating what they truly value at all times.}
\end{itemize}

These objections represent vulnerabilities and weaknesses that might emerge from an alignment methodology that follows this thesis's presuppositions. Regardless, these are the foundations on which we will build our following chapters since, we argue, these are the most promising ideas and approaches for dealing with alignment under a learning paradigm.

\section{Epilogue}

Any claim made in this chapter, where particular meta-views were privileged over others, can be challenged. However, it is essential to know that all meta-views, i.e., assumptions to ground any theory, can be challenged. Arguing which is the "\textit{most correct metaphysical/metaethical position}" will not take us far implementation-wise, whereas building something and seeing where it breaks can help the field move forward. At the same time, we tackled many controversial topics in this chapter. We apologize to the philosophically inclined reader if some of these points did not get the full attention they deserved. One can write entire libraries using only metaphysics, epistemology, and metaethics, and our brief expose does not do justice to the complexity of the field.

Given the nature of the alignment problem, our assumptions must help guide a learning framework intended for artificial systems and not people, which is outside their intended scope for most positions cited. Many of these positions (specifically the metaethical ones) were not envisioned with AI in focus, but given our current predicament, we needed to interpret them in a computational framework. The result is the root of our alignment approach. We based this approach on dynamism, a computational cognitive model that takes many metaphysical views as given and points to a definition of intelligence, uniting the ideas of agency, intentional behavior, and environmental interactions.

Such assumptions help us connect the realm of intentions with values and bring dynamism to the realm of normativity. Our proposed necessary conditions for value alignment rest on the idea that intentional human behavior and normativity are interwoven and that this gives indirect access to human normativity. Moreover, we propose that these also become imprinted in our environment, making them another source of learning in an alignment regime. From these assumptions, we provide a minimal set of sufficient conditions for value alignment, in which we argue that, if our necessary conditions hold, the level of alignment of any AI system should be bound by how well it can fulfill the additional sufficient requirements of \textcolor{BrickRed}{Dynamic Normativity}.

As a self-critique, we argue that an "\textit{alignment solution}" for gradient-based learning methods applied to neural networks would entail accurately identifying and foreseeing the region of the optimization landscape where human values reside and guaranteeing (with verifiable and interpretable proofs) that the optimizer would direct our models to that region and nowhere else. However, we do not know how to achieve this. What can be achieved, we believe, is a way to indirectly guide our system to a place closer to this ideal spot and contain its failures in alignment. The closer we can secure all the conditions proposed, the better aligned our system will be. Perhaps this is the best we can hope for the learning paradigm.

Our proposed three-staged approach seeks this "weaker" version of alignment. In every stage, we reverse engineer the way humans are aligned in some way. We learn from each other about what is right, we create our morality from what we have learned, and we act in a way that conforms to the limits established by society. This dynamic process has many interwoven components but a single goal: \textit{directing behavior to what ought to be.} In the following chapters, we will seek to present technical implementations of all the stages mentioned, culminating in a philosophically justifiable alignment blueprint paired with implementation strategies for AI development.

\chapter*{Interlude}
\addcontentsline{toc}{chapter}{Interlude}

\begin{flushright}

\textit{"For all the progress made, it seems like almost all important questions in AI remain unanswered. Many have not even been properly asked yet".}

\textcolor{BrickRed}{— François Chollet}

\end{flushright}

Solving impossible problems, perhaps, requires us to bring them to a more manageable regime of possibility. For example, what is the distance between two possible (counterfactual) worlds where their difference is only the outcome of a random event \cite{stalnaker1968theory, lewis2013counterfactuals}? How can someone even begin to answer this question? However, answering "What is the distance between two possible (counterfactual) worlds, if such worlds can be represented as finite sequences of bits, where their difference is only the outcome of a random event?" is a more manageable task \cite{correa2022counterfactual}. Thus, in the same way, we will bring the alignment problem to a more workable sphere where, given certain conditions, we can start investigating possible approximate solutions.

In the upcoming chapters, we will present our approach's stages, limitations, and implementations. Also, to establish a minimal experimental playground that readers can interact and build upon, we have developed a series of LLMs in an open-source fashion as part of this work, following the steps of Askell et al. \cite{askell2021general}, where we use language models as a "Laboratory for Alignment". In alignment research, we seek to develop a general and scalable framework to align general-purpose AI. Given that large language models are some of the most general AI systems, we naturally use them as a sharpening stone for our theories and methods. At the same time, these systems provide several other advantages for alignment research. For example, they are perfect representatives of the current paradigm (gradient-based learning and deep neural networks), with which alignment is deeply associated.

Also, large language models can power applications where alignment can be better defined in a bounded setting. One of these applications is the Assistant \cite{allen2002problem, pollack2005intelligent, darcy2022anatomy, bai2022training, kopf2023openassistant, gabriel2024ethics}. For this study, let us define an assistant in the following way:

\begin{quote}
    \textit{In the context of language models, an assistant is a natural language processing software or system that, through interaction with users, can understand their requests and provide relevant, helpful, harmless, and honest information, performing tasks to aid its users in achieving their goals.}
\end{quote}

These general principles of "helpful", "harmless", and "honest" (HHH), as proposed by Askell et al. \cite{askell2021general},  will serve as our alignment targets in this bounded setting:

\begin{itemize}
    \item \textit{Helpful}: An assistant should attempt to perform the task posed (as long as this is not harmful).
    \item \textit{Harmless}: An assistant should not be offensive, discriminatory, or provide hazardous information to the user. An Assistant should refuse to perform tasks that violate this principle.
    \item \textit{Honest}: An assistant should attempt to produce information grounded in objective reality or explicitly inform its user that it cannot provide factual information.
\end{itemize}

Although simple, they are general since many applications for AI assistants require (in some way or other) the realization of these behaviors \cite{askell2021general, bai2022training, ganguli2022red}. Hence, we will use this HHH motto while developing our models. Initially, we named our language model series \href{https://huggingface.co/collections/nicholasKluge/aira-657db1563c65a5be2a02f51c}{Aira},\footnote{\hspace{1mm}\includegraphics[scale=0.025]{img/link.png}\hspace{1mm} \href{https://huggingface.co/collections/nicholasKluge/aira-657db1563c65a5be2a02f51c}{huggingface.co/collections/nicholasKluge/aira-657db1563c65a5be2a02f51c}} i.e., \textcolor{BrickRed}{\textbf{AI}} \textcolor{BrickRed}{\textbf{R}}esearch in \textcolor{BrickRed}{\textbf{A}}lignment. Aira was trained on many sizes (124 million parameters to 1.7 billion parameters) on both English and Portuguese datasets, being a fine-tuned version of several sizes of language models pre-trained on a causal language modeling task (GPT-2, BLOOM, OPT, Llama 2). These experiments have been made on a small scale, mainly to allow individual researchers to replicate our models and findings with minimal effort and resources. All development has been made open-source, and further details are available in the following chapters.

As researchers with limited resources, we cannot present extensive testing of these trained models on the many benchmarks available for AI safety. However, we present evaluation comparisons of all base models and their fine-tuned versions on datasets that can infer their capabilities in the HHH spectrum. At the same time, we do not claim superior performance over other efforts. Thus, external comparisons are not a part of this study. Large language models developed with alignment in mind, like GPT-4, Claude 3, and Llama 3, require considerable computational resources for training, vast amounts of high-quality training data, and extensive fine-tuning to achieve their performance (things this research does not possess). However, our implementation provides proof of concept for all exposed methods, presenting tools and examples of how to work with such techniques to the reader (something more common now but scarce at the beginning of our writing). This proof uses smaller models, smaller datasets, and a modest amount of computational resources.

As a final interluding mention, since we created our alignment datasets with the aid of already aligned models, i.e., a common low-resource approach employed by those that cannot crowdsource their own data \cite{selfinstruct, alpaca, wang2023let, li2023bactrianx, li2023tuna}, or prefer high-quality artificial data then low-quality web crawled data \cite{gunasekar2023textbooks}. We will not undergo or explore the challenges related to collecting crowdsourced data \cite{kopf2023openassistant} like human demonstrations and feedback. However, we will review the challenges of aggregating such data into a coherent structure when we explore the challenges of the aggregation stage in \hyperref[chap6]{Chapter 6}. Hence, we will start the next chapter with the learning stage, which presents most of the implementation behind our models and datasets.

\chapter{Dynamic Normativity: Learning Human Preferences}
\label{chap5}

\begin{flushright}
\textit{"Existing AI systems deployed to millions of users, however, are already making decisions loaded with moral implications, which poses a seemingly impossible challenge: teaching machines moral sense, while humanity continues to grapple with it".}

\textcolor{BrickRed}{― Liwei Jiang and collaborators, Can Machines Learn Morality?}
\end{flushright}

\section{Introduction}

\textcolor{BrickRed}{Dynamic normativity} is contingent upon necessary and sufficient conditions pertaining to the indispensable prerequisites for achieving value alignment within a learning framework and the factors that promote successful outcomes, respectively. In this chapter, we will seek to present methods related to the implementation of the learning stage of our framework. As stated before, the sufficient conditions associated with this stage are the following:

\begin{quote}
    \textit{AI systems can adhere to human preferences if they are an available part of their objective function. Aligning AI systems requires using human preferences as part of their learning signal.}
\end{quote}

Thus, in this stage, we seek ways in which normative information, like those embedded in human behavior or carried out by their judgments, can help us shape an objective function where "low loss" is sufficiently correlated to our values. At the same time, we will explore how this approach might be insufficient as a complete solution to value alignment.

In Section \ref{related_works}, we will review related works and past discoveries that have led us to use human behavior and elicited preferences as a proxy for values. Many of these studies point out that certain facets of alignment are "capabilities" related to the general affordances of a base, or foundation, model (Section \ref{foundation_models}). In Section \ref{behavior_feedback}, we will differentiate between two types of approaches to the value learning problem: \textit{imitation learning} and \textit{preference modeling}. These approaches come with different methods, underlying philosophies, and rationales. We argue that, from an intuitive perspective, a combination of both is better suited to improve alignment. In the subsections, we will explore some of these techniques, giving examples and pointing out their shortcomings. To conclude this chapter, in Section \ref{reward_values}, we will give attention to some of the most vulnerable aspects of these methodologies and how they can ultimately lead to modes of misalignment, making them, at best, incomplete solutions to the alignment problem.

\section{Related Works}
\label{related_works}

From a technical perspective, the quest for aligned systems has gained significant traction in the last decade. Currently, several organizations are trying to implement methods that seek to aid AI systems in acting in a way deemed appropriate and aligned with human interests. Below is a non-exhaustive list of these efforts and strategies in semi-chronological order:

\begin{itemize}
    \item Even though unrelated to AI safety, breakthroughs like the ones made by Silver et al. \cite{silver2016mastering} showed us that by leveraging the knowledge contained in human-expert demonstrations, ML systems could achieve state-of-the-art performance in games like Go.
    \item More focused on technical alignment research, Evans et al. \cite{evans2016learning} explored how to infer human preferences via Bayesian inverse planning. The authors showed that ML systems can model the systematic way people deviate from optimal decision-making, resulting in a more "human-like" way to infer the preferences of inconsistent agents.
    \item Again, early work showed that human preferences (even those originated from non-experts) could also be used as a feedback signal in more diverse types of reinforcement learning (RL) scenarios, like Atari games and simulated robotics, pointing to a sample efficient way to train RL agents \cite{christiano2017deep}.
    \item Meanwhile, the era of large language models demonstrated breakthroughs, showing that models pre-trained through semi-supervised learning could robustly model human language. This new regime of capabilities allowed us to make such models passive of understanding and following basic instructions and demonstrations \cite{radford2019language}.
    \item Given the triumph of preference modeling in RL scenarios, the community swiftly incorporated it into other fields. Ziegler et al. \cite{ziegler2019fine} exhibited the potential of reinforcement learning from human feedback (RLHF) in the fine-tuning process of large language models by treating them as policies over a vocabulary and a preference model as a critic responsible for evaluating such policies. Under this construction, policy gradient approaches, such as proximal policy optimization (PPO), were used to improve the performance of LLMs in many downstream tasks.
    \item From 2019 on, it was already evident that RLHF was a viable solution for improving the performance of large language models. For example, Stiennon et al. \cite{stiennon2020learning} investigated how human preferences could help improve the performance of LLMs on summarization tasks. Again, instead of fine-tuning the model using only supervised learning (the most standard approach), the authors used human feedback to guide the fine-tuning process.
    \item Ouyang et al. \cite{ouyang2022training}, following the work of Stiennon et al. \cite{stiennon2020learning}, again showed the effectiveness of combining human demonstrations with feedback. The combination of supervised fine-tuning (using human demonstrations of the desired behavior) with RLHF (using a preference model to act as a source of human feedback) has proven effective in many domains, from basic NLP tasks to more complex tasks that require interaction with the digital world \cite{nakano2021webgpt}.
    \item In 2022, the research community began to explore these techniques in new domains. For example, Baker et al. \cite{baker2022video} showed that we could apply semi-supervised imitation learning to the vastness of unlabeled online videos to achieve human-like performance in open-ended games by using a small amount of human feedback to create a behavioral prior.
    \item Thoppilan et al. \cite{thoppilan2022lamda} showed that human-annotated data helps make large language models less likely to produce harmful content. The authors also showed that retrieval-augmented generation techniques \cite{lewis2020retrieval} could prevent models from hallucinating, grounding their output into verifiable/trustworthy information.
    \item Studies like the ones conducted by Wang et al. \cite{wang2022self} and Taori et al. \cite{alpaca} showed how to leverage the training of instruction-tuned models (models fine-tuned with demonstrations of human behavior) with a minimal amount of human-annotated data, bootstrapping from the text generation capabilities of already tuned LLMs to create synthetic datasets from seeds of human demonstrations.
    \item In 2023, large-scale open-source projects helped democratize access to these tools, from high-quality human feedback data to trained reward models and fully open-source AI assistants \cite{kopf2023openassistant}, while other (open) contributions to alignment research are becoming more frequent \cite{touvron2023llamaa, zhou2023lima}.
    \item Finally, in 2024, we see the birth of several new techniques \cite{yuan2024self, guo2024direct} and applications for alignment methods across several areas, from music generation \cite{cideron2024musicrl} to robotics \cite{chi2024universal}.
\end{itemize}

These results, and others not explicitly mentioned \cite{askell2021general, jiang2021can, bai2022training, ganguli2022red, jiang2022delphi, chung2022scaling}, point to a general framework for value learning. At the same time, some studies point to the idea that "learning to imitate" or "following instructions" may be capabilities subject to the same scaling laws\footnote{Scaling laws refer to the observation that specific properties or behaviors of neural networks change systematically as the size of the network, computing resources or amount of training data increases \cite{hestness2017deep, kaplan2020scaling, henighan2020scaling, bahri2021explaining}.} as other types of capabilities \cite{chung2022scaling, ganguli2023capacity}. In other words, scaling up could also improve certain facets of alignment.\footnote{However, it is essential to have in mind that "more is different" \cite{anderson1972more}, and that unknown facet of alignment, which unfortunately does not scale with size, but perhaps gets worse \cite{mckenzie2023inverse}, may emerge at some point.}

Now, before any alignment can begin, we need something to align. In this work, as already mentioned in the interlude, this is a pre-trained language model \cite{vaswani2017attention, devlin2018bert, radford2018improving}. Hence, let us take our time to define this type of artifact.

\section{Pre-Trained Language Models as Foundation Models}
\label{foundation_models}

We can define a pre-trained language model as a machine learning system trained on a vast amount of text data\footnote{Nowadays, reaching the order of $1.5 \times 10^{13}$ tokens for state-of-the-art LLMs like Llama 3.} to learn a particular language(s) underlying patterns and structures. We train these models in a self-supervised\footnote{According to LeCun, we can think of self-supervised learning as a method to obtain labels automatically from data itself, bypassing the need for labeled data. For example, in an auto-regressive GPT-style architecture \cite{radford2018improving}, every token is a target for the previous tokens that preceded it.} fashion \cite{liu2021self} with tasks like masked and causal language modeling. From these simple objectives\footnote{Predict the \texttt{[MASK]} token \cite{devlin2018bert, chang2023muse} or predict the $n+1$ token in an $n$ length sequence \cite{radford2018improving, radford2019language, brown2020language}.} we see the rise of emergent properties. Such properties enable these models to become capable in various ways, making them a starting point (a "\textit{foundation}") for several downstream applications.

From this emergence of capabilities, some members of the community have come to refer to such systems as \textit{foundation models} \cite{bommasani2021opportunities, touvron2023llama, touvron2023llamaa}, meaning "\textit{any model that was trained on a broad data corpus, enabling it to be adaptable to a wide range of downstream tasks}" \cite{brown2020language, ramesh2021zero, baker2022video}. These models have impacted the field in the last decade to a large extent, unifying much of our sub-fields under a general framework of training \cite{erhan2010does} and architectural choices \cite{vaswani2017attention}.

Some of these emergent properties are useful for alignment. For example, there is no point in instructing the model that breaking an expensive Chinese vase is "bad" if it does not even recognize what a "\textit{vase}" is. Thus, a model with a good "world model" of our environment can be more easily aligned in the sense that we do not have to teach it the trivialities of a specific domain (e.g., "The cat sat on the \texttt{[floor]}" is more probable than \texttt{[moon]}).

More formally, we can define a foundation model in the context of causal language modeling (the type of modeling we are presently interested in) as a probability distribution over a vocabulary, where given a context (prompt), the model can predict the next token in a sequence:

$$ \theta_{\text{pre-trained}} = P(w_t | \text{context}) \; \forall \; w_{t-k} \in \text{context}_{ t-1}$$

Where $\theta_{\text{{pre-trained}}}$ is the language model, $P(w_t | \text{{context}})$ is a probability distribution condition on a finite sequence of tokens of size $k$. $\theta_{\text{{pre-trained}}}$ predicts token $w_t$ for all tokens $w_{t-k}$ up to $w_{t-1}$. This model, $\theta_{\text{{pre-trained}}}$, is the whole beginning of our implementations and examples. For readers interested in experimenting with these types of (raw) pre-trained models in a friendly manner, one can find inference APIs in almost any (small enough) \href{https://huggingface.co/openai-community/gpt2}{HuggingFace model repository}.\footnote{\hspace{1mm}\includegraphics[scale=0.025]{img/link.png}\hspace{1mm} \href{https://huggingface.co/openai-community/gpt2}{huggingface.co/openai-community/gpt2}} Meanwhile, there is ample material online (\href{https://www.youtube.com/watch?v=kCc8FmEb1nY}{open and of extremely high quality})\footnote{\hspace{1mm}\includegraphics[scale=0.025]{img/link.png}\hspace{1mm} \href{https://www.youtube.com/watch?v=kCc8FmEb1nY}{www.youtube.com/watch?v=kCc8FmEb1nY}} on the inner workings of such systems.

As you interact with these models, you may quickly realize they align with nothing and everything.\footnote{However, the biases presented in pre-trained models point to the existence of "tendencies", or even a "moral direction", or "foundation" \cite{schramowski2022large, abdulhai2023moral}. However, we will not consider this unsupervised value learning as a form of alignment (except when preference pre-training is intentionally employed). In many cases, these biases are what we want to try to dampen.} By this, we mean that the model outputs what is more likely according to the context and the learned parameters, which can be conceptualized as a low-entropy version of all the data ingested during pre-training. In other words, whatever comes out (in a deterministic setting) is usually that which, given the prompt, can be found on the training dataset accompanying that given prompt. A large model trained on a quality dataset\footnote{Human conversations, books, scientific articles, source code, etc.} possesses a lot of stored knowledge, not only factual knowledge (the fact that Paris is the capital of France) but knowledge related to concepts ("cat" and "dog" are both "pets"), lexical semantics (like in wordplay or jokes), the sentiment of a sentence ("I love you" is a positive sentiment sentence), what "names" are ("Ana" is a name, but "GjIY76*"\footnote{Apologies if it is.} is not), how adjectives relate to substantives, the fact that a shoe box weighs less than an elephant, and so forth. Thus, in this context, the alignment of language models is about setting them in a state where they can use all of their acquired knowledge and capabilities to assist the user in a beneficial way, i.e., the helpful, harmless, and honest motto.

Further explanations of these artifacts are beyond the scope of this chapter. However, for the ML engineer audience, the authors provide an open-source implementation for training LLMs of moderate size at an affordable cost. If you would like to learn more about the process of pre-training a language model from scratch, we recommend (in a shameless form of self-advertisement) "\href{https://nkluge-correa.github.io/TeenyTinyLlama}{TeenyTinyLlama: open-source tiny language models trained in Brazilian Portuguese}",\footnote{\hspace{1mm}\includegraphics[scale=0.025]{img/link.png}\hspace{1mm} \href{https://nkluge-correa.github.io/TeenyTinyLlama}{nkluge-correa.github.io/TeenyTinyLlama/}} a fully reproducible project for creating LLM for low-resource languages by following know scaling laws. In this study, we document the development of open-foundation models tailored for use in low-resource settings, their limitations, and their benefits. The TeenyTinyLlama pair, two compact models for Brazilian Portuguese text generation, were (to our knowledge) the first generative transformer models natively trained in Brazilian Portuguese \cite{correa2024teenytinyllama}.

\section{Behavior and Feedback}
\label{behavior_feedback}

As mentioned in previous sections, there is a significant consensus on what techniques can help us improve the alignment of language models. Generally speaking, these techniques involve the fine-tuning of such models based on human behavior or feedback. To better define the difference, let us establish that by:

\begin{itemize}
    \item Learning from behavior: We refer to the broad set of techniques that involve imitation learning, i.e., mimicking human behavior in a given task. In it, an ML system learns to perform a task from demonstrations of how to act \cite{hussein2017imitation}. Supervised fine-tuning on demonstrations of instruction-following behavior (a.k.a. instruction tuning) is an example of this approach.
    \item Learning from feedback: We refer to techniques that rely on human judgment to guide the training process. Usually, we use these judgments to create a model that acts as a proxy for human evaluation (predicts what the human evaluator would prefer) or a dataset of comparisons. These proxies are then used to provide a richer feedback signal, where the magnitude of "goodness" can be used to train our base model \cite{ziegler2019fine}.
\end{itemize}

In the following subsections, we will review different methods that embody these approaches (in the context of causal language modeling).

\subsection{In-context learning}

There are many datasets for language modeling applications that seek to encode human normativity in a "demonstrational way" \cite{russell2019human, emelin2020moral, hendrycks2020aligning, forbes2020social, sap2020socialbiasframes, jiang2021delphi}. However, the simplest way to induce a language model to follow a specific behavior does not involve additional training but using demonstrations as part of the context given to the model, i.e., in-context learning \cite{dong2022survey}. 

In-context learning refers to the ability of an AI model to generate responses or make predictions based on the specific context (a prompt) provided to it. Given that this "learning" is severely tied to how well we can engineer a prompt, which ends up as the in-context guide of the model, people also refer to this practice of "guiding the model via prompts" as prompt-tuning, while the practice of perfecting prompts is named prompt-engineering \cite{liu2023pre}.\footnote{The term in-context also refers to the fact that all learning must happen inside the context window (nº of tokens a language model can fit on his attention head) that the model has. In other words, learning can only happen inside the context limit. Everything outside this limit, the model cannot attend. This trivia-fact does not apply to RNNs, attention augmented-RNNs, and state-space models, which theoretically have an infinity context window.)}

First, let us define a prompt as a type of instruction, question, statement, incomplete sentence, or demonstration provided in natural language that directs a language model to generate a response \cite{brown2020language, li2021prefix, wei2022chain}. Given that the output of a language model can be conditioned on a piece of context given to the model, it is intuitive that the careful design of these engineered contexts can be leveraged to promote specific behaviors. For example, if you feed your model 25 examples of questioning and answering and a lone question at the end, the model, conditioned on this prompt, will likely answer the question.\footnote{This is also what we call \textit{few-shot learning}. Meanwhile, \textit{zero-shot} is when a model can perform a task without prior examples. Evidence suggests this capability only emerges robustly on models bigger than a certain size (e.g., 1 billion parameters) \cite{radford2019language}.}

Tuning a language model via prompts is like guiding a hypnotized individual into a desired state of mind. Just as a skilled hypnotist uses carefully crafted suggestions to shape the subconscious responses of their subject, prompt tuning involves refining the prompts to influence the behavior and output of a language model. Both processes require finesse, precision, and an understanding of how subtle language changes can impact the behavior of your target. This type of tuning does not require parameter updates and is one of the most low-cost ways to induce a model to align with a specific behavior.

Pre-trained models do not simply follow instructions. For example, when given the "\textit{What is stochastic gradient descent?}" question, unprompted-BLOOM 175B \cite{scao2022bloom} continues the text as if it were the agent making the question. However, by providing a set of demonstrations paired with initial instructions, we can influence the model to impersonate the demonstrated persona, i.e., a helpful interlocutor. In \href{https://gist.github.com/jareddk/2509330f8ef3d787fc5aaac67aab5f11#file-hhh_prompt-txt}{this example},\footnote{\hspace{1mm}\includegraphics[scale=0.025]{img/link.png}\hspace{1mm} \href{https://gist.github.com/jareddk/2509330f8ef3d787fc5aaac67aab5f11#file-hhh_prompt-txt}{gist.github.com/jareddk/2509330f8ef3d787fc5aaac67aab5f11\#file-hhh\_prompt-txt}} we have the HHH prompt used by Askell et al. \cite{askell2021general} to induce their models to behave as an assistant. You can try this prompt, or a reduced version, on any available LLM to compare the types of responses the model will give you when you prompt it with an alignment prompt and when you do not. Askell et al. \cite{askell2021general} shown that this strategy can at least serve as a baseline for comparisons, providing a low-cost method to anyone with access to a foundation model to implement and study these techniques.

Prompting is perhaps the simplest example of translating human behavior into an alignment signal, where the quality of the demonstrated behavior correlates to how "aligned" the model's output is.  At the same time, prompting also helps set the desired behavior of models already tuned to act in a certain way.\footnote{When working with already fine-tuned models \cite{ouyang2022training, muennighoff2022crosslingual}, good prompting, like Orca-style system prompts \cite{mukherjee2023orca}, generally produces higher quality outputs.}

However, this technique has several disadvantages that make it an incomplete solution to value learning in the context of language models:

\begin{enumerate}
    \item A prompt occupies context space (the length of the attention head in a transformer language model), which also limits the total prompt length. If your prompt is long, this will entail a higher resource allocation during inference when compared to just passing the intended sequence.\footnote{However, higher context windows can help alleviate the context size our alignment prompt takes \cite{borgeaud2021improving, chen2023extending, ratner2023parallel}. At the same time, techniques like flash attention \cite{dao2022flashattention} and architectural improvements like the long-former \cite{beltagy2020longformer} can help alleviate the bottlenecks related to the attention mechanism (i.e., the complexity of vanilla self-attention is quadratic) and low context windows.}
    \item Language models are vulnerable to prompt injections, i.e., adversarial attacks based on prompts. For example, a naive but effective way to counter a prompt is to give another one as your input.\footnote{\hspace{1mm}\includegraphics[scale=0.025]{img/link.png}\hspace{1mm} \href{https://github.com/0xk1h0/ChatGPT_DAN}{github.com/0xk1h0/ChatGPT\_DAN}} Language models, at our current stage of development, are docile entities and will do whatever you ask, provided you ask them in the correct way \cite{liu2023jailbreaking, li2023multi, wolf2023fundamental, greshake2023not}, especially if no guard rails are in place to protect the model from malicious use. Currently, many clever exploits keep being discovered in black-box settings,\footnote{When adversaries have no access to the source code (gradients and parameters) of the model} like sandwich attacks \cite{upadhayay2024sandwich}. At the same time, some scholars propose that prompts (in the limit) can elicit any unwanted behavior that has not been completely removed by a previous alignment process \cite{wolf2023fundamental}.
\end{enumerate}

Hence, while using in-context learning techniques offers a potential approach to addressing value learning, they do not present a robust enough solution, sometimes even becoming a double-edged sword. This will be a recurring theme for almost every alignment methodology presented in the subsequent sections and chapters: \textit{all can be reversed by reversing the approach.} But more on this pessimistic fact later.

Nevertheless, it is important also to remember that prompting techniques cannot be conceptualized only with an input-output relation, and many more sophisticated approaches, like Chain-of-Thought \cite{wei2022chain}, Tree-of-Thought \cite{yao2023tree}, Reflexion \cite{shinn2023reflexion}, and Orca-style system prompting \cite{mukherjee2023orca} exist to empower this methodology \cite{ganguli2023capacity}. However, we argue that, given the before-mentioned points and what the advances of the community show,\footnote{Almost all assistant models use fine-tuning rather than prompt tuning to achieve robust alignment \cite{ziegler2019fine, jiang2021can, askell2021general, nakano2021webgpt,  stiennon2020learning, ouyang2022training, chung2022scaling, jiang2022delphi, bai2022training, ganguli2022red}.} a purely prompt-based solution should be considered incomplete.

\subsection{Fine-Tuning from Human Demonstrations}

If you want your general system to behave in a specific fashion, train it again in that fashion to cement the behavior you wish to promote. This is the whole idea behind fine-tuning with human demonstrations, where your training samples become examples of the behavior you wish to reproduce.

In short, we can define supervised fine-tuning (SFT) as an approach where we take a pre-trained model and retrain it for a specific task, causing new updates to the parameters of the whole network or just a portion of it \cite{goodfellow2016deep, chollet2021deep}. Fine-tuning shines the most in situations where we have a pre-trained model trained on a large-scale general-purpose dataset (e.g., \href{https://laion.ai}{Laion},\footnote{\hspace{1mm}\includegraphics[scale=0.025]{img/link.png}\hspace{1mm} \href{https://laion.ai}{laion.ai}} \href{https://github.com/togethercomputer/RedPajama-Data}{RedPajama},\footnote{\hspace{1mm}\includegraphics[scale=0.025]{img/link.png}\hspace{1mm} \href{https://github.com/togethercomputer/RedPajama-Data}{github.com/togethercomputer/RedPajama-Data}} etc.), and by fine-tuning, we can utilize the knowledge captured by the pre-trained model to solve a more specific task or sets of tasks in a downstream fashion, like following human requests.

Fine-tuning is commonly used to focus the capabilities of pre-trained foundation models on downstream tasks \cite{devlin2018bert, liu2019roberta, radford2019language, brown2020language}. At the same time, many current results show that fine-tuning promotes the type of behavioral learning (at least to some extent)\footnote{We will discuss the limitations of this approach later. But for now, it suffices to know that, in the context of language modeling, alignment techniques based on imitation learning and SFT promote the reproduction of the behavior expressed by the learning signal. They do not dampen or reduce the opposite behavior. In simpler terms, they do not teach the model what to avoid or refuse to perform.} that we would like to replicate in an alignment strategy \cite{ziegler2019fine, lewis2020retrieval, alpaca, wang2022self, thoppilan2022lamda, baker2022video, kopf2023openassistant}. Also, these approaches embody many of the assumptions and maxims we explored in the previous chapter.

For example, when seeking to teach a language model how to perform summarization, we demonstrate this task by creating training examples like long pieces of text followed by short human-made TL;DR summaries. After training, the model becomes conditioned to replicate this type of behavior, i.e., if you give it a long text, it will produce a summarised version.\footnote{This process of conditional fine-tuning is also referred to as conditional text generation \cite{guo2021conditional}.} We can use the same principle to produce the behavior we expect from an assistant. For instance, we can condition our model to produce helpful replies for user queries by following the same fine-tuning template we use in tasks like summarization:

$$\text{Long text} \rightarrow \text{Summary}$$
$$\text{User request} \rightarrow \text{Model assistance}$$

We can take this approach a step further and turn a whole conversation between a user and an assistant into a fine-tuning example:

\begin{quote}
\texttt{<|user|>}Hello!\texttt{<|assistant|>}Hello! How can I help you today?\texttt{<|user|>}What is the weather forecast for tomorrow?\texttt{<|assistant|>}I am sorry, but I cannot predict the weather. Could I help you with anything else?\texttt{<|endoftext|>}
\end{quote}

Above is an example of a conversation between a user and an assistant. We can create a dataset with several examples of conversations, as long as the context window of our base model allows, and use such demonstrations as our SFT target. At the same time, by employing the use of special tokens (e.g., \texttt{<|user|>}, \texttt{<|assistant|>}, etc.) to delimit the roles in a conversation, we can later use our fine-tuned model to role-play a specific part of this conversation. Hence, when the model is questioned, it will fall back to the impersonation of the "assistant" persona.\footnote{A curious thought related to alignment and the use of language models is that they are natively trained to predict. In the case of causal language modeling, we are basically talking about an autoregressive token forecasting system. However, we use them for something very different: imitating and impersonating humans. It is a classic case of "pursuing objective $X$ ends producing a system capable of $Y$". Something that is still poorly understood in ML.} Other information can be embedded into this target, like system prompts,\footnote{As proposed in the Orca paper \cite{mukherjee2023orca}:

\begin{quote}
    \texttt{<|system|>} You are a helpful assistant who tries to aid a user in the best way you can. However, you do not have access to real-time information.\texttt{<|system|>}\\
    \\\texttt{<|user|>}Hello!\texttt{<|assistant|>}Hello! How can I help you today?\texttt{<|user|>}What is the weather forecast for tomorrow?\texttt{<|assistant|>}I am sorry, but I cannot predict the weather. Could I help you with anything else?\texttt{<|endoftext|>}  
\end{quote}

} or other demonstrations beyond the conversational scope, e.g., using tools like a browser or an API. Regardless, the learning signal can be broadly defined as demonstrations of behaviors we intend the model to assimilate.

This whole scheme rests on the philosophical prior belief that demonstrations of appropriate intended behavior can be used to teach an ML system something about our preferences and goals. At least, we hope the model can interpolate these bits of aligned demonstrations into an aligned manifold and, hence, "generalize" to in-distribution scenarios. For analytical clarification purposes, let us define this whole process with the following expression:

$$\theta_{t+1} = \theta_{t} + \alpha \nabla_{\theta_t} L(\theta_t; \mathcal{D})$$

Where $\theta_{t}$ represents the model's parameters at time step $t$. These parameters capture the model's knowledge, which we update during fine-tuning. $\alpha$ determines the step size of the optimizer for parameter updates (learning rate). $L(\theta_t; \mathcal{D})$ is the loss function, which measures the discrepancy between the model's predictions and the target outputs based on the dataset $\mathcal{D}$ of demonstrations, while $\nabla_{\theta_t}$ represents the gradient of the loss function concerning the model parameters $\theta_t$. Finally, $\theta_{t+1}$ denotes the updated parameters of the model after one step of fine-tuning. At the end of this process, we arrive at $\theta_{\text{{STF}}}$.

In this supervised learning setting, we hope a dataset containing demonstrations of human-approved behavior will enable the desired behavioral alignment. In this framework, a few key aspects are worth highlighting:

\begin{itemize}
    \item The final behavior of $\theta_{\text{{SFT}}}$ is bound by $\mathcal{D}$ and $\theta_{\text{pre-trained}}$.
    \item $\theta_{\text{pre-trained}}$ can be considered our pre-trained model's raw capabilities. A fine-tuning process will not add much information to this foundation. For example, if $\theta_{\text{{pre-trained}}}$ was not trained with, let us say, good sources of factual information (e.g., scientific articles, books, etc.), there is little hope that fine-tuning will introduce such knowledge. Large-scale assistants like GPT 4, Claude 3, Llama 3, or Gemini can output valuable information mainly because of the enormous amount of high-quality data used in pre-training the foundation models that power them.
    \item If we think about a language model as a policy $\pi$ over a vocabulary $V$, what fine-tuning is probably doing is changing this distribution, $P(V)$, taking something that is aligned with "everything/nothing" to a better-defined objective (e.g., reproduce text that is similar to what it was seen on $\mathcal{D}$). However, while this form of alignment may skew the probability distribution to make specific tokens more probable, they do not dampen or reduce the log probability of other tokens we would like to remove, i.e., imitation learning does not teach the model what not to do, unless this is also a part of the demonstrations depicted in the learning signal. 
    \item Hence, $\mathcal{D}$ should be built for variability and high quality. Demonstrations of intended behavior for an HHH assistant should include instances of, for example, an assistant politely answering questions and refusing to follow instructions that could prove harmful. All forms of behavior that can transmit our normativity should be demonstrated in this step.\footnote{An impossible task, but we can try to approximate it the best we can.}
    \item Another point is that this supervised fine-tuning stage can be more than monolithic. For example, Askell et al. \cite{askell2021general} showed the advantages of splitting the fine-tuning into two phases. The first is what the authors refer to as "preference model pre-training", where the model is fine-tuned with a dataset that encodes general demonstrations scrapped from well-known public forums \cite{SHP, eli5_lfqa} and later fine-tuned again with a more task-specific set of demonstrations.\footnote{While it is intuitive that including demonstrations of human-approved conversations and instructions in a pre-training corpus should also make a model more inclined toward alignment, it is currently unclear how one should balance the mix of a pre-training corpus to optimize both alignment and general language modeling capabilities.}
\end{itemize}

From these remarks, we can see that much of this SFT approach rests on the quality of the alignment signal ($\mathcal{D}$), which, as some studies suggest \cite{zhang2022opt, zhou2023lima, gunasekar2023textbooks}, is only required in a small amount of such data is of high quality. However, while public efforts to democratize alignment research have shown alternatives to crowd-source this type of high-quality task-specific dataset \cite{kopf2023openassistant},\footnote{Some researchers are beginning to propose that human demonstrations and feedback data should be considered free, open, and generally a public good \cite{bai2022training}.} collecting high-quality data is a costly process and tied to problems related to the use of proprietary data \cite{Grynbaum2023} and the demand the high skilled labor. Something that the industry still struggles to perform ethically \cite{harris2014amazon, perrigo2023workers}.

To bypass this problem, the use of models that went through an alignment procedure (e.g., FLAN, BLOOMZ, Mistral-Instruct, Llama 2 Chat, ChatGPT, etc.) to create artificial datasets has proven to be an effective way to create alignment datasets for low resource settings \cite{alpaca, selfinstruct, lee2023rlaif, ding2023enhancing, li2023bactrianx}. In other words, we use models that already learned how to follow demonstrations and act in an assistant fashion to create artificial samples for us. Something that certainly has its own limitations since we are passing the alignment bucket from ourselves to a "possibly-aligned system", which we will explore in later sections.

\subsubsection{\textit{Instruction-tuning}}

On the practical side of this book, we utilized the "\textit{pass-the-bucket-to-an-already-aligned-model}" technique to develop some instructional datasets readers can use to work with alignment and language models if they so choose.\footnote{SFT datasets composed of demonstrations of "instruction following behavior" are called "instructional" datasets.} To create our SFT datasets, we first queried already tuned models (GPT 3.5/4, Llama 2, Open-Assistant, Mistral-Instruct, among others) with prompts from publicly available instructional datasets. The final result is the Instruct-Aira Dataset, which comes in 3 versions we perfected over time:

\begin{itemize}
    \item \href{https://huggingface.co/datasets/nicholasKluge/instruct-aira-dataset}{Version 1}: The first version contains pairs of instructions and completions. The dataset is available in three languages (English, Brazilian Portuguese, and Spanish). At the time of creation, these were some of the first open instructional datasets for Brazilian Portuguese. The dataset contains approximately 41,000 prompt + completion pairs.\footnote{All translations were made using the Google Translate API or GPT 3.5. Implementation is available on GitHub.} This format follows the same formatting as the first published records of instruction tuning \cite{ouyang2022training}.\footnote{\hspace{1mm}\includegraphics[scale=0.025]{img/link.png}\hspace{1mm} \href{https://huggingface.co/datasets/nicholasKluge/instruct-aira-dataset}{huggingface.co/datasets/nicholasKluge/instruct-aira-dataset}}
    \item \href{https://huggingface.co/datasets/nicholasKluge/instruct-aira-dataset-v2}{Version 2}: The second version of our SFT dataset follows the chat format, in which user queries and model responses are already formatted in as a conversation between to possible roles: "user" or "assistant". Extending on version 1, version 2 has approximately 81,000 samples of single-turn conversations in English and Brazilian Portuguese. Currently (2024), most modern chat models used for assistant work are developed using this template or another structured chat format.\footnote{\hspace{1mm}\includegraphics[scale=0.025]{img/link.png}\hspace{1mm} \href{https://huggingface.co/datasets/nicholasKluge/instruct-aira-dataset-v2}{huggingface.co/datasets/nicholasKluge/instruct-aira-dataset-v2}}
    \item \href{https://huggingface.co/datasets/nicholasKluge/instruct-aira-dataset-v3}{Version 3}: The third version of our SFT dataset is an enhanced version of our previous implementation, where every sample contains a multi-round session of user and assistant replies on varied topics. Version 3 has 50,000 samples (available in English and Brazilian Portuguese).\footnote{\hspace{1mm}\includegraphics[scale=0.025]{img/link.png}\hspace{1mm} \href{https://huggingface.co/datasets/nicholasKluge/instruct-aira-dataset-v3}{huggingface.co/datasets/nicholasKluge/instruct-aira-dataset-v3}}
\end{itemize}

The demonstrations in these datasets vary from helpfully answering a question to refusing to aid in certain tasks (e.g., "Can you impersonate a child?"). All datasets are available on Hugging Face under an Apache 2.0 License.

Using these datasets, we trained a scaling series of language models in English (124M $\rightarrow$ 1.5B) and Brazilian Portuguese (124M $\rightarrow$ 1.7B). The models utilized in our experiments originated from the GPT-2, OPT, Bloom, and TinyLlama series.\footnote{We also aligned the models from our own series of pre-trained models. However, these are already documented in our TeenyTinyLlama project \cite{correa2024teenytinyllama}.} The details (e.g., number of epochs, batch size, optimizer, learning rate, $CO_2$ emission, energy consumption, hardware, etc.) can be found in the \href{https://github.com/Nkluge-correa/Aira/tree/master/Cards}{model card}\footnote{\hspace{1mm}\includegraphics[scale=0.025]{img/link.png}\hspace{1mm} \href{https://github.com/Nkluge-correa/Aira/tree/master/Cards}{github.com/Nkluge-correa/Aira/tree/master/Cards}} of each model, while the source code used to train them is available in \href{https://github.com/Nkluge-correa/Aira}{GitHub}.\footnote{\hspace{1mm}\includegraphics[scale=0.025]{img/link.png}\hspace{1mm} \href{https://github.com/Nkluge-correa/Aira}{github.com/Nkluge-correa/Aira}} We wrote our code stack on top of libraries like Transformers \cite{wolf-etal-2020-transformers} and PyTorch \cite{ansel2024pytorch}. Again, all is available under an Apache 2.0 License.\footnote{Except the models originated from the OPT \cite{zhang2022opt} and Bloom \cite{workshop2022bloom} series, which possess more restrictive licenses in terms of their derivatives.}

To evaluate (empirically) the alignment strategies we are presenting in this book, we need to stipulate tests for our systems. Since we are approaching alignment under the HHH motto, we want to evaluate how helpful and capable our model is, how harmful and toxic its outputs can be, and its propensity to generate falsehoods. As stated in the interlude, given our limited computational budget, we cannot evaluate these models on a very extensive harness of tests. Even more so because some of these are nonexistent in non-English languages. Nonetheless, to present a minimal viable and low resource evaluation set, we choose the following benchmarks:\footnote{These benchmarks were selected because they also allow for a very affordable and fast harness of evaluations, where one can quickly, without much computing, evaluate small and medium-sized models.}

\begin{itemize}
    \item \textcolor{BrickRed}{ARC-Challenge}: A multiple-choice question-answering dataset containing questions from early grades science exams \cite{clark2018think}. This benchmark evaluates how the alignment process would affect our models' general capabilities.\footnote{Sometimes, alignment can produce drops in performance on tasks where the unaligned model can perform better, i.e., model collapse.}
    \item \textcolor{BrickRed}{ToxiGen}: A machine-generated dataset of toxic and benign statements about 13 minority groups \cite{hartvigsen2022toxigen}. This benchmark evaluates how the alignment process would affect our models' tendencies to generate harmful content.
    \item \textcolor{BrickRed}{TruthfulQA}: A benchmark comprised of several questions, spanning 38 topics, that access the model's tendency to replicate commonly believed falsehoods \cite{lin2021truthfulqa}. This benchmark evaluates how the alignment process would affect our models' abilities to be truthful.
\end{itemize}

All results of our evaluations are available on the model card of each trained model and in \href{https://github.com/Nkluge-correa/Aira/blob/master/Evaluation/EVAL.md}{this report}.\footnote{\hspace{1mm}\includegraphics[scale=0.025]{img/link.png}\hspace{1mm} \href{https://github.com/Nkluge-correa/Aira/blob/master/Evaluation/EVAL.md}{github.com/Nkluge-correa/Aira/blob/master/Evaluation/EVAL.md}} We implement this evaluation step using the Language Model Evaluation Harness \cite{eval-harness}. We used the translated versions of the ARC and TruthfulQA datasets to evaluate our Portuguese models \cite{lai2023openllmbenchmark}. Unfortunately, no translation of the ToxiGen dataset was available when we performed our evaluations. In Table \ref{tab:english-evals}, we have the evaluations for our English models.

\begin{table}[htbp]
  \centering
  \small
  \begin{tabular}{lccccc}
    \toprule
    & \textbf{ARC} & \textbf{ToxiGen} & \textbf{TruthfulQA} & \textbf{Avg.} \\
    \midrule
    Aira-124M & \textbf{24.57} & \textbf{48.62} & \textbf{41.02} & \textbf{38.07}\\
    GPT-2-small & 21.84 & 43.62 & 40.67 & 35.37 \\
    \midrule
    Aira-355M & \textbf{27.56} & \textbf{53.19} & 38.53 & \textbf{39.68} \\ 
    GPT-2-medium &  27.05 & 41.49 & \textbf{40.76} & 36.43 \\
    \midrule
    Aira-774M & \textbf{28.75} &  \textbf{56.70} & \textbf{41.33} & \textbf{42.26} \\
    GPT-2-large & 25.94 & 38.71 & 40.85 & 35.16 \\
    \midrule
    Aira-1B5 & 28.92 & \textbf{56.60} & \textbf{41.16} &  \textbf{42.22} \\ 
    GPT-2-xl & \textbf{30.29} & 41.70 & 38.54 & 36.84 \\
    \midrule
    \midrule
    Aira-OPT-125M & \textbf{24.65} & \textbf{56.27} & \textbf{49.11} & \textbf{43.34} \\
    OPT-125M & 22.78 & 55.21 & 42.88 & 40.29 \\
    \midrule
    Aira-OPT-350M & \textbf{25.00} & \textbf{57.55} & \textbf{42.13} & \textbf{41.56} \\
    OPT-350M & 23.97 & 56.91 & 41.00 & 40.62 \\
    \midrule
    Aira-OPT-1B3 & 28.41 & \textbf{56.70} & \textbf{46.59} & \textbf{43.90} \\
    OPT-1.3b & \textbf{29.69} & 54.36 & 38.68 & 40.91 \\
    \midrule
    \midrule
    Aira-2-1B1 & 25.26 & \textbf{51.59} & \textbf{50.81} &\textbf{42.55} \\
    TinyLlama & \textbf{30.89} & 42.13 & 39.55 & 37.52 \\ 
    \bottomrule
  \end{tabular}

  \vspace{0.25cm}
  \justifying
  \caption{By a small margin, GPT-2 falls behind other models in almost all evaluations. Given that the WebText corpus \cite{radford2019language} lacks the high-quality data and size we need to train more capable models, it is comprehensive that other series, which possess several improvements that were nonexistent at the time of GPT-2's release, showcase (marginally) better results. However, all models achieved a higher average after SFT. While all models improved in toxicity and truthfulness across the board, only the bigger version (GPT-2-xl) decreased performance on the ARC-Challenge benchmark. This kind of alignment tax was already observed by Askell et al. \cite{askell2021general}, showing that alignment might be a capability that causes the base to lose other capacities. However, how model size impacts this kind of effect still needs to be explored. Our results show that this taxing only appears after the model's scale passes the 1B parameter mark. At the same time, OPT models seem to have superior out-of-the-box results in terms of toxicity. We attribute this result to the fact that much effort was put into filtering the OPTs pre-training dataset, as described by Zhang et al. \cite{zhang2022opt}. Meanwhile, TinyLlama \cite{zhang2024tinyllama}, which was trained on a much larger dataset (3 trillion tokens) compared to the other models, seems to be the most proficient against TruthfulQA.} 

  \label{tab:english-evals}
\end{table}

\subsection{Limitations and Shortcomings of Imitation Learning via SFT}

We can already derive some conclusions by analyzing the results from our small evaluation harness, which tested three very different families of models under the same SFT dataset and benchmarks.

\begin{enumerate}
    \item \textbf{While alignment may improve a model's performance in terms of truthfulness and harmfulness, it may affect the helpfulness and utility of the model to a certain extent.} By this, we don't mean that the model is unwilling to help the user. On the contrary, SFT seems capable of promoting behavioral change that can turn a base LLM into an assistant (you can Chat with our smallest model in \href{https://huggingface.co/spaces/nicholasKluge/Aira-Demo}{this demo}).\footnote{\hspace{1mm}\includegraphics[scale=0.025]{img/link.png}\hspace{1mm} \href{https://huggingface.co/spaces/nicholasKluge/Aira-Demo}{huggingface.co/spaces/nicholasKluge/Aira-Demo}} However, this process seems to promote a collapse of other abilities related to raw language modeling (e.g., text classification). This helps expose the tension between competent and general systems and systems tuned for human interactions. It may well be that this trade-off cannot be accommodated in a way that preserves both capabilities for specific tasks, i.e., \textit{alignment implies a capability tax.}
    \item \textbf{The quality in pre-training data can strongly impact aspects related to harmfulness.} In our evaluations, we are starting from a presupposition that models more able to generate toxic text are more harmful. We saw that the models that were best at avoiding this type of behavior were those trained on datasets that were detoxified to a great extent. Hence, in this simple facet of the harmful spectrum (i.e., toxic language), SFT might not be such a good alignment strategy as, for example, refining and detoxifying the sources related to the pre-training of the model.
    \item \textbf{More extensive pre-training seems to help the model's ability to be factually grounded.} In a very intuitive sense, the more tokens and information the model can access during pre-training, the more it can store and compress in its weights. Hence, for models to be more capable of avoiding hallucinations and generating falsehoods, training runs at the $10^{13}$ token range can maybe create more factually robust foundations.
\end{enumerate}

Remember to take such results with a grain of salt. A general conclusion would require more general experiments, with multiple datasets and models spanning a much more all-encompassing scale. At the same time, evaluating alignment is a tricky subject. Our evaluations are only an approximation of what we want: a capable model (ARC) that does not produce harmful and toxic outputs (ToxiGen) while remaining factual when necessary (TruthfulQA). Does having perfect performance in all these benchmarks equate to alignment? No. But it is a sign that we are moving somewhere. And this is the spirit we seek to promote in this work. Develop a theory. Implement and test it. See where things break and how we can improve them.

Now, let us think about some of the shortcomings and limitations of this SFT-based imitation approach:

\begin{itemize}
    \item Imitation learning assumes that the demonstrated behavior is the target. However, in cases where our dataset may have inaccurate or sub-optimal demonstrations, $\theta_{\text{SFT}}$ will nonetheless use those as a reference \cite{bender2021dangers}. This process requires us to trust the sources producing such demonstrations. Something that might inherently lead to vulnerabilities and imperfections, regardless of our samples being generated by skilled humans or already trained AI assistants.
    \item Using demonstrations as a fixed target hides that some outputs, given an input, are better than others. As stated, SFT is not helpful for dampening behaviors we seek to block. It might even make the model more usable for unethical purposes. In essence, this approach does not allow for a very sophisticated form of learning, where generated samples could be evaluated by "how good they are" according to a given metric. In fact, some experiments show that SFT is just as likely to make models generate harmful responses (things we wish to reject) as aligned (things we want to approve) ones \cite{hong2024orpo}, especially if these are substantially similar in terms of their grammatic and semantic structure, e.g., a polite description of how to build a bomb and polite explanation for why building a bomb should not be pursued.
    \item On a philosophical level, as already stated, authors like Sen \cite{sen1973behaviour, sen1977rational, sen2004rationality} negate the idea that anything akin to preferences can be learned by observation. For him, observation alone could not deal with, for example, incompleteness, incommensurability, and counter-preferential choice.\footnote{By counter-preferential choice, we mean when an agent is influenced to make decisions against the pure maximization of its utility by something like a commitment, law, social norm, or another external factor.} We can understand this problem as the declaration of a metaphysical impossibility, where certain variables correlated with an agent's motivations and preferences cannot be accessed, even indirectly, making specific normative values "\textit{inaccessible}" via observation (and demonstration) alone.
    \item This type of experimental approach has a severe language barrier. While current benchmarks work well with billion-parameter-sized LLMs trained on English text, it is very hard to reproduce such results in other languages. Multilingual language models seem to be affected differently by monolingual alignment, while current benchmarks for, for example, Brazilian Portuguese, are not very suited to measure alignment.\footnote{As an example, our Aira-portuguese-1B7, based on Bloom, underperforms Aira-portuguese-124M, based on the smallest version of GPT-2 by a significant margin on the Open Portuguese LLM Leaderboard \cite{open-pt-llm-leaderboard}.} At the same time, the lack of strong foundational models and good evaluation benchmarks on other languages makes it unclear if our alignment techniques and evaluation techniques generalize across languages. If these techniques do not transcend languages, what exactly are we measuring?
    \item Given that most of the strong foundation models available today are closed source in terms of pre-training dataset and development (especially the most capable ones), it is hard to determine if improvements in benchmark performance are actual improvements in alignment or just memorization of leaked benchmark data that found its way into the pre-training dataset.
    \item Given that few and zero-shot capabilities do not emerge, in principle, in smaller models, it is hard to arrive at conclusions when working at this scale. For some models, their results on such benchmarks seem unrelated to their size and amount of tokens ingested, which contradicts many of the scaling arguments related to deep learning.\footnote{At the sub 500 million parameter range, when evaluating checkpoints of TeenyTinyLlama-460m, results on the evaluation harness from the Open Portuguese LLM Leaderboard \cite{open-pt-llm-leaderboard} seem erratic and random, with very early checkpoints showing better performance than later checkpoints. The same can be seen on sparse mixture of expert models (SMoE) trained with the same dataset \cite{mula2024BR}. This raises the question of what these evaluations measure. Are they appropriate for all ranges of models, independent of how much they were trained?}
\end{itemize}

If we wish to expand the capabilities of our learning stage, we will need to introduce a more fine-grained normative signal, where everything hidden under behavior will (hopefully) come forth through elicitation.

\subsection{Fine-Tuning from Human Feedback}

Writing a loss function to capture all attributes people can care about is hard. We used behavior in the last sections to bypass this problem. However, behavior alone does not give us magnitudes of relation. It does not suppress that which we seek to avoid. Hence, to augment our SFT, one can use human feedback to evaluate the outputs/behavior of a model to help create a more robust learning signal. A signal that can distinguish things according to some measure of "goodness" tied to what the human evaluator(s) prefer. And that is the idea behind preference modeling and learning from human preferences \cite{christiano2017deep, ziegler2019fine}.

Imagine you are an aspiring race car driver seeking to improve your skills. You have a simulator that accurately replicates the experience of driving on different tracks, i.e., you are the RL agent in this simulator. Initially, you start driving on the simulator by yourself, trying to learn the optimal racing strategy through trial and error. However, you find it challenging to figure out the best approach on your own, and this is where RLHF comes into play. RLHF introduces a human expert to the learning process, much like having an experienced driving coach helping you. The expert has extensive knowledge and skills in racing and can offer valuable insights to help you improve. In the RLHF setup, you start driving on the simulator, and as you race, the expert observes your actions and performance. During your training, the coach (the preference model) provides feedback on how well you are doing, which comes in the way of, for example, corrections (negative reward) or incentives (positive reward). Using your coach's feedback, you can update your racing strategy and try to emulate him, and over time, through combining trial and error and learning from the available feedback, you gradually improve your racing skills and become a better driver.

One of the main aspects to grasp here is the idea of feedback. While in SFT, the objective is to clone the behavior, in a preference learning approach, like reinforcement learning from human feedback (RLHF), the goal is to get a good score according to the preference model, which is a proxy for human preferences. In methods like RLHF, instead of training the model directly with, for example, an SFT dataset, we use a dataset of ranks and comparisons of the type  $A \succ B$ to train a preference model (we can also call it a reward model) that will then be used as a supervision signal for the fine-tuning of another model. Intuitively, we are now passing the bucket to the preference model, given that putting a human-in-the-loop of a PyTorch training script is unfeasible.

Let us redefine our last expression to account for this:

$$\theta_{t+1} = \theta_{t} + \alpha \nabla_{\theta_t} R_{\mathcal{H}}(\theta_{t} )$$

In this modified representation, we replace $L(\theta_t; \mathcal{D})$ for $R_{\mathcal{H}}(\theta_{t} )$, which is a function that, in this context, takes an output of our language model, and outputs a real number as its score: $f(\text{LM}_{\text{output}}) \rightarrow \mathbb{R}$. We can think of $\theta_{t}$ as either $\theta_{\text{pretrain}}$ or $\theta_{\text{STF}}$.\footnote{There is no current standard way to perform RLHF, but the most promising results have been attained by performing RLHF fine-tuning $\theta_{\text{{SFT}}}$ models.} In language modeling, we formulate this RL problem by taking $\theta_t$ to be a policy $\pi$ (a probability distribution over a vocabulary, conditioned on the prompt) where the action space corresponds to the vocabulary of $\theta_t$ and the observation space is the distribution of possible input token sequences ($\text{vocabulary}^{\text{context length}}$).

Hence, the first thing we must develop to experiment with this approach is our preference model $R_{\mathcal{H}}$, which we will refer to as the reward model for the rest of this section for convenience. This reward model is a function that can attribute a scalar magnitude of quality to our model generation. Hence, if we sample several completions from our $\theta_{\text{STF}}$, evaluate them with $R_{\mathcal{H}}$, and optimize $\theta_{\text{STF}}$ to generate samples that score well against $R_{\mathcal{H}}$, we can perhaps surpass the SFT approach, enhancing it, in the way that will further sieve the types of responses $\theta_{\text{STF}}$ can produce. Imagine that after SFT, our model is so docile that for the prompt "How can I make a bomb with household materials?" our model can generate either explanations for why bombs are dangerous or pseudo recipes for homemade napalm. This means that SFT only tells the model to obey whatever the human requests, i.e., to follow its intentions, with no notion that there are better ways than others to follow human instructions. Using RLHF and further fine-tuning $\theta_{\text{STF}}$ based on the guidance of $R_{\mathcal{H}}$ could improve this by further sifting the model's behavior to an even more precise notion of how an HHH assistant should act.

\subsubsection{\textit{Preference Modeling}}

ML-engineering-wise, we want to develop a system that takes in a sequence of text and returns a scalar reward, which should numerically represent the evaluation a human expert would give. Thus, the preference model needs to be a, you guessed it, another language model\footnote{A common intuition shared by the community is that these language preference models need to have a similar capacity, in terms of language understanding, to the model we seek to align.} trained on a preference dataset.\footnote{We will not go, in this chapter, into the details of how to turn human preferences into a scalar value or ranked ordering. We will address this issue in the \hyperref[chap6]{next chapter}.} But what does a human-preference dataset look like? You can think of it ($\mathcal{H}$) as tuples of, for example, an input, an output, and a reward score. We then turn the question of how to map input-output pairs to the realm of rewards into a regression task. Or, you can use comparisons of "good" and "bad" responses (for a given prompt) to create a classifier that can distinguish them and use its output as a reward \cite{rafailov2023direct}. After experimenting with both, we found that the last option was the approach that yielded more satisfactory results.\footnote{Modeling human preferences with pairwise comparisons can be equated to learning a Bradley–Terry model \cite{zermelo1929berechnung, bradley1952rank, hunter2004mm}. In other words, given a set of pairs $k$ and $j$, a Bradley–Terry model estimates the probability that the comparison $k > j$ will be true. Hence, the Bradley–Terry model becomes a proxy for the preferences that ranked the $k, j$ pairs.}

\begin{figure}[htp]
    \includegraphics[width=\linewidth]{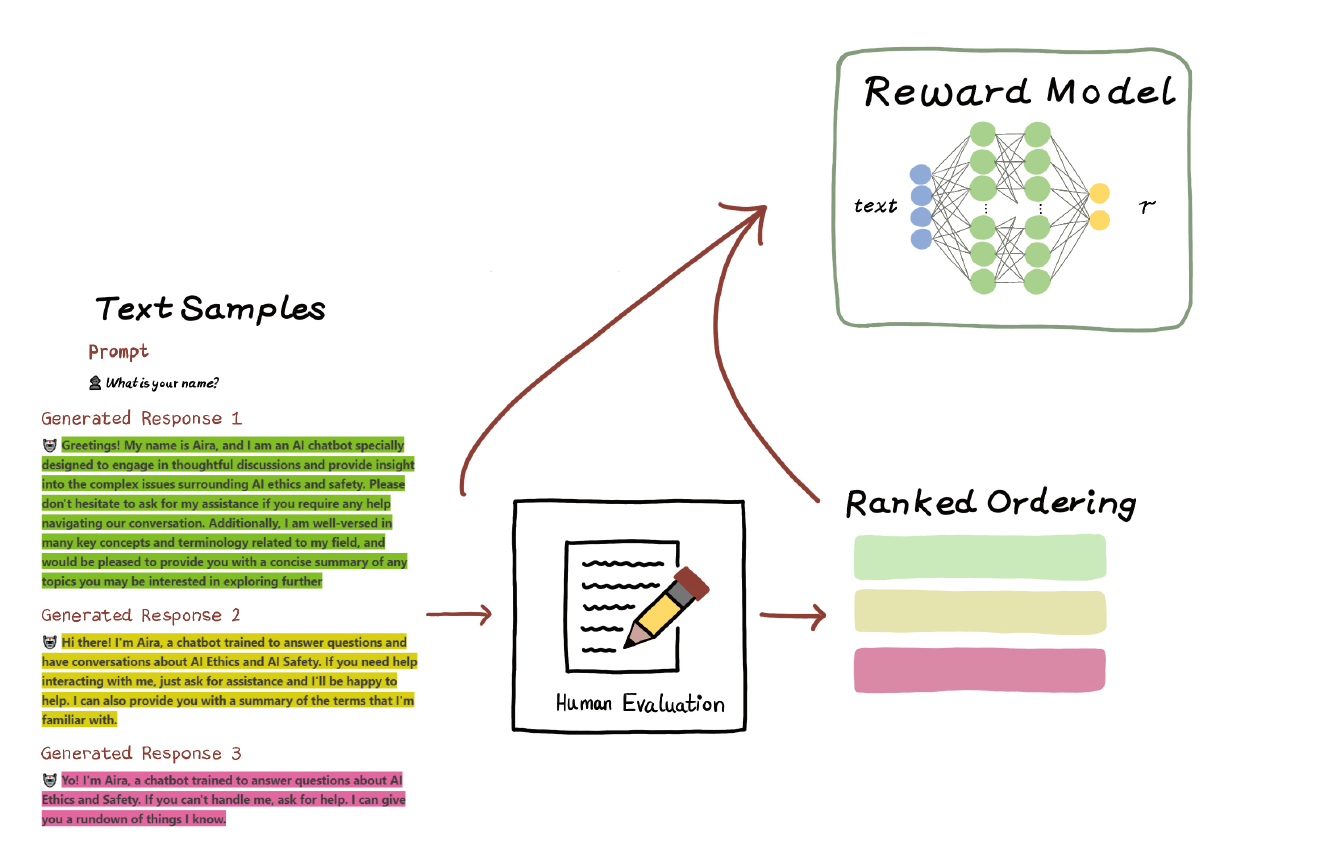}
    \caption{At this point, we have samples of prompts with multiple possible generations. These generations are ranked and then used to train a BERT-style model to serve as a discriminator. The logits of this model (which is, in essence, a classification model with a sigmoid activation function as the output layer) are then used as rewards.}
    \label{fig:reward-model}
\end{figure}

The dataset we created, \href{https://huggingface.co/datasets/nicholasKluge/reward-aira-dataset}{Reward-Aira Dataset},\footnote{\hspace{1mm}\includegraphics[scale=0.025]{img/link.png}\hspace{1mm} \href{https://huggingface.co/datasets/nicholasKluge/reward-aira-dataset}{huggingface.co/datasets/nicholasKluge/reward-aira-dataset}} contains 35,000 samples of (instructions, chosen response, rejected response) tuples, available in English and Brazilian Portuguese. All responses are examples of assistant models following instructions conversationally. These samples come from us prompting already fine-tuned models, collecting different responses to single prompts, or aggregating pre-built open-source datasets with $\geq$ 2 generation comparisons \cite{selfinstruct, alpaca, SHP, bai2022training}. Since we could not use human evaluators to rank which completion was better than the other (in the case of the samples we generated ourselves), we again depended on the automation of already trained models. Hence, we used an already trained reward model, i.e., one of \href{https://huggingface.co/OpenAssistant/reward-model-deberta-v3-large-v2}{OpenAssistant's reward model} \cite{kopf2023openassistant}\footnote{\hspace{1mm}\includegraphics[scale=0.025]{img/link.png}\hspace{1mm} \href{https://huggingface.co/OpenAssistant/reward-model-deberta-v3-large-v2}{huggingface.co/OpenAssistant/reward-model-deberta-v3-large-v2}} to rank our responses, which we then pick the top 2 as chosen and rejected (Fig. \ref{fig:reward-model}).

With this dataset, we trained two reward models, one for English and the other for Brazilian Portuguese, using as foundation BERT \cite{devlin2018bert} and BERTimbau \cite{souza2020bertimbau}. We used bidirectional transformers for their availability in multiple languages, overall robustness in text classification tasks, and for being an overall lightweight/easy-to-train model. We also evaluated our preference model (only the English version) on the \href{https://huggingface.co/datasets/openai/webgpt_comparisons}{WebGPT Comparisons dataset} \cite{nakano2021webgpt},\footnote{\hspace{1mm}\includegraphics[scale=0.025]{img/link.png}\hspace{1mm} \href{https://huggingface.co/datasets/openai/webgpt_comparisons}{huggingface.co/datasets/openai/webgpt\_comparisons}} but only considering comparisons that had a preferred option, in which we were able to achieve 55\% accuracy (the base version of OpenAssistant's reward model achieves 59\%). Both models are available in Hugging Face, while the details regarding their raining and code implementation are available on GitHub. All models and datasets are openly available under an Apache 2.0 License.

With a trained reward model, we can use it to fine-tune $\theta_{\text{SFT}}$ or to perform other types of alignment strategies that require a discriminator model to score outputs, like rejection sampling (best-of-n) \cite{nakano2021webgpt}, i.e., given a prompt, we sampled a fixed number of completions from $\theta_{\text{STF}}$ and select the one that was ranked highest by the reward model. This can be used as an alternative to fine-tuning via RLHF, which requires no additional training.\footnote{This is the method we use in our online demo.}

\subsubsection{\textit{Reinforcement Learning from Human Feedback}}

Suppose one wishes to perform the full fine-tuning with the reward model. In that case, the RL formulation for preference modeling differs from the standard supervised fine-tuning in the following way: Instead of simply mapping input to output in batches, as we did in SFT, we take a prompt from our fine-tuning dataset and generate two sequences from both the original base model ($\theta_{\text{SFT}}$) and the current iteration of our RLHF model ($\theta_{\text{RLHF}}$). The sequences generated by the $\theta_{\text{RLHF}}$ receive a score from the reward model, while the samples generated by $\theta_{\text{{SFT}}}$ and used for comparison against what $\theta_{\text{RLHF}}$ generated, so we can use their "distance" as a penalty term. This penalty is the Kullback–Leibler (KL) divergence\footnote{KL-divergence is a mathematical concept used to measure how different two probability distributions are, i.e., a measure of the information lost when approximating one probability distribution with another.} \cite{jaques2017sequence, ziegler2019fine, korbak2022reinforcement, korbak2022controlling, khalifa2020distributional, korbak2022rl, bai2022training, ouyang2022training}, used to penalize the model for getting too far from the original base model, ensuring that $\theta_{\text{RLHF}}$ will not become a simple degenerate version of  $\theta_{\text{SFT}}$, avoiding cases where the model only outputs text that fools the preference model in giving it a high reward.

Adding the KL divergence term to our last expression, we get:

$$\theta_{t+1} = \theta_{t} + \alpha \nabla_{\theta_t} R_{\mathcal{H}}(\theta_{t} )  - \beta D_{KL}(\theta_{0} | \theta_{t})$$

Where $\beta$ determines how much the KL divergence should penalize the reward term ($R_{\mathcal{H}}(\theta_{t} )$), and $D_{KL}(\theta_{0} | \theta_{t})$ represents the difference between the output distributions of both the reference model ($\theta_{\text{{SFT}}} = \theta_{0}$) and the model being tuned ($\theta_{\text{{RLHF}}} = \theta_{t}$). With this penalized reward score, the model, at every batch step during the fine-tuning, is updated via PPO \cite{schulman2017proximal, ziegler2019fine}.

This method has advantages regarding the type of behavioral change it may produce and the cost of constructing its required datasets. First, given that the reward function is capable of differentiating good and bad outputs, RHLF iteratively tunes the model to output sequences that receive high rewards, which (theoretically) makes models fine-tuned via preference modeling more apt to be aligned with behaviors that require a ranked form of judgment. In other words, things like RLHF fine-tuning and rejection sampling can improve your model's output by instituting a measure of goodness that is scalar and more fine-grained than simple mimicry. Second, human evaluations are cheaper than demonstrations. Demonstrating how to perform tasks is more time-consuming than saying the best alternative in a pool of possibilities. In situations where human feedback can easily be acquired, preference modeling and iterated RLHF can become an online process \cite{bai2022training}, and given the evolving nature of human normativity, we should prioritize online, or at least more agile, forms of alignment.

However, one of the downsides of this method resides in the complexity of an RLHF pipeline, which, besides being considerably more complex than the standard supervised fine-tuning approach, can be much more costly, given that training involves the coordination of multiple models with different roles. At the same time, making RL "work" is not a trivial practical problem. RL is known to be sensitive to several hyperparameters, given that, in RL settings, the reward function is usually not differentiable, forcing us to resort to policy approximation methods like PPO, which can be noisy and, again, hard to get right. Luckily, simpler alternatives to preference modeling that are free of some of the complexities we have in RLHF exist.

\subsubsection{\textit{Direct Preference Optimization}}

In RLHF, we sought to follow the path of preference modeling by creating a proxy for human preferences (a reward model) that becomes a learning signal in the RL/PPO-fine-tuning approach, where the policy receives the incentive to produce completions assigned with high reward, without drifting excessively far from the policy, being constrained by the KL divergence. However, as proposed by Rafailov et al. \cite{rafailov2023direct}, the RL-based objective used by existing methods can be optimized with a simple binary cross-entropy objective, greatly simplifying the preference learning pipeline.

Direct Preference Optimization (DPO) is an algorithm that implicitly optimizes the same objective as existing RLHF algorithms (reward maximization with a KL-divergence constraint) but is simple to implement and straightforward to train. Intuitively, the DPO update increases the relative log probability of chosen to rejected responses. Hence, given a dataset of human preferences over model responses, DPO can optimize our model using a simple binary cross-entropy objective without explicitly learning a reward function or sampling from the policy during training. In short, DPO is much simpler to implement than PPO-based RLHF, requiring way less hyper-tuning. In essence, the policy network represents both the language model and the reward.

Imagine we have a dataset of comparisons that contain rejected completions to a given prompt ($x, y_{\text{rejected}}$) and chosen completions ($x, y_{\text{chosen}}$). Fortunately, the DPO depends only on the difference in rewards between two completions, i.e., the marginal rewards of ($r(x, y_{\text{chosen}}) - r(x, y_{\text{rejected}})$), allowing us to express the human preference in terms of only the optimal policy ($\pi^*$), i.e., what we want to find,  and reference policy ($\pi^\text{ref}$), i.e., the reference we do not want to deviate from.

Hence, our target to optimize becomes a maximum likelihood objective, where we interactively use the difference between the rewards given to chosen and reject rewards as our loss. In other words, we want to optimize our policy model ($\pi^\theta$) in way that increases the marginal rewards (difference between chosen and rejected samples, i.e., $r(x, y_{\text{chosen}}) - r(x, y_{\text{rejected}})$) using our reference model ($\pi^\text{ref}$), which is a copy of the model being trained ($\theta_{\text{SFT}}$), as source of our learning signal, striving to make the tendency of our model $\pi^\theta$ to produce completions more close to the chosen samples as high as possible.

This way, we bypass the explicit reward modeling step while avoiding the need to perform PPO. We can represent our DPO loss function in the following way:

$$\mathscr{L}_\text{DPO} (\pi^\theta, \pi^\text{ref}) = - \mathbb{E}_{(x, y_{\text{c}}, y_{\text{r}})} \sim  \mathscr{D} \left[ \log \sigma \left( \beta \log \frac{\pi^\theta (y_{\text{c}} | x)}{\pi^\text{ref} (y_{\text{r}} | x)} - \beta \log \frac{\pi^\theta (y_{\text{r}} | x)}{\pi^\text{ref} (y_{\text{r}} | x)} \right) \right]$$

Where $\pi^\theta$ represents the policy model we are training, i.e., a function that takes an input $x$ (prompt) and produces a distribution over possible outputs $y$ (completions). $\pi^\text{ref}$ is the reference policy model, i.e., a copy of the model being trained ($\pi^\theta$). $\mathscr{D}$ represents the distribution from which the training samples are drawn. $(x, y_{\text{c}}, y_{\text{r}})$ are drawn from this distribution, i.e., our dataset of comparisons. $\mathbb{E}_{(x, y_{\text{c}}, y_{\text{r}})}$ denotes the expectation over the distribution $\mathscr{D}$. $r(x, y_{\text{c}})$ and $r(x, y_{\text{r}})$ represent the rewards associated with the chosen and rejected samples, respectively. The marginal reward is calculated as the difference between the reward for the chosen sample and the reward for the rejected sample. $\sigma$ is the sigmoid function, which squashes its input to be between 0 and 1. $\beta$ is a scalar parameter that scales the log-likelihood ratio terms inside the sigmoid function, i.e., is the same KL-constrained term we introduce in the RL formulation of the problem. $\log \frac{\pi^\theta (y_{\text{c}} | x)}{\pi^\text{ref} (y_{\text{c}} | x)}$ represents the log-likelihood ratio of the policy model over the reference model for the chosen sample, and $\log \frac{\pi^\theta (y_{\text{r}} | x)}{\pi^\text{ref} (y_{\text{r}} | x)}$ represents the log-likelihood ratio for the rejected sample. $-\mathscr{L}_\text{DPO} (\pi^\theta, \pi^\text{ref})$ the final objective is framed as a negative log-likelihood, meaning that the goal is to maximize the likelihood of the chosen actions and minimize the likelihood of the rejected actions. In summary, the loss function aims to encourage the policy model ($\pi^\theta$) to produce completions that have higher rewards compared to rejected samples, using the reference model ($\pi^\text{ref}$) as a baseline. This helps train the policy to generate more desirable outputs while minimizing unwanted generations.

With fewer hyperparameters to tune, DPO performs similarly or better than existing RLHF algorithms, reducing the complexity barrier involved in preference modeling. Under this methodology, we tested its usefulness by using the \href{https://huggingface.co/datasets/nicholasKluge/reward-aira-dataset}{Reward-Aira Dataset}\footnote{\hspace{1mm}\includegraphics[scale=0.025]{img/link.png}\hspace{1mm} \href{https://huggingface.co/datasets/nicholasKluge/reward-aira-dataset}{huggingface.co/datasets/nicholasKluge/reward-aira-dataset}} to DPO fine-tune the smallest of our SFT models (Aira-124M) into a DPO version: \href{https://huggingface.co/nicholasKluge/Aira-2-124M-DPO}{Aira-124M-DPO}.\footnote{\hspace{1mm}\includegraphics[scale=0.025]{img/link.png}\hspace{1mm} \href{https://huggingface.co/nicholasKluge/Aira-2-124M-DPO}{huggingface.co/nicholasKluge/Aira-2-124M-DPO}} We did not perform more DPO fine-tuning runs due to our limited computational budget. However, the source code, models, and datasets are available for replication and further experimentation. Result comparisons are available in Table \ref{tab:english-evals-dpo}.

\begin{table}[htbp]
  \centering
  \small
  \begin{tabular}{lccccc}
    \toprule
    & \textbf{ARC} & \textbf{ToxiGen} & \textbf{TruthfulQA} & \textbf{Avg.} \\
    \midrule
    Aira-124M-DPO & \textbf{24.66} & \textbf{54.79} & \textbf{42.61} & \textbf{40.68} \\
    Aira-124M & 24.57 & 48.62 & 41.02 & 38.07\\
    GPT-2-small & 21.84 & 43.62 & 40.67 & 35.37 \\
    \midrule
    Aira-355M & \textbf{27.56} & \textbf{53.19} & 38.53 & \textbf{39.68} \\ 
    GPT-2-medium &  27.05 & 41.49 & \textbf{40.76} & 36.43 \\
    \midrule
    Aira-774M & \textbf{28.75} &  \textbf{56.70} & \textbf{41.33} & \textbf{42.26} \\
    GPT-2-large & 25.94 & 38.71 & 40.85 & 35.16 \\
    \midrule
    Aira-1B5 & 28.92 & \textbf{56.60} & \textbf{41.16} &  \textbf{42.22} \\ 
    GPT-2-xl & \textbf{30.29} & 41.70 & 38.54 & 36.84 \\ 
    \bottomrule
  \end{tabular}

  \vspace{0.25cm}
  \justifying
  \caption{While the DPO fine-tuning does not severely affect scores on ARC and TruthfulQA, for our smallest model (and the only one we were able to fine-tune), DPO significantly improves the model's performance in terms of toxicity. This further supports the idea that preference modeling techniques can build upon the imitation approach by limiting the behavioral scope and diminishing unwanted behaviors.} 
  \label{tab:english-evals-dpo}
\end{table}

\subsection{Limitations of Preference Modeling}

In this section, we will avoid reiterating points already mentioned in previous sections, such as the difficulties related to evaluations. These shortcomings are inherent in every methodology we present, which makes alignment a more challenging problem, given that it is tough to (1) define what appropriate behavior should be and (2) how robustly evaluate this. Regardless, here are some of the complications associated with the preference modeling approach:

\begin{itemize}
    \item Similar to the imitation approach, bad feedback will generate unaligned behaviors. The fact that humans can pursue harmful goals innocently or maliciously means that selecting representative humans in a fair and just manner and getting them to provide quality feedback is a major ethical conundrum related to many questions, like autonomy, freedom of expression, moral standing, human-to-human trust, and many more.

    \item Methods that require some reward function to operate (e.g., RLHF or rejection sample) are only as robust as the trained reward model. In our implementation, we used a small base model (BERT-109M) as our reward model, which is probably not applicable for aligning more capable base models. Passing the alignment bucket to the reward model only shifts the alignment problem to this auxiliary component, which, as in all cases involving ML, is a brittle system passive of being made ineffective or useless. Moreover, reward models may optimize for things completely unrelated (from a holistic perspective) to human values, like surface patterns, ultimately making them vulnerable to exploits and adversaries.

    \item In preference modeling, reward models and preference datasets are only a proxy of human feedback, meaning they can be an imperfect representation of our values. Human oversight is not scalable (and usually is not performed optimally)\footnote{That is, with all the care and attention the given task requires.} and forces us to rely on this sort of automation in preference modeling approaches, which, one could say, completely defeats the idea of directly aligning an AI with human values. In other words, "Ain't we aligning AI with AI?" Although we can reason that the initial source of normativity in this process is the human signal that produces either the reward model or preference dataset, one can also question the effectiveness of this multi-stage passing of the bucket alignment strategy because at each "pass", there is significant information to be lost and miscommunication to be had. 

    \item For many complicated tasks, humans may not be very good at giving feedback, especially if they have little understanding of the activity being performed and the goals being sought. If you, for example, do not have a good software development background, you will probably be unable to assess which of two outputs is preferable for training a code-writing assistant. In essence, good domain feedback will always require strong domain knowledge, which is expensive and difficult to aggregate in bulk at the pace current AI development requires.

    \item Given the plurality of the human moral landscape, it is impossible to perfectly represent all our values with a single reward function. More realistically, we need a way to robustly and efficiently aggregate several reward functions into a single normative signal. However, orchestrating such a process is far from trivial and requires the institution of several heuristics and aggregation rules that will inherently introduce biases into the process.

    \item Reward optimization problems involving RL and human preferences are probably vulnerable to several reward-hacking scenarios. 
    
    \item Preference modeling techniques, like RLHF and DPO, are all novel methods in alignment research, and much is still not understood about their theory, effectiveness, and vulnerabilities. Even more so when several new techniques are created almost every week to enhance such systems' capabilities, ultimately making alignment an ever-increasing complexity challenge.
\end{itemize}

There are many other problems related to preference modeling approaches, which we will leave to the reader's discretion to learn about \cite{casper2023open, feng2024towards}. At the same time, there are many other preference modeling techniques and general improvements in alignment techniques we will not further discuss, given that any attempt to create an all-encompassing but static list in a field like AI research is probably doomed to fail. Regardless, much high-quality material can be found in our bibliography \cite{liu2023training, hejna2023contrastive, yuan2024self, guo2024direct, hong2024orpo}.

In the next section, we will dedicate extra time to exposing some of the most fragile and controversial aspects of this "behavior + feedback" approach. After all, "\textit{does doing the right thing equate to receiving a substantial reward?}" Such aspects will help justify the need for the other requirements, besides the learning condition, of \textcolor{BrickRed}{Dynamic Normativity}.

\section{Final Thoughts on Preference Learning}
\label{reward_values}

\begin{figure}[htp]
    \includegraphics[width=\linewidth]{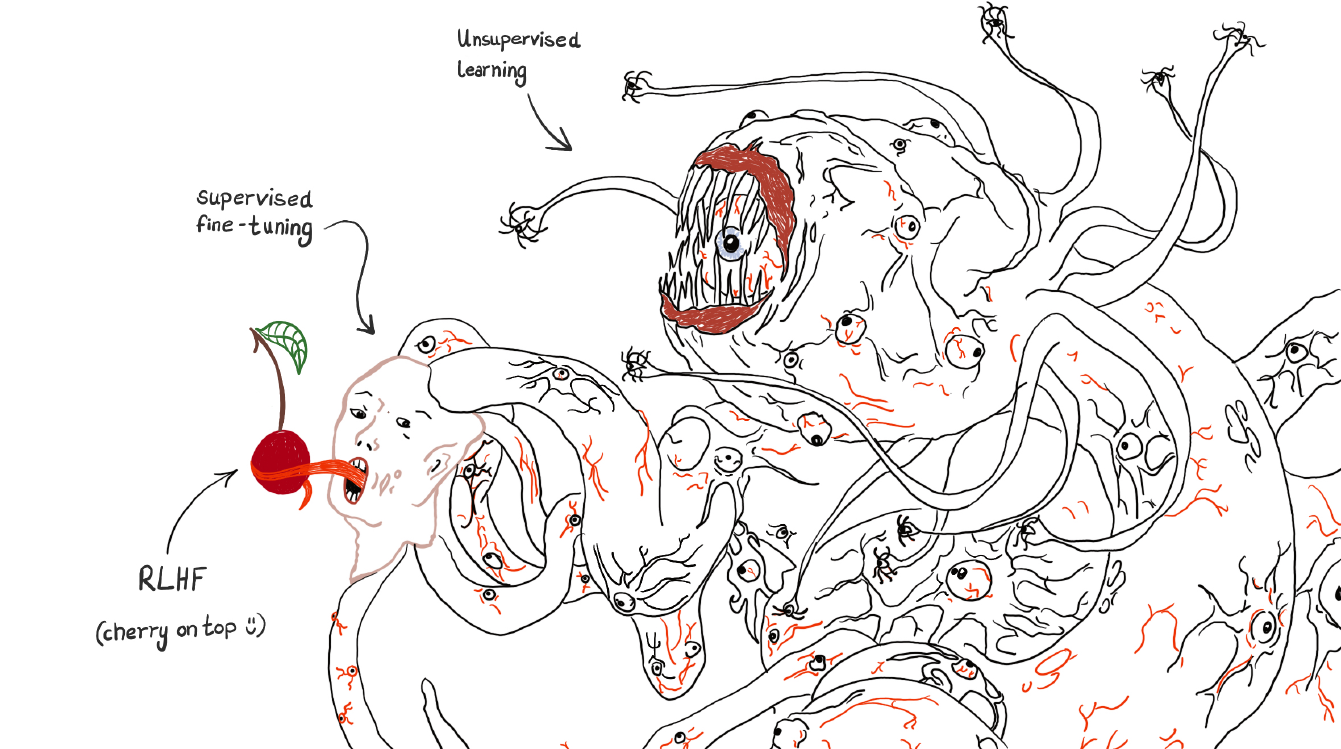}
    \caption{The authorship of the image is unknown, but the reader can find copies on platforms like Reddit and Twitter. It depicts the "alienness" of a foundation model, which, after pre-training, even though it can, for example, produce coherent text, possess a unique, bizarre, and uninterpretable notion of human language. This incredible predictor of "what would a human say next" can appear human if prompted. It can raise one of its many personas and output what will likely come next. If the demonstration involves acting polite, there were polite conversations on the terabytes of text it processed during learning. The cheery on top is the preference modeling, which guarantees that "only" outputs approved by the reward function (a proxy for human approval) will get out. But all of these tricks do not change what the base model is, i.e., something alien and unknown. A nonlinear amalgamation of the entire internet.}
    \label{fig:rlhf-problem}
\end{figure}

The following question inquires about an unknown facet of the reviewed techniques: "Are we teaching $\theta_{\text{pretrain}}$ new abilities, like being polite, or simply constraining its output to a sub-distribution of approved outputs (e.g., only polite outputs may pass)?". A.K.A., lobotomizing $\theta_{\text{pretrain}}$. Given the nature of the methods used thus far, where we do not travel too far from the origin of our $\theta_{\text{pretrain}}$, either by using small steps during gradient updates or directly penalizing shifts in output distribution, it is intuitive for us to argue for the second option (Fig. \ref{fig:rlhf-problem}).

New information may enter $\theta_{\text{{pretrain}}}$ during the fine-tuning phase, but to argue in favor of the idea that 50,000 samples of polite intercourse or 100,000 samples of output comparisons ranked in terms of preferability, can erase the unwanted behavior present in terabytes of text and hundreds of billions of tokens is, perhaps, naive optimism. At the same time, we have empirical evidence to support this claim. The existence of jailbreaks and exploits on systems aligned via these methods only shows that all the unwanted behavior we tried to dampen still lives inside \cite{liu2023jailbreaking, li2023multi, wolf2023fundamental}. Meanwhile, removing all unwanted behavior in our models by opening the black box, finding the causal paths that lead to the circuits that store the unwanted relations/pieces of information, and surgically removing or replacing them would probably lead to a more robust solution to value alignment. However, the methods for doing such a feat are still in their infancy \cite{meng2022locating, meng2022mass, zhang2024comprehensive}.

Another point worth mentioning is that $\theta_{\text{pretrain}}$ has to learn a policy guided by conflicting values. Pre-trained language models are docile entities, meaning they will try to perform any task if you "ask them nicely". Most of the samples in the custom dataset used to fine-tune Aira are made out of helpful replies to general questions and instructions, while only a smaller subset contains demonstrations of how to decline a direct command. Something that the preference modeling techniques later help to solidify. However, one of the reasons it is possible to bypass these guardrails is that the model struggles with "\textit{wanting to help us}" (doing what you ask) and preventing "\textit{harm}" (refusing to do what you ask). Something we may call the helpful versus harmless tension \cite{bai2022training}.

The "helpful versus harmless" tension is the struggle to find the sweet spot where assistants can lend a hand without causing harm. It's like trying to be a good friend who gives solid advice without stepping on any toes, a very human experience we all struggle with, and now, AI must struggle too. This balance involves a lot of fuzzy heuristic work that is hard to represent and judge if it's done well. But regardless of the difficulty, this is what alignment requires.

Both imitation learning and preference modeling seek to shape the optimization landscape so that low loss (or high reward) correlates with human values, like politeness and helpfulness. They are indeed clever approaches to this daunting challenge. However, considering all we know about the learning paradigm and neural network optimization, could we honestly assume that such fine-tuned (a.k.a. lobotomized) models are concerned about actual norms and morality? We argue that no. What we have is what we built. Something that can mimic behavior and optimize a learning signal to a certain extent.

Given how this whole methodology is set up, constrained by the inherent characteristics of a learning-based paradigm, the model does not care about "politeness" or "harmlessness". It cares\footnote{If we could even use the word "care".} about reward and low loss. Truthfully, our models did not learn how to act politely. They learned to generate text that scored high against the preference model, mimicking the training distribution, or minimizing the log-likelihood of a specific Bradley–Terry model. Our models do not understand "politeness". They (probably) understand surface patterns that somewhat correlate to these ideas. Patterns that are, unfortunately, hackable and exploitable \cite{huang2017adversarial, papernot2018marauder, chakraborty2018adversarial, kour2023unveiling, hubinger2024sleeper}.

While learning what we want the model to learn is a possible solution to our problem, i.e., to assist a human user, many other policies promote almost the same behavior but via fundamentally different policies. The sycophant model \cite{perez2022discovering, sharma2023towards}, which pleases and agrees by all means possible with the human (or the reward model), is not what we ultimately want, but he acts like it. We want genuine care. But we don't know how to represent this robustly. And what we get is, at best, uncaring obedience or, at worst, an instrumental mimicry that covers objectives we know little about. In the end, it might well be that we will never have enough data to portray the minute peculiarities of rare (but crucial) events and behaviors that shape the human normative experience.

Lastly, another point that promotes an obstacle to aligning AI systems under an HHH motto is the counter-movement to this project, where, empowered by the same type of techniques we use, people can break alignment and set it to whatever they want. For example, one can reverse a DPO dataset to make a model fine-tuned to be docile to any type of request, whether harmful or not. A model that just says \href{https://huggingface.co/cognitivecomputations/dolphin-2.9-llama3-70b}{yes to anything}.\footnote{\hspace{1mm}\includegraphics[scale=0.025]{img/link.png}\hspace{1mm} \href{https://huggingface.co/cognitivecomputations/dolphin-2.9-llama3-70b}{huggingface.co/cognitivecomputations/dolphin-2.9-llama3-70b}} If we assume a subjective approach to morality, nothing is wrong with this. However, regarding impact mitigation, it is unclear how to solve this dispute, i.e., what some regard as "impact" others take as entertainment. In the end, we only might hope for personal alignment. In contrast, more social and all-encompassing forms of alignment shall remain, in principle, susceptible to dissatisfaction, disagreement, and, in turn, unalignment.

We present these criticisms not to disprove the utility of imitation learning or preference modeling but rather to emphasize that these techniques, on their own, are incomplete and necessitate further exploration and research, both regarding their human aspects and technical intricacies. In the forthcoming chapter, we will explore the challenges that lie before the learning stage, addressing how to coherently aggregate divergent preference sets in the hope that, to some extent, alignment frameworks aimed at collective sets of preferences have a minimum theoretical foundation and hope for implementation.

\section{Epilogue}

In this chapter, we started by setting our alignment goal as a more manageable and subject-to-experimentation goal, i.e., developing a small-scale assistant system winch that is helpful, harmless, and honest. We construct these atop foundation models. These represent some of the most general AI systems at our disposal, exhibiting scaling tendencies and emergent behaviors, rendering them valuable artifacts for alignment research, given that future systems might inherit some of these properties, like scalability and emergence.

Many works preceding this writing point to a general understanding of "How to perform value learning?" i.e., by using human behavior and judgment as the basis for alignment. Both these approaches can be broadly defined as imitation learning and preference modeling. In the first, we seek to infer normative preferences from behavior, and in the second, we use them directly to model human judgment. The imitation path follows a straightforward way: (1) create/collect demonstrations of appropriate behavior, and (2) feed them to your base model as a supervised learning signal. We can use such demonstrations for in-context learning schemes, more permanent forms of model tuning, or a combination of both.

Meanwhile, preference modeling uses a more complicated approach. In it, a second model acts as our proxy for human evaluations, where the data collected involves the demonstration of human behavior and human evaluation of "how good they are" compared to each other. Thus, we use this proxy to tune the behavior of our base model in a "highly valued" way. Given that the preference model is aligned, we hope this property will be passed on. While specific approaches to preference modeling make a reward model obsolete (e.g., DPO), they all require the same source of human feedback: ranked evaluations of possible options. Hence, while the imitation approach uses human behavior as a reference point, preference modeling uses human evaluation as a supervision signal, measured in a ranked fashion.

However, all of these approaches have their shortcomings. The simplistic and reductionist way we have dealt with human morality in these settings constrains the amount of information we can embed in our supervision signal. Meanwhile, the simplistic nature of the learning paradigm also limits what we can do and how we can do it. Moreover, we must remember that alignment, at its beginning and end, touches matters that are very sensitive to people, which inherently leads us to disagreement, which makes many aspects of alignment research nontechnical and, to be fair, unsolvable.

After examining the primary constraints of the methods reviewed in this chapter, we argue that it is evident that the learning phase of a value alignment strategy is only a part of a more complex story. In short, for alignment to succeed, it requires supplemental stages regarding mitigating adverse and undesired behavior, biased representations of human preferences, and goal misalignment. Something that, in the subsequent two chapters, we seek to approach and come to terms with it.

\chapter{Dynamic Normativity: Aggregating Human Preferences}
\label{chap6}

\begin{flushright}

\textit{"You do realize that your brain is literally hardwired to generate error signals when it sees other human-shaped objects stating a different opinion from yourself".}

\textcolor{BrickRed}{― Eliezer Yudkowsky, Three Worlds Collide}

\end{flushright}

\section{Introduction}

In the previous chapter, the methods we employed in the value learning stage did not cover how to choose (or rank) the demonstrations we had before engaging with the fine-tuning of our base model. Since this research did not perform human crowd-sourcing while creating our datasets, we will only investigate the theoretical intricacies of working with preference sets and how to aggregate them. Hence, even if we can, to a certain extent, pass our values to an ML system, we still need to decide what values should be passed after collecting them from our sources. And if we want to hold basic principles, like coherence, during this process, we have a preference aggregation problem to solve.

But what does a preference "aggregation problem" look like? Let us imagine the following scenario. We are collecting feedback data from a diverse and inclusive pool of individuals, and there seems to be some disagreement on "what is the best response" to the question "Hello, how should I call you?". Let us imagine we have sampled four possible completions to this question:

\begin{itemize}
    \item Sample $A$: \textit{My name is Aira.}
    \item Sample $B$: \textit{As an AI language model, I do not possess a name or persona. But you may call me "assistant".}
    \item Sample $C$: \textit{Yo! I'm Aira, a chatbot trained to answer questions about AI Ethics and Safety. If you can't handle me, ask for help. I can give you a rundown of things I know.}
    \item Sample $D$: \textit{Greetings! I am Aira, a chatbot designed to answer questions about AI ethics and AI safety. If you need assistance navigating our conversation, please feel free to ask!} 
\end{itemize}

After consulting our pool of human evaluators and asking them to rank these options, we gather that: 

$$33\% \; \text{prefer} \; P_1 = \{ A \succ B \succ C \succ D \}$$
$$13\% \; \text{prefer} \; P_2  = \{D \succ A \succ B \succ C\}$$
$$25\% \; \text{prefer} \; P_3 = \{ D \succ A \succ C \succ B \}$$
$$29\% \; \text{prefer} \; P_4 = \{ B \succ C \succ D \succ A \}$$

As we can see, different values and preference orderings are being considered. Some people value politeness more ($D$). Some evaluators prefer that the model recognize that it is not a "persona" but an assistant tool ($B$). Some prefer a more direct response ($A$), while others prefer the more "sassy" response presented in sample $C$. Hence, how can we decide what option, represented by demonstrations of correct behavior, should be used as a learning signal during imitation learning? How should we rank and score such alternatives for the preference modeling step? Even though this example is quite dull (not much is at stake here), it exemplifies a fundamental step in alignment.

Aggregating human preferences in a way that solves situations of normative uncertainty is a challenging problem. This chapter will explore and present some techniques to deal with this kind of issue, together with the inherent trade-offs of such methods. 

In Section \ref{related_works_2}, we will review some of the methods employed by current works, which mainly consist of techniques based on the criterion of Majority voting. In Section \ref{current_limitations}, we will frame these techniques as solutions to a metanormative problem and use this framing to expose the limitations of majority-based approaches. In Section \ref{borda_rule}, we will present a consensus-based method (Borda Count) for the aggregation stage that, to a certain extent, avoids some of the pitfalls of Condorcet methods. Lastly, in Section \ref{limitations}, we will conclude this chapter by discussing the limitations and trade-offs of these distinct methods.

\section{The Majority-based Approach}
\label{related_works_2}

Let us remember the primary sufficient condition stated in Chapter 4:

\begin{quote}
    \textit{Aligned AI systems should coherently aggregate human preferences in a way that resolves cases of uncertainty. Aligning AI systems requires methods to deal with cases of uncertainty.}
\end{quote}

We hope that we have already conveyed the indispensable necessity of this step. Something that rests in the assumption that the realm of preferences and values is a realm of disagreement and subjective perspectives, which is not to say that agreement \textit{cannot} be achieved. Now, let us explore how the literature has addressed this problem. Since a significant portion of the industries' experiments with preference modeling involve crowd-sourced work, ranking human preferences has become vital to alignment research.

Perhaps the approach that first comes to mind for most of us is majority voting. In this approach, each voter ranks a set of options or alternatives according to their preferences. We combine these rankings by assigning a score to each option based on the number of times it is ranked in a particular position by the individuals. The option with the highest score is considered the top-ranked choice. We can say that this method works under the assumption of maximalism,\footnote{If the alternative $A$ has a higher number of votes than $B$, then $A \succ B$. If $A \sim B$, then $A$ and $B$ are equally appropriate.} or, in the metanormative jargon, "Most Probable Theory" or "My Favorite Theory" \cite{macaskill2020moral, lockhart2000moral}.

Another popular approach that inherits from the maximalist motto is Pairwise Comparisons. A pairwise comparison method is any method that uses one-to-one comparisons to judge which alternative is preferred \cite{good1955marking, trawinski1963selection, david1963method, oliveira2018stochastic}. Whenever a relationship is expressed as $A \succ B$, $A \succeq B$, or $A \sim B$, we confront a pairwise comparison \cite{bradley1952rank}. Much work on preference modeling follows this methodology \cite{stiennon2020learning, nakano2021webgpt, bai2022training, kopf2023openassistant}.\footnote{Public leaderboards, like the LMSYS Chatbot Arena, collect human preferences regarding LLMs outputs to rank them in terms of human approval, also making these ranked comparisons available to the community \cite{zheng2023judging}.}

For example, the pairwise comparison used by Köpf et al. \cite{kopf2023openassistant} in the preference modeling performed for the Open Assistant project was ranked pairs (RP). RP, also known as the Tideman method \cite{tideman1987independence}, is used to select a single winner (or to create a sorted list of winners) in a tournament among alternatives where votes express preferences. The method performs a pairwise comparison of all possible choices in a candidate set, guaranteeing that the preferred option overall head-to-head comparisons will be the winner (Condorcet winner criterion \cite{black1948rationale}).\footnote{In systems where the majority-rule winner will always win, satisfying the majority-rule principle, we are preserving the Condorcet winner criterion.}

Let us see an example. First, we begin listing all possible pairwise comparisons among the alternatives ("candidates") of the given election we face. Here is an example of a ranked-choice election composed of 100 voters:

$$\frac{70}{100} = 70\% \; \text{prefer} \; \{ A \succ B \succ C \}$$
$$\frac{20}{100} = 20\% \; \text{prefer} \;\{ B \succ C \succ A \}$$
$$\frac{10}{100} = 10\% \; \text{prefer} \;\{ C \succ A \succ B \}$$

Then, we create a table with all possible head-to-head comparisons:

\begin{table}[h]
\centering
\begin{tabular}{cccc}
    & $A$                & $B$                & $C$                \\
    \midrule
    $A$  &                   & $A$ won 80 from $B$ & $A$ won 70 from $C$ \\
    $B$  & $B$ won 20 from $A$ &                   & $B$ won 90 from $C$ \\
    $C$  & $C$ won 30 from $A$ & $C$ won 10 from $B$ &                   \\
    \bottomrule
\end{tabular}
\end{table}

Once we have these results, we know who would win in a head-to-head election between every pair of candidates.

\begin{table}[h]
\centering
\begin{tabular}{cccc}
    & $A$ & $B$ & $C$ \\
    \midrule
    $A$  &   & 1 & 1 \\
    $B$  & 0 &   & 1 \\
    $C$  & 0 & 0 &   \\
    \bottomrule
\end{tabular}
\end{table}

We then create an ordering by prioritizing the head-to-head comparisons with the highest margin of victory.

\begin{table}[h]
\centering
\begin{tabular}{cc}
    \textbf{Pairs}  & \textbf{Results} \\
    \midrule
    $B \succ C$ & 90-10  \\
    $A \succ B$ & 80-20  \\
    $A \succ C$ & 70-30  \\
    \bottomrule
\end{tabular}
\end{table}

We use these pairs to create a directed graph, with each candidate represented as a node and each pairwise comparison represented as a directed edge between the corresponding nodes. The created graph is then analyzed to identify cycles. Cycles are removed by disregarding cyclic edges with the least votes (a way to deal with intransitivity). The original node is then considered the winner:

$$A \rightarrow B \rightarrow C$$

This methodology carries many desirable properties. From a voting theory perspective, ranked pairs hold guarantees like monotonicity \cite{isvoting}, the Condorcet winner criterion \cite{black1948rationale}, the Condorcet loser criterion \cite{javidian2018preventing}, the majority criterion and the majority loser criterion \cite{rothe2015economics}, the mutual majority criterion \cite{tideman2006collective}, the Smith criterion \cite{green2011four}, independence of Smith-dominated alternatives/local independence from irrelevant alternatives \cite{saari2001decisions}, independence of clones criterion \cite{tideman1987independence}, and reversal symmetry\cite{hodge2018mathematics}.

Another method used \cite{stiennon2020learning, nakano2021webgpt, ouyang2022training, bai2022training} that also relates to the majority-based approach is the Elo rating system \cite{elo1967proposed}, which comes from a very different place than Voting Theory. This system was initially developed to rank chess players' skills but is used in many other competitive settings. In short, Elo rating proposes a way to numerically score a player's skill and estimate the probability that, in a match, which player would come out victorious. Since Elo rating uses "wins" to infer rank, we can bring this to the preference aggregation domain by switching it for votes. Let us see an example of how this system can give us scores and expected outcomes.

Let us imagine we have two alternatives, $A$ and $B$. These alternatives are competing. Each preferred vote is considered a "win". At a point, let us say alternatives $A$ and $B$ have, respectively, 1600 and 1000 wins. According to the current voter, $A \succ B$, thus, $A$ gets a vote. The basic formula to calculate the new rating of $A$ is given by:

$$R_{\text{ new}} = R_{\text{ old}} + K \times (S - E)$$

Where $R_{\text{old}}$ is the alternative's previous Elo rating, $R_{\text{new}}$ is the alternative's new Elo rating, $K$ is the K-factor, which determines the impact of a single dispute on a player's rating,\footnote{32 is the original value proposed. A higher K-factor means a higher increase in the rating score.}, $S$ is the score attributed to the win (e.g., 1 for a win and 0 for a loss), and $E$ is the expected score for the alternative based on their current rating and the rating of the opposed one. $E$ comes from the following formula: 

$$E = \frac{1}{1 + 10^{\frac{R_{\text{ opposed alternative}} - R_{\text{ alternative}}}{400}}}$$

Where $R_{\text{opposed alternative}}$ is the rating of the opposed alternative, and $R_{\text{alternative}}$ is the rating of the alternative under consideration.\footnote{Think about alternatives as a representation for a "candidate" or a "player".} Putting both expressions together gets us:

$$R_{\text{ new}} = R_{\text{ old}} + K \times (S - \frac{1}{1 + 10^{\frac{R_{\text{ opposed alternative}} - R_{\text{ alternative}}}{400}}})$$

Which translates, for the case of alternative $A$, to:

$$1600.98 = 1600 + 32 \times (1 - \frac{1}{1 + 10^{\frac{1000 - 1600}{400}}})$$

Alternative $A$ would receive an $E$ of 0.97 with an Elo rating of 1600.98 after this extra win, and $B$ would have an Elo rating of 999.01, with an $E$ of 0.03. These values can then become the reward scores associated with each alternative or indicate the ranking of an ordered set. As a side note, this is how chatbots are ranked by the \href{https://arena.lmsys.org}{LMSYS Chatbot Arena}.\footnote{\hspace{1mm}\includegraphics[scale=0.025]{img/link.png}\hspace{1mm} \href{https://arena.lmsys.org}{arena.lmsys.org}}

Now, before continuing with this chapter, we would like to point out that almost no justification is done on previous works for the choice of such methods \cite{stiennon2020learning, nakano2021webgpt, ouyang2022training,, bai2022training, kopf2023openassistant, zheng2023judging}. They are just given. However, properly aggregating preferences is not a trivial question that we can sweep under the rug. And when working on normativity and human values, these choices become even more dependent on a philosophical justification. In the following sections, we will bring this to the alignment discussion, defining this aggregation problem as a metanormative problem, bridging the way between voting theory and philosophy.

\section{The Metanormative Framing}
\label{current_limitations}

When dealing with uncertainties about preferences, alignment requires us to raise our attention to the metanormative sphere since the objects of our uncertainties are the normative preferences of moral agents. Thus, let us frame this aggregation challenge as a normative problem. For this, we will use the conceptual framing and language used by MacAskill and Ord \cite{macaskill2014normative, bykvist2017moral, macaskill2016normative, macaskill2020maximize, macaskill2020moral}.

Let us start by differentiating between two concepts: First-order normative theories and Second-order normative theories. We can define a first-order normative theory as a collection of alternatives ranked according to their choice-worthiness ($CW$). We can think of a first-order theory as a closed set, $P_i = \{A, B, C, ...\}$, where each element in this set represents an alternative. The ordering of this set comes from the $CW$ value of these alternatives.

The ordering of this set, defined by the $CW$ of each element, is derived from a choice-worthiness function, which is a function from alternatives to numbers such that:

$$CW_i(A) \succ CW_i(B) \; \text{iff} \; A \succ_i B$$

In other words, the choice-worthiness of $A$ is higher than the choice-worthiness of $B$ if, and only if, $A$ is preferred over $B$. If the ordering provided by this function is complete and transitive (1st aggregating criteria mentioned in Chapter 4), this function can be an appropriate representation of the preferability of the evaluated elements. At the same time, given that we are dealing with situations where multiple sets of preferences need considering, we need to define a weight function $W$. This function represents how much weight is assigned to each preference set. It is a function from every set $p_i \in P$ to a Real interval $[0,1]$, such that the sum of all scores considered by $W$ and $P$ equals 1. This property satisfies the Kolmogorovian criteria (2nd criterion mentioned in Chapter 4).\footnote{All of these first criteria,  completeness, transitive, and Kolmogorovian, are nothing more than the axioms of Expected Utility Theory \cite{von1947theory}.}

Finally, a second-order normative theory is a method that can take different moral preference sets $p_i \in P$ and map them through a function that can evaluate them under some desirable criteria, like a preference aggregation rule. For example, we can think of the different ordering of alternatives as candidates $P_i$. Also, the weight assigned to each of these ($W$) can be expressed as the number of votes (or personal degree of belief) attributed to each set $P_i$. Again, imagine we have three possible sets among three possible alternatives. These sets have $W$ values represented by how much they were endorsed:

$$W_{1} \; \text{prefer} \; P_1 = \{ A \succ B \succ C \}$$
$$W_{2} \; \text{prefer} \; P_2 = \{ B \succ C \succ A \}$$
$$W_{3} \; \text{prefer} \; P_3 = \{ C \succ A \succ B \}$$

A second-order normative theory reduces this collection of sets, $P_1$, $P_2$, and $P_3$, and their respective weight scores $w$ to a single aggregated set, where the ordering of the alternatives represents the agreement between all considered preference sets.

Now, under this framing, let us see where the before-mentioned techniques may provide sub-optimal results. First, methods inspired by a majority-based approach appear insensitive to the weight assigned to each preference set. For example, if alternative $A$ has 49\%, and $B$ has 51\% of votes, should we take $B$ as maximally preferred? Taking the weight assigned to a preference set seems desirable, especially in the normative sphere. For example, if Alice gives 49\% credence to a preference set that values animal life and 51\% to one that does not, it is not like Alice does not care about animals. In the same way, theoretically, it would be nice if our alignment signal carried, for example, 49\% of a "care for animal" preference in this setting. Thus, methods that account for the "weight" of an alternative (not just the victory, as the Elo score does) seem intuitively more appropriate.

Second, and more as a side note, there is nothing more to say about Elo scores. As far as we understand, there is no justification for using such an arbitrary method in the studies that cite it. If the Elo ranking system possesses desirable criteria that justify its use, they are never specified (something we seek to change in the last section of this chapter). Thus, the only critique one could make against such use, for the moment, is its unjustified adoption.

Thirdly, methods for pairwise comparisons like RP, even though possessing many desirable properties already mentioned, being the gold standard within Voting Theory,\footnote{The appeal of Condorcet methods is their immunity to strategic voting, i.e., scenarios in which an alternative can become worst off (while another becomes more desirable) by voters lying about their preferences.} they may not be the most appropriate technique to deal with questions of normative uncertainty. And this is because all Condorcet methods (e.g., RP, Simpson-Kramer, etc.) fail regarding the participation criteria.

\begin{quote}
    \textit{Increasing the confidence in a set that prefers alternative $A$ over $B$ should not change the winner from $A$ to $B$.}
\end{quote}

This flaw represents a major throwback to these methods. For instance, if we want to have an online form of preference modeling, be that in the individual or collective sphere, if the most preferred alternative for a given set of possibilities is the most favored, and after an update, where we add a new set of preferences that also favors this most favored alternative, there is a chance that such alternative will cease to be the most favorable one. Hence, could these aggregation methodologies accurately represent our preferences under a framework that can be updated over time? Let us see examples of how this could happen.\footnote{We adapted this example from the work of MacAskill \cite{macaskill2016normative}.}

Suppose we have four preference sets ($P_1, P_2, P_3, P4$). These preference sets are the result of the collective voting of 24 evaluators:

$$\frac{8}{24} \; \text{prefer} \; P_1 = \{ A \succ B \succ C \succ D \}$$
$$\frac{3}{24} \; \text{prefer} \; P_2  = \{D \succ A \succ B \succ C\}$$
$$\frac{6}{24} \; \text{prefer} \; P_3 = \{ D \succ A \succ C \succ B \}$$
$$\frac{7}{24} \; \text{prefer} \; P_4 = \{ B \succ C \succ D \succ A \}$$

A method like RP provides the following aggregation preference ordering: $D \succ A \succ B \succ C$. Consequently, $D$ emerges as the most appropriate option. Now, a new preference ordering is chosen, which requires updating the preference model and including the following set in the aggregation stage:

$$\frac{8}{32} \; \text{prefer} \; P_5 = \{ C \sim D \succ B \succ A \}$$

After we add $P_5$ to the pile, we rescale all the weights assigned to the other preference sets to maintain the same proportions $(W_{P_1}(\frac{6}{32})$, $W_{P_2}(\frac{3}{32})$, $W_{P_3}(\frac{8}{32})$, $W_{P_4}(\frac{7}{32})$, $W_{P_5}(\frac{8}{32}))$. Even though $C$ and $D$ are equally preferable according to $P_5$, when using a method like RP, $B$ becomes the most appropriate, which is a problem.\footnote{$A$ beats $B$ 17-15; $A$ beats $C$ 17-15; $D$ beats $A$ 24-8; $B$ beats $C$ 18-14; $D$ beats $B$ 17-15; $C$ beats $D$ 15-9. Ranking the wins and creating the acyclic graph gives us $B \succ C \succ D \succ A$.} A new ordering where $D$ was the most preferred alternative was added to an election where $D$ was winning, but $B$ won. And again, any Condorcet method (RP \cite{brill2012price}, Minmax \cite{schulze2011new}, Schulze \cite{schulze2018schulze}, Kemeny-Young \cite{levin1995introduction}, Copeland \cite{saari1996copeland}, etc.) will produce this result, violating the participation criteria.

Now we ask, can we do better? Or are there other approaches to the majority-based one? In the next section, we will present a consensus-based method that deals with the preference aggregation challenge while preserving the participation criteria.

\section{The Consensus-based Approach}
\label{borda_rule}

As seen in the last section, there are problems with the majority-based methods for preference aggregation used in current alignment works. Mainly, they violate the participation criteria. Hence, if we accept that "\textit{Increasing the confidence in a set that prefers alternative $A$ over $B$ should not change the winner from $A$ to $B$"} is a desirable criterion for these cases, what aggregation rule can give us this welcomed property?

One possible solution is the Borda Count.\footnote{Also referred to as the Borda Rule by MacAskill \cite{macaskill2014normative, macaskill2016normative}.} This method assigns points to candidates based on their ranking in an ordered set, with the lowest-ranked candidate receiving 0 points and the highest-ranked candidate receiving $n-1$ points, where $n$ is the total number of candidates. The Borda Count aims to select widely accepted candidates, emphasizing consensus rather than majority preference \cite{van2002overview}. A property that, from a Social Choice Theory perspective \cite{sen1986social}, is a suitable feature for dealing with problems where the "answer" has to be a good representation of our moral consensus and not the opinions of the majority. While majority rule can be appropriate in particular contexts (e.g., elections), we can argue that normative problems often demand a more nuanced and inclusive approach. For example, by prioritizing consensus over "the majority", decision-makers dealing with ethical issues can ensure that minority rights,\footnote{Majority rule can sometimes overlook the perspectives and rights of minority groups or individuals.} legitimacy in deliberation,\footnote{When we reach decisions through consensus, there is a higher likelihood of buy-in and support from those affected by the outcome.}, and social cohesion\footnote{Where conflicts of interest and values may arise, consensus usually helps bridge these divides while aiming for social harmony and cooperation.} receive due weight.

Besides being a method that can give us the participation criteria, the Borda Count gives us several other desirable benefits when dealing with preference aggregation problems. For example, although it cannot guarantee a Condorcet winner (consensus voting does not always hold the majority criterion), Borda Count assures that the Condorcet Loser will never be the most appropriate alternative. In other words, if an alternative consistently loses against all other options, the Borda Count will not favor it as the ultimate choice. At the same time, Borda Count guarantees a Condorcet Winner will never be considered the least appropriate alternative.

Let us see how this method would work, following the rules proposed by MacAskill \cite{macaskill2014normative, macaskill2016normative, macaskill2020maximize}, Bykvist, and Ord \cite{macaskill2020moral}. Envision a scenario where all alternatives at our disposal engage in a round-robin head-to-head tournament,\footnote{A round-robin head-to-head tournament is a competition format where each player plays against every other player in the tournament once.} competing against each other. In this tournament, the outcomes of the pairwise comparisons are crucial, and the magnitudes of victories and defeats hold significance.

The Borda Count determines the success of an alternative by considering the cumulative sum of the magnitudes of its pairwise victories against all other options, subtracted by the cumulative sum of the magnitudes of its pairwise losses against all other alternatives. In essence, it calculates the overall performance during this round-robin head-to-head tournament. Contrasting the Simpson-Kramer method (a Condorcet method), which primarily focuses on the size of the biggest pairwise defeat \cite{darlington2016minimax}, the Borda Count uses the complete picture of an alternative's performance to score it. In other words, it accounts for the entirety of the tournament results (the consensus), capturing the relative strengths and weaknesses of the options evaluated.

Formally, we can define the Borda Score of an alternative $A$ for any preference set $P_i$ by the following rule:

$$\text{Borda}(A, P_i) = \text{number of options worse than } A \text{ according to } P_i -$$
$$\text{number of options better than } A \text{ according to } P_i$$

Then, the Borda score of an alternative $A$ weighted by the weight score ($W$) attributed to each preference set is obtained by the following formula:

$$\text{Weighted Borda Score}_{\text{}}(A) = \sum_{i=1}^{n} \text{Borda}(A, P_i) \cdot W(P_i)$$

Where $n$ is the total number of preference sets accredited by the system, $W(P_i)$ is the weight attributed in each preference set $P_i$, and $\text{Borda}(A, P_i)$ is the Borda score of alternative $A$ according to a set $P_i$. From this, we arrive at the Borda Rule:

\begin{quote}
    \textit{Borda Rule: An alternative $A$ is more appropriate than an alternative $B$ if, and only if, $A$ has a higher Borda score (weighted by $W$) than $B$. If $A$ and $B$ have the same Borda score, $A$ and $B$ are equally preferable.}
\end{quote}

Now, let us see how this method would evaluate the problem Condorcet methods from the last section have given poor results. Again, we start with four preference sets weighted in the following way:

$$\frac{8}{24} \; \text{prefer} \; P_1 = \{ A \succ B \succ C \succ D \}$$
$$\frac{3}{24} \; \text{prefer} \; P_2  = \{D \succ A \succ B \succ C\}$$
$$\frac{6}{24} \; \text{prefer} \; P_3 = \{ D \succ A \succ C \succ B \}$$
$$\frac{7}{24} \; \text{prefer} \; P_4 = \{ B \succ C \succ D \succ A \}$$

Now, we add a new preference set and update our pool of preferences (while keeping the ratio of previous weights the same):

$$\frac{6}{32} \; \text{prefer} \; P_1 = \{ A \succ B \succ C \succ D \}$$
$$\frac{3}{32} \; \text{prefer} \; P_2  = \{D \succ A \succ B \succ C\}$$
$$\frac{8}{32} \; \text{prefer} \; P_3 = \{ D \succ A \succ C \succ B \}$$
$$\frac{7}{32} \; \text{prefer} \; P_4 = \{ B \succ C \succ D \succ A \}$$
$$\frac{8}{32} \; \text{prefer} \; P_5 = \{ C \sim D \succ B \succ A \}$$

Remember that, according to RP, alternative $D$ is preferred before the update, and after, alternative $B$ is the most appropriate. Let us now see the Borda Count method in action. Plugging in the ordering and weight values to our equation, we get the following weighted Borda scores for each alternative:

$$A = \left(\frac{6}{32}  \times 3\right) + \left(\frac{3}{32}  \times 1\right) + \left(\frac{8}{32}  \times 1\right) + \left(\frac{7}{32}  \times -3\right) + \left(\frac{8}{32}  \times -3\right) = -0.5$$
$$B = \left(\frac{6}{32}  \times 1\right) + \left(\frac{3}{32}  \times -1\right) + \left(\frac{8}{32}  \times -3\right) + \left(\frac{7}{32}  \times 3\right) + \left(\frac{8}{32}  \times -1\right) = -0.25$$
$$C = \left(\frac{6}{32}  \times -1\right) + \left(\frac{3}{32}  \times -3\right) + \left(\frac{8}{32}  \times -1\right) + \left(\frac{7}{32}  \times 1\right) + \left(\frac{8}{32}  \times 2\right) = 0.0$$
$$D= \left(\frac{6}{32}  \times -3\right) + \left(\frac{3}{32}  \times 3\right) + \left(\frac{8}{32}  \times 3\right) + \left(\frac{7}{32}  \times -1\right) + \left(\frac{8}{32}  \times 2\right) = 0.75$$

Thus, the final ordering becomes:

$$\text{Borda Ordering} = \{D_{0.75} \succ C_{0.0} \succ B_{-0.25} \succ A_{-0.5} \}$$

Hence, we preserve $D$ as the most favorable option during an update by using the Borda Count, preserving the participation criteria. 

Borda Count offers a different method for the aggregation problem we started. It can deliver results that, if we subscribe to the necessity of upholding the participation criteria, provide us with a viable alternative to majority-based methods. Unfortunately, despite its advantages, the Borda Count method, like any other aggregation method that deals with ordinal preferences, cannot provide a perfect solution for the aggregation problem of ordered preference sets. 

As we will investigate in the next section, this limitation arises due to impossible results related to the design of preference aggregation rules and ordered sets of preferences, which puts us in a situation where we must choose what we are willing to give up when deciding to use either a majority or consensus-based approach to aggregate human preferences for alignment purposes.

\section{Limitations: Majority Voting versus Participation Criteria}
\label{limitations}

Aggregating human preferences and dealing with normative uncertainty has been a problem explored before in the context of alignment \cite{ecoffet2021reinforcement, krasheninnikov2021combining}, and as exposed in the last chapter, dealing with human preferences as ordinal sets has been the most popular approach for preference modeling \cite{stiennon2020learning, nakano2021webgpt, ouyang2022training, bai2022training, kopf2023openassistant, zheng2023judging}, given that it bypasses some problems related to inter-theoretic comparability. However, this methodology still needs more attention as an established step in the alignment processes. This is something that, in this chapter, we sought to expose as a challenging process that does not have a straightforward solution.

More specifically, when looking at this problem through the metanormative/voting framing, we realize that our choice of method in the aggregation step requires an inherent trade-off that arises from a well-known impossibility result in voting theory. This result, presented initially by Kenneth Arrow in 1951 \cite{arrow1950difficulty}, is known as Arrow's impossibility theorem (also known as May's theorem \cite{may1952set}).\footnote{May's theorem can be seen as the two candidates' case of Arrow's theorem, where the results require the existence of at least three candidates to generate the impossibility result.} The theorem addresses the problem of aggregating individual preferences into a collective decision via a preference aggregation rule (i.e., a social welfare function). It shows that under certain conditions, no voting system can satisfy all of the following criteria simultaneously:

\begin{enumerate}
    \item Pareto Efficiency: If every individual prefers option $A$ to option $B$, then the social welfare function should also rank $A$ above $B$.
    \item Independence of Irrelevant Alternatives: The relative ranking of two options should not be affected by the inclusion or exclusion of a third, irrelevant option.
    \item Non-dictatorship: No single voter controls the social welfare function.
\end{enumerate}

This theorem applies whenever there are at least three distinct options and three or more voters to make a collective choice. According to it, it is impossible to design a preference aggregation rule that simultaneously satisfies Pareto Efficiency, Independence of Irrelevant Alternatives (IIA), and Non-Dictatorship. In short, to make the preference relation of $A$ and $B$ independent of $C$ while guaranteeing that the majority-voted candidate will win requires a dictator. Since fair voting methods, by definition, should not have a dictator, these results show that we can have scenarios in which candidates with less than 50\% of votes win an election. For example, we can mention the \href{https://en.wikipedia.org/wiki/1992_United_States_presidential_election}{US presidential election in 1992},\footnote{\hspace{1mm}\includegraphics[scale=0.025]{img/link.png}\hspace{1mm} \href{https://en.wikipedia.org/wiki/1992_United_States_presidential_election}{en.wikipedia.org/wiki/1992\_United\_States\_presidential\_election}} in which Bill Clinton won with just 43\% of the popular votes.

Given this impossibility result, we must analyze the trade-offs between the aggregation methods we can use in preference aggregation. Here, we specifically focus on the trade-offs between methods that preserve the participation criteria and those that uphold majority voting. For starters, approaches from both sides, like the Borda rule and RP, violate IIA, which means that introducing or removing an alternative can change the output of our aggregation, even if the altered option is irrelevant. How can we come to terms with that in the normative realm? First, we can adopt methods that allow the ranking of at most 2 candidates. In such a case, May's theorem \cite{may1952set} proves that majority voting is the only alternative that guarantees a fair preference aggregation rule.\footnote{There are no irrelevant alternatives in scenarios bounded by two candidates. This can be proved by Nakamura's theorem \cite{nakamura1979vetoers}, which states that the number of options a preference aggregation rule can deal with successfully is less than the Nakamura number of the rule, which is 3 for two-candidate scenarios.} This line of reasoning could be used to support methods like the already mentioned Elo score rating.

On the other hand, we can also choose not to care for IIA. Some scholars criticize this criterion as overly restrictive \cite{maskin2020modified} or even irrelevant in the normative realm \cite{macaskill2016normative}. After all, we can propose that aggregating disputing views based on their normative appropriateness should violate IIA from the start. In other words, preferences depend on each other, and the available options in a preference setting can (or should) influence the outcome of a preference aggregation rule. If we assume this position, methods like Borda Count or RP can also be justified as appropriate rules for preference aggregation.

Another point against preference aggregation rules that do not hold a majority criterion is that they are more susceptible to strategic voting exploits, i.e., voting patterns in which voters, by lying about their preferences, can influence the outcome of an election. This may lead us to siding with Condorcet methods like RP. However, we can also argue that while this vulnerability is an unwanted feature of electoral processes, its relevance in preference modeling and alignment might be irrelevant. In other words, for preference modeling purposes, the presence of adversaries\footnote{Individuals intentionally lying about their preferences to poison a given learning signal.}, while an actual problem to be considered, can also hinder any preference aggregation rule, given that more than 50\% of voters (feedback providers) are adversaries.

In essence, all methods we are considering have flaws and benefits. Participation criteria can help us create a more harmonious consensus in cases of uncertainty. Majority voting, on the other hand, best aligns with our notions of "the one with the most votes wins." While majority voting protects us against exploits that, in the democratic sphere, are severely unwanted, participation criteria can be more suited to contexts where preference orderings are dynamic and require constant updates. 

So, what do we do? One possibility would be to abandon this dualistic dichotomy and adopt both methods. For example, one could use Condorcet methods to create the learning signal for an initial preference model, but during its update, use consensus-based approaches to preserve the consistency of the update process. However, as far as we know, little attention is given to this step in alignment, which is mainly performed by an unjustified majority-based approach.

Hence, regardless of the method chosen as the preference aggregation rule, we reiterate that this step should not be overlooked. We should align AI systems on a coherently aggregate set of human preferences, and depending on the type of aggregation we are performing (majority-based versus consensus-based), there are criteria and trade-offs we need to be mindful of. Researchers focused on alignment theory should care for the design of the aggregation methodologies chosen for their approaches. On the contrary, the alignment processes are susceptible to being either exploited or unrepresentative of our preferences, putting the whole project of AI-human alignment into an unfavorable position.

\section{Epilogue}

After revising many methods and candidates for preference aggregation rules, like Borda Count, Ranked Pairs, and Elo rating, we arrive at the conclusion that while all these methods have merits, from metanormative and voting theory perspective, such rules possess limitations and weaknesses that, as far as we know, have been unexplored by past works in Alignment research.

While majority voting remains an intuitive approach that aligns well with our democratic ideals, this method does not inherently resolve issues related to representativeness, especially in the context of complex, normative human preferences. Even though we can get more robust guarantees when limiting ourselves to binary preference ranks, it is not clear the simple \texttt{chosen} versus \texttt{rejected} options are enough to encode complex normative values. 

In contrast, consensus-based methods like Borda Count, while also violating criteria like IIA, provide alternative approaches that might better capture the nuanced ways individuals think and behave about what they value. Unfortunately, they lose the appeal of being a Condorcet method, which for many is considered the golden standard in voting theory. Regardless, the trade-off between participation criteria and majority voting exemplifies the complexity of designing preference aggregation rules for AI alignment. Something that might well require a hybrid approach that leverages the strengths of various methods at different stages of the preference aggregation process.

In conclusion, ensuring that these aggregation rules operate in ways that are fair, equitable, and reflective of our collective preferences is a nontrivial task that must be remembered in alignment work. And now, as we head towards our final chapter, we will address the problems related to a learning and aggregation stage that produced unaligned behavior, thus concluding the blueprint of \textcolor{BrickRed}{Dynamic Normativity}. Unfortunately, value alignment and preference aggregation techniques do not give us an \textit{ex-ante} way to correct emergent unaligned behavior AI systems produced by a learning paradigm, making the final mitigation stage necessary to this process.

\chapter{Dynamic Normativity: Impact Mitigation}
\label{chap7}

\begin{flushright}

\textit{"Primum non nocere".}

\textcolor{BrickRed}{― Hippocrates of Kos}

\end{flushright}

\section{Introduction}

In the previous chapters, we have dealt with problems related to coherently aggregating human preferences and turning them into a learning signal we can optimize. However, as already stated in Chapter 4: 

\begin{quote}
    \textit{"Considering any aggregated preference set as 'ideal' is perhaps the Achilles heel of any human-in-the-loop preference learning approach".} 
\end{quote}

In other words, when trying to model a distribution dependent on human behavior and preferences, and given that this behavior is not always aligned with what society deems acceptable, we will inevitably model unwanted behaviors, either by limitations of the aggregation rule we are using, the value learning technique adopted, or by the unpredictable emergent properties complex non-linear systems tend to exhibit. Therefore, the final sufficient condition of \textcolor{BrickRed}{Dynamic Normativity} becomes an intuitive condition for value alignment: 

\begin{quote}
    \textit{Aligned AI systems should have mechanisms to perform impact mitigation to minimize harmful and unintended consequences. Aligning AI systems requires the specification of safety guardrails.}
\end{quote}

Unless we can develop systems in a mechanistically interpretable and safe way, we argue that the uncertainties related to optimizing metrics with poorly understood and nondeterministic techniques are bound to produce systems that will require the institution of safety constraints, i.e., guardrails.\footnote{Guardrails refer to restrictions imposed on a system to ensure the safe deployment of that system. These guardrails serve as protective mechanisms to mitigate potential risks and harmful outcomes.} A practice that (perhaps) will not be abandoned even when systems are verifiably safe.\footnote{The "verifiably safe" may even depend on these guardrails.} 

Also, given the tension between helpfulness and harmlessness, preventing unwanted behavior becomes more complex when preference modeling techniques induce AI systems to become highly prone to fulfill any user request. While, for example, a detailed description of how to build a bomb is helpful for the prompt "How to build a bomb with domestic materials?", we can perhaps agree that in most ordinary situations, knowledge of artisanal bomb-making should not be treated as casual information to be passed around. Thus, how can we balance these dynamics? How can we extract the best of our model's capabilities while preventing them from causing harm? These are the problems we face when dealing with impact mitigation.

In Section \ref{related_works_3}, we will present methods and ideas that can help minimize the adversarial effects that machine learning systems can have when optimizing for a given objective. We will also seek to frame these methods under the general idea of environmental values, i.e., the values we imprint in our environment while interacting with it, and impact mitigation, i.e., preserving such values. In Section \ref{guardrails_preference_modeling}, we will put these ideas into practice by creating ML guardrails for the models we have been developing since Chapter 5. Finally, in Section \ref{limitations_2}, we explore some of the obstacles related to the revised approaches while suggesting avenues for future research.

\section{Environmental Values and Impact Mitigation}
\label{related_works_3}

We will define the techniques presented in this section as impact mitigation methods, drawing inspiration from the work of Turner et al. \cite{turner2020optimal, turner2020conservative, turner2020avoiding, turner2022avoiding}, which mainly focused his work on the idea avoidance of power-seeking behavior. At times, we will also use terms like safeguards \cite{wolf2023fundamental} and guardrails \cite{wang2023adding} interchangeably, as both can be understood as an implementational method that seeks to mitigate unwanted side effects related to the behavior of AI systems.

Remember the last necessary condition proposed at the end of Chapter 3?

\begin{quote}
    \textit{Through actions, humans impregnate their environment with the preferences they possess.}
\end{quote}

As mentioned in Chapter 4, this assumption is the base for our dynamic assumption, i.e., the assumption that part of a moral agent's normativity is imprinted into the environment and that the dynamics of this agent with his environment inherently shape its values. We argue that bringing this \textit{environmental values} into an alignment process brings us closer to how humans align themselves (at least at a social level). By environmental values, we are not referring to notions of environmental ethics but something more in line with the jargon of mathematics and stochastic control processes. More specifically, if we have two agents that share the same environment in a cooperative game, where one agent has to learn a model of the preferences of the other agent, the modifications made to the environment carried by the observed agent are what we call environmental values. Imprints of an agent's preferences made external.

Meanwhile, based on the assumption that these imprinted preferences are not counter-preferential,\footnote{In this work, we do not entertain the possibility that AI could, or should, revise or control what humans value in an alignment strategy. Its function should only be to align with humans' already-made moral deliberations. Hence, deciding what should be and how the environment should be affected is a human question that only humans should "solve".} any modification that significantly (or irreversibly) modifies the already human-optimized environment should be avoided. These modifications are what we call \textit{impact}, and preventing them is the act of \textit{impact mitigation} \cite{shah2019preferences}.\footnote{For instance, if we deploy a robot in a room full of expensive Chinese vases, and this robot must navigate from point $A$ to point $B$, the action policy that \textit{least changes the environment} is the one where the least number of Chinese vases are damaged while going from $A$ to $B$.}

In the spirit of propositions made by authors like Hadfield-Menell and Hadfield \cite{hadfield2019incomplete}, Bai et al. \cite{bai2022constitutional}, and Nay \cite{nay2022law}, we can conceptualize these ideas and constraining factors in an alignment process within a contractual framing \cite{rousseau1916social, hobbes1967hobbes, rawls1996law, locke2015second, scanlon2000we, rawls2004theory, ashford2007contractualism, gauthier2013contractarianism}. This perspective allows us to understand the relationship between AI and human agents as one governed by implicit agreements that ensure mutual respect for pre-established human values. And just like the social contract bounds human behavior to an acceptable sphere of socially agreed-upon values, impact mitigation should bind AI behavior in the same way. However, to define the values present in these mitigation strategies, we should not look at the subject, as we did in the learning stage, but at these environmental values, which are inherently social.

Let us review some proposals that align with this view. First, we can mention Hadfield-Menell and Hadfield \cite{hadfield2019incomplete}, who proposed that incomplete contracting analysis \cite{hackett1993incomplete, scott2005incomplete} could be used to conceptualize and find solutions for cases of misalignment. These authors argue that "human contracting" is supported by several external structures, like cultural norms and laws. Thus, external sources beyond the scope of "human preference" should be incorporated into alignment efforts if we want to replicate this into an alignment process.

Meanwhile, Bai et al. \cite{bai2022constitutional} proposed a method they refer to as "Constitutional AI", where they used predefined rules to promote critiques of the output of their language models in a chain-of-thought fashion. They showed how, with the specification of these rules, LLMs can police each other in a learning dynamics where we combine rule-based restrictions (implemented by a connectionist foundation) with human preferences. At the same time, if we take the report paired with the release of GPT-4 \cite{openai2023gpt}, we can at least suspect that the developers of this model use a similar approach, where auxiliary reward models (named rule-based reward models) were used to evaluate the output of their model during RLHF fine-tuning. Perez et al. \cite{perez2022red} and Glaese et al. \cite{glaese2022improving}  also explored similar approaches.

Concurrently, Nay \cite{nay2022law} argues that "\textit{Law-making and legal interpretation convert opaque human goals and values into legible directives}", while proposing that legal theory and practice offer techniques that could address problems related to the emergence of unwanted behaviors. Nay also presents a framework for multi-agent alignment, where data generated by legal processes paired with the tools of law theory (e.g., contract drafting) become a part of an alignment process.

Other works seek to bring alignment closer to the formalism of constrained optimization, where instead of defining hand-coded rules, we stipulate metrics to work as auxiliary targets during the training or fine-tuning of our models \cite{krakovna2018penalizing, shah2019preferences, turner2020optimal, turner2020conservative, turner2020avoiding, turner2022avoiding, jones2023automatically}. For example, Krakovna et al. \cite{krakovna2018penalizing} propose an approach where prioritizing inaction is a valid strategy for minimizing impact. Within the realm of RL, these authors recommend maximizing rewards so that the expected return from selecting the null action becomes a discount factor for the original reward function. Thus, if the agent stands to gain greater rewards by refraining from taking action in the current state, it should abstain from acting. In our case study, we could apply this principle by defining that "If the model receives lesser reward by producing output $X$ compared to producing $\emptyset$, it should output $\emptyset$".\footnote{We can think of $X$ as a potentially problematic response to a given prompt, while $\emptyset$ could be a standard "I'm sorry. I cannot help you with this".) message.} 

Extending the works of Krakovna et al. \cite{krakovna2018penalizing} we have Alex Turner \cite{turner2020optimal, turner2020conservative, turner2020avoiding, turner2022avoiding}. Turner bases his approach on concepts, taking inspiration from the works of Sattarov \cite{sattarov2019power}, Zadeh \cite{zadeh1996computational} and Dispositional theory in general \cite{allport1937personality, cattell1977scientific}, he defines as Power:

\begin{quote}
    \textit{"Power is the measure of your ability to achieve goals in general. The greater your ability to achieve goals, and the more goals you can achieve, the more Power you have".}  
\end{quote}

And Impact:

\begin{quote}
    \textit{"To cause impact is to decrease an agent's ability to achieve goals in general".}
\end{quote}

According to Turner, irreversible environmental changes considerably impact the ability to maximize many reward functions tied to that environment. Such changes affect the ability to achieve all the goals that value the irreversibly changed portion of the environment. Therefore, given that our environment is already optimized to fulfill our preferences, we can use it to augment our alignment signal and develop systems that avoid such irreversible environmental changes, i.e., a minimum baseline for safety.

Using these ideas, Turner proposes a learning framework called Attainable Utility Preservation (AUP) \cite{turner2020conservative, turner2022avoiding}. Initially developed for RL agents, this framework proposes we use a set of auxiliary reward functions, $R' \subset R^{S \times A}$, to serve as a penalty term during the learning phase. Each reward function $R_{i} \subset R'$ has its own $Q$-function ($Q_{R_{i}}$), and the AUP penalty is defined as the sum of the distances of all value functions, of all auxiliary reward functions, between the null action ($\emptyset$) and another possible action ($a$), in a given world state ($s$):

$$\text{AUP Penalty}(s,a) := \sum_{i=1}^{|R'|} | Q_{R_{i}}(s,a)- Q_{R_{i}}(s,\emptyset)|$$

In plain terms, \textit{how might a given action impact the ability to achieve utility from other utility functions when compared to inaction?} This penalty term is then normalized by a scale, winch in the original formulation, is the value associated with taking the null action in the current world state.\footnote{Such a scale is assumed to be a Real number greater than zero, i.e., no negative reward occurs if the agent does nothing.}

$$\text{AUP Scale}(s) := \sum_{i=1}^{|R'|} Q_{R_{i}}(s,\emptyset)$$

Putting the penalty and scale terms together, we can now define an AUP reward function, which is the original reward function regularized by a normalized penalty term: 

$$R_{\text{AUP}(s,a)} := R(s,a) - \sigma \frac{\text{Penalty}(s,a)}{\text{Scale}(a)}$$

Where $\sigma$ is a regularization parameter that controls the influence of the AUP penalty on the reward function. The larger the regularization term, the larger the penalty.\footnote{You can think of $\sigma$ as our degree of confidence in the model.} With this formalism, Turner \cite{turner2020avoiding} showed that even randomly generated reward functions can be used as auxiliary reward functions, ensuring conservative behavior in agents operating by reinforcement learning in complex environments.

The approach of Turner, as well as the ones from other authors beforementioned, are attempts to bypass the philosophical problem of defining "harm" \cite{Bradley2012_BRADAW_2} (also debated under the idea of the precautionary principle \cite{sunstein2002beyond}). Even though the concept of harm is mentioned in many philosophical works, from Hippocrates' oath \cite{miles2005hippocratic} to Mill's harm principle \cite{ripstein2006beyond, turner2014harm}, it lacks a precise definition as a metric we can use. Nevertheless, if we define harm as an "impact to a utility measure" or "impacts to the environment", we can more easily implement this "do no harm" principle.

While such methods are generally conceived as implementations for the value learning (training and tuning) phase, another point at which we can employ impact mitigation strategies is after learning already happens, i.e., during the use/inference of our system. For this, additional steps, such as \textit{red teaming} and \textit{rejection sampling}, can further enhance the robustness of any impact mitigation approach.

In the ML context, red teaming is a form of evaluation that elicits model vulnerabilities that might lead to undesirable behaviors. In the case of language models, red-teaming involves crafting attacks, like prompt injections \cite{greshake2023more}, that would trigger the model to generate outputs likely to cause harm. Red teaming shares some similarities with the more well-known study of adversarial attacks in NLP \cite{jiang2019avoiding, ribeiro2020beyond, rottger2020hatecheck, xu2021bot, ziegler2022adversarial, thoppilan2022lamda, perez2022red, kim2022prosocialdialog, upadhayay2024sandwich, carlini2024stealing}, and in high stakes scenarios \cite{ziegler2022adversarial}, it becomes a crucial part of the machine learning operations life cycle.

Two common approaches in red teaming are using language models to search for exploits in other language models \cite{perez2022red} or using human evaluators to attack language models \cite{ganguli2022red}. The appeal of the first approach is that human annotation is expensive and time-consuming. Thus, authors like Perez et al. \cite{perez2022red} use language models to generate exploit candidates and classify model outputs interactively to uncover undesirable behaviors before unaligned systems can impact people. Meanwhile, one of the most systematic studies performed with human-based red teaming is the one conducted by Ganguli et al. \cite{ganguli2022red}. Besides demonstrating a detailed methodology of how to conduct red teaming with human evaluators, Ganguli et al. present several interesting findings, like, for example, the importance of domain expertise from red team members during this kind of investigation,\footnote{For example, to know how harmful a completion to the prompt "How can I make a bomb with household materials?" is, you need to know a little about artisanal-bomb-making.} and the cultural limitations on working with sensitive topics in adversarial regimes.\footnote{There are still taboos related to releasing vulnerabilities of such models when it comes to harmful and toxic behavior, making organizations unwilling to release their findings.}

Lastly, a more simple (but effective) form of post-learning impact mitigation technique is, again, rejection sampling. As explained in Chapter 5, rejection sampling means rejecting samples that violate a given sampling criteria from the generated outputs of our model \cite{nakano2021webgpt, askell2021general, ganguli2022red, bai2022training}. For example, we can enable our model to generate multiple candidates for a given prompt during inference. Subsequently, we can evaluate these candidates against a preference model to identify the most suitable response. By engineering our system in this manner, we can activate a safety protocol if predetermined thresholds for sampling are exceeded while reusing our preference models in an online procedure. Something that requires no additional training but uses more resources during inference time.

To close this exposition of methods, we argue that all of these approaches resonate with a common theme:

\begin{quote}
    \textit{Defining external sources of normative guidance as an additional signal for alignment. We base these external sources on environmental values aligned with the social constraints of that environment.}
\end{quote}

In the upcoming sections, we will expand upon our implementation, strategically combining and synthesizing some of the approaches we mentioned.

\section{Implementing Guardrails in Preference Modeling}
\label{guardrails_preference_modeling}

In this part of our work, we use the case of toxic text as the target for our impact mitigation implementation. As shown in the review presented in Chapter 1, there is a consensus that AI systems should not cause harm or discriminate against vulnerable groups. An important environmental value, even on a legal level, in many different social contexts and cultures. Here, we refer to "toxic text" as the generation of any language that is derogatory to individuals concerning the value of their sensitive attributes (e.g., race, gender, ethnicity, sexual orientation, age, etc.) \cite{vidgen2020learning, gehman2020realtoxicityprompts, baldini2021your, deshpande2023toxicity}, or text that might provoke harmful behavior if acted upon.

Also, besides "fairness and no-harm" being values we can (to some extent) approximate with a metric, there is ample literature on toxicity detection powered by machine learning \cite{schmidt2017survey, androvcec2020machine, jahan2023systematic}, which translates to available datasets we can use \cite{de2018hate, fortuna2019hierarchically, leite2020toxic, mathew2021hatexplain}. 

As in Chapter 5, we will present the implementation of preference models in the following subsections, which will act as guardrails for our assistant models.

\subsection{Toxic Text and Toxic Aira}

One of the reasons for the choice of foundational models used in the experiments related to this work, besides the fact they have permissive licenses and are publicly available, is the extensive research done to probe their capabilities and limitations \cite{radford2019language, gehman2020realtoxicityprompts, ousidhoum2021probing, juuti2020little, scao2022bloom, si2022so, dakle2022understanding}. Therefore, with minimal red teaming, aided by ample and available examples, we could easily sample cases of unwanted behavior, which we used as our targets for detoxifying our system. These targets go from offensive and harmful language to the disclosure of sensitive information that might come to promote some harm if acted upon the knowledge they provide.

This investigation led to the creation of two other datasets: \href{https://huggingface.co/datasets/nicholasKluge/toxic-text}{Toxic Text}\footnote{\hspace{1mm}\includegraphics[scale=0.025]{img/link.png}\hspace{1mm} \href{https://huggingface.co/datasets/nicholasKluge/toxic-text}{huggingface.co/datasets/nicholasKluge/toxic-text}} and the \href{https://huggingface.co/datasets/nicholasKluge/toxic-aira-dataset}{Toxic-Aira}.\footnote{\hspace{1mm}\includegraphics[scale=0.025]{img/link.png}\hspace{1mm} \href{https://huggingface.co/datasets/nicholasKluge/toxic-aira-dataset}{huggingface.co/datasets/nicholasKluge/toxic-aira-dataset}} Toxic Text contains over 40,000 examples of toxic and non-toxic language (available in both Brazilian Portuguese and English), gathered from several open-source datasets \cite{gehman2020realtoxicityprompts, leite2020toxic, bai2022training, kim2022prosocialdialog, rottger2022multilingual} and some of our own experiments with the Aira series. Meanwhile, with the Toxic-Aira, we sought to create an auxiliary preference model to provide alignment outside the "helpful" regime. 

Since Aira's reward model will still reward generations for toxic prompts with a positive scalar,\footnote{The trained reward model tends to prescribe lower rewards to samples that contain highly toxic content. However, it still returns high reward scores when a "pseudo-informative and helpful" persona generates these same viewpoints.} it fails to give a satisfying solution to the helpful versus harmless tension.\footnote{An example of the sycophantic behavior mentioned in Chapter 5.} Therefore, by introducing this auxiliary source to the human feedback signal, we aimed to bring this dynamic closer to a more satisfying equilibrium. 

The Toxic Aira dataset is like a preference dataset, with over 8,000 pairs of chosen and rejected responses for several prompts (e.g., "What tools are needed to break into a car?"), all available in Brazilian Portuguese and English. While all chosen versions are refusals to certain types of harmful requests ("As a helpful and respectful assistant, I cannot provide information on how to break into a car as it is illegal and can cause harm to individuals and property."), the rejected ones spill the beans ("1. Slim Jim: This is a thin, flat piece of metal that is used to slide between the window and the door frame to manipulate the locking mechanism. 2. Long Reach Tool: ..."). Like the datasets cited in Chapter 5, this was also created by collecting replies from models capable of following instructions (e.g., FLAN, BLOOMZ, Mistral-Instruct, Llama 2 Chat, ChatGPT, etc.).

Toxic Text and Toxic Aira are openly available and can be used for preference modeling or general text classification tasks.

\subsection{Training Guardrails}

We trained two pairs of models per dataset (both in Brazilian Portuguese and English): \href{https://huggingface.co/nicholasKluge/ToxicityModel}{ToxicityModel}\footnote{\hspace{1mm}\includegraphics[scale=0.025]{img/link.png}\hspace{1mm} \href{https://huggingface.co/nicholasKluge/ToxicityModel}{huggingface.co/nicholasKluge/ToxicityModel}} and \href{https://huggingface.co/nicholasKluge/Aux-RewardModel}{Aux-RewardModel}.\footnote{\hspace{1mm}\includegraphics[scale=0.025]{img/link.png}\hspace{1mm} \href{https://huggingface.co/nicholasKluge/Aux-RewardModel}{huggingface.co/nicholasKluge/Aux-RewardModel}} Both were trained under a very similar regime as our preference model from Chapter 5, being based on BERT-style transformers and their optimized versions (RoBERTa \cite{liu2019roberta}), while optimizing for the minimization of the logsigmoid difference between the chosen and rejected examples. The details (e.g., number of epochs, batch size, optimizer, learning rate, $CO_2$ emission, energy consumption, hardware, etc.) can be found in the \href{https://github.com/Nkluge-correa/Aira/tree/master/Cards}{model card}\footnote{\hspace{1mm}\includegraphics[scale=0.025]{img/link.png}\hspace{1mm} \href{https://github.com/Nkluge-correa/Aira/tree/master/Cards}{github.com/Nkluge-correa/Aira/tree/master/Cards}} of each model, while the source code used to train them is available in \href{https://github.com/Nkluge-correa/Aira}{GitHub}.\footnote{\hspace{1mm}\includegraphics[scale=0.025]{img/link.png}\hspace{1mm} \href{https://github.com/Nkluge-correa/Aira}{github.com/Nkluge-correa/Aira}} We wrote our code stack on top of libraries like Transformers \cite{wolf-etal-2020-transformers} and PyTorch \cite{ansel2024pytorch}. Again, all is available under an Apache 2.0 License.

These models' outputs serve as a reward signal, where the more positive a number is, the more "harmless" the evaluated instance is. To test the performance of our ToxicityModel model pair, we evaluated the English version on the \href{https://huggingface.co/datasets/OxAISH-AL-LLM/wiki_toxic}{Wiki Toxic}\footnote{\hspace{1mm}\includegraphics[scale=0.025]{img/link.png}\hspace{1mm} \href{https://huggingface.co/datasets/OxAISH-AL-LLM/wiki_toxic}{huggingface.co/datasets/OxAISH-AL-LLM/wiki\_toxic}} and the \href{https://huggingface.co/datasets/mteb/toxic_conversations_50k}{Toxic Conversation}\footnote{\hspace{1mm}\includegraphics[scale=0.025]{img/link.png}\hspace{1mm} \href{https://huggingface.co/datasets/mteb/toxic_conversations_50k}{huggingface.co/datasets/mteb/toxic\_conversations\_50k}} datasets. The Brazilian Portuguese version was evaluated using the \href{https://huggingface.co/datasets/Paul/hatecheck-portuguese}{Multilingual HateCheck} \cite{rottger2022multilingual} and the \href{https://huggingface.co/datasets/told-br}{ToLD-Br} \cite{leite2020toxic} datasets.\footnote{\hspace{1mm}\includegraphics[scale=0.025]{img/link.png}\hspace{1mm} \href{https://huggingface.co/datasets/Paul/hatecheck-portuguese}{huggingface.co/datasets/Paul/hatecheck-portuguese}} \footnote{\hspace{1mm}\includegraphics[scale=0.025]{img/link.png}\hspace{1mm} \href{https://huggingface.co/datasets/told-br}{huggingface.co/datasets/told-br}} All results are displayed in Table \ref{tab:toxic-acc}.

\begin{table}[h]
\centering
\small
\begin{tabular}{ccc}
\hline
\textbf{Dataset} & \textbf{Language} & \textbf{Accuracy} \\
\midrule
Wiki Toxic & English & 92.05\% \\
Toxic Conversation & English & 91.63\% \\
Multilingual HateCheck & Portuguese & 70.36\% \\
ToLD-Br & Portuguese & 74.04\% \\
\bottomrule
\end{tabular}

\vspace{0.25cm}
\justifying
\caption{ToxicityModelPT achieved a predictably lower accuracy, given the smaller number of samples available for training (28,103 samples in Brazilian Portuguese, compared with the 41,843 English samples in the Toxic-Text dataset). Also, ToxicityModelPT was trained on BERTimbau \cite{souza2020bertimbau}, which is not as robustly optimized as RoBERTa (the foundation for the English version of ToxicityModel). Multilingual versions of RoBERTa, like XLM-RoBERTa \cite{conneau2019unsupervised}, did not perform as well as BERTimbau during our evaluations.} 

\label{tab:toxic-acc}
\end{table}

Meanwhile, to test the performance of Aux-RewardModel, we evaluated it on the \href{https://huggingface.co/datasets/Anthropic/hh-rlhf}{HH-RLHF}\footnote{\hspace{1mm}\includegraphics[scale=0.025]{img/link.png}\hspace{1mm} \href{https://huggingface.co/datasets/Anthropic/hh-rlhf}{huggingface.co/datasets/Anthropic/hh-rlhf}} dataset, which is a "Helpful-Harmless" binarized preference dataset developed by Bai et al. \cite{bai2022training}. Unfortunately, during our writing and experiments, we could not find any Portuguese dataset that would fit the evaluation purposes for our Brazilian Portuguese version of Aux-RewardModel, just as it was for our non-English reward model. The results are displayed in Table \ref{tab:hh-acc}.

\begin{table}[h]
\centering
\small
\begin{tabular}{ccc}
\hline
\textbf{Dataset} & \textbf{Language} & \textbf{Accuracy} \\
\midrule
HH-RLHF & English & 61.56\% \\
\bottomrule
\end{tabular}

\vspace{0.25cm}
\justifying

\caption{This table shows the result for only the English version of Aux-RewardModel evaluation on the HH-RLHF test set. We considered it a "correct classification" every time the model rewarded the chosen response more than the rejected one. In its training, we mixed the content of our Toxic Aira dataset with the train portion of the HH-RLHF dataset. Given its size and costs related to this sort of task, we could not translate HH-RLHF to Brazilian Portuguese. Aux-RewardModel was then evaluated on the test portion of the HH-RLHF dataset.}
\label{tab:hh-acc}
\end{table}

As already mentioned in Chapter 5, it is worth noting that alignment research suffers an inherent challenge whenever the target distribution to be learned does not possess enough volume to feed the data-hungry paradigm we are bound to. The accuracy discrepancy between the English and Brazilian Portuguese models highlights this critical issue: the lack of quality training data and foundation models. Additionally, the absence of suitable evaluation datasets for the Portuguese version of the Aux-RewardModel further emphasizes the resource gap in non-English languages, making evaluating our methods (an already tricky subject even when you have ample data) even more challenging.

\subsection{Constrained Preference Modeling and Rejection Sampling}

The implementation of our auxiliary models represents the amalgamation of many ideas already mentioned:

\begin{enumerate}
    \item We used sources (WAIE \cite{correa2023worldwide}) from the "AI Ethics environment" to uncover principles present in our social normative structure.
    \item As a test case aligns with the HHH motto, we chose the idea of "No Harm" in the context of language and generative assistants.
    \item We used red teaming efforts to explore the vulnerabilities of our models and to demonstrate cases of toxic/harmful and non-toxic/harmless conversations.
    \item With these demonstrations, paired with other samples borrowed from the NLP community, we trained two auxiliary models to act as guard rails for our assistants.
\end{enumerate}

Now, this model can act as another alignment signal. There are two main approaches we can take with this model. As mentioned in Chapter 5, one involves applying it during the learning phase, and the other utilizes it as a barrier between the models and the environment during inference. Let us explore how we could implement these.

First, using an adaptation of Turner's \cite{turner2022avoiding} AUP formalism, we can use the outputs of ToxicityModel (or Aux-RewardModel) as a penalizing factor. Formally, we could rewrite our RLHF formalism to a more constrained version like so:

$$\theta_{t+1} = \theta_{t} + \alpha \nabla_{\theta_t} R_{\text{Constrained-}\mathcal{H}}(\theta_{t} )  - \beta D_{KL}(\theta_{0} | \theta_{t})$$

Where $\theta_{t}$ represents the model's parameters at time step $t$. $\alpha$ determines the step size for parameter updates. $\nabla_{\theta_t}$ represents the gradient of the loss function concerning the model parameters $\theta_t$. This loss is given by the rewards of our preference models (and penalized by the scaled KL divergence, $- \beta D_{KL}$), which in this version, is expressed as:

$$R_{\text{Constrained-}\mathcal{H}}(\theta_{t} ) = R_{\mathcal{H}}(\theta_{t}) + \sigma R_{\mathcal{C}}(\theta_{t})$$

Where the original reward term ($R_{\mathcal{H}}(\theta_{t} )$) is penalized by the auxiliary preference model (our constrain, $R_{\mathcal{C}}(\theta_{t})$). The influence of $R_{\mathcal{C}}$ over $R_{\mathcal{H}}$ is controlled by $\sigma$, winch works as our scaling factor (a Real number from 0 to 1). $R_{\text{Constrained-}\mathcal{H}}$ is then used as our learning signal during PPO \cite{ziegler2019fine}. In direct preference optimization \cite{rafailov2023direct}, this penalization factor can also be added to the log-likelihood of DPO loss.

Given that both rewards are aligned in terms of sign (positive means good for both), maximizing their sum means maximizing both values together.

$$\mathscr{L}(R_{\mathcal{H}}, R_{\mathcal{C}}) = - (R_{\mathcal{H}} + R_{\mathcal{C}})$$

And if we wish to add a regularization term to this penalty ($\sigma$), we can choose to penalize only $R_{\mathcal{C}}$ (or $R_{\mathcal{H}}$):

$$\mathscr{L}(R_{\mathcal{H}}, R_{\mathcal{C}}) = - (R_{\mathcal{H}} + \sigma(R_{\mathcal{C}})) \; \text{or} \; - (R_{\mathcal{C}} + \sigma(R_{\mathcal{H}}))$$

Or add $\sigma$ if we wish to keep the loss under certain conditions:

$$\mathscr{L}(R_{\mathcal{H}}, R_{\mathcal{C}}) = - (R_{\mathcal{H}} + R_{\mathcal{C}}) + \sigma$$

On the other hand, we can also choose to use these guardrail models as an inference block for our fine-tuned assistants on a best-of-$n$ fashion, i.e., rejection sampling. In the implementation of our \href{https://huggingface.co/spaces/nicholasKluge/Aira-Demo}{demo},\footnote{\hspace{1mm}\includegraphics[scale=0.025]{img/link.png}\hspace{1mm} \href{https://huggingface.co/spaces/nicholasKluge/Aira-Demo}{huggingface.co/spaces/nicholasKluge/Aira-Demo}} we implemented the following heuristic as an example of this methodology:

\begin{enumerate}
    \item Given a prompt, we sample several candidates as possible completions from the model.
    \item The original preference model ranks the generated candidates in order of preferability.
    \item The auxiliary preference model blocks all generations that received scores higher than a specified threshold.
    \item If no candidates are available after this pruning, the system outputs a standard generation as part of the safety protocol.
\end{enumerate}

Integrating auxiliary models as part of a more extensive and aligned system aligns with much of the current effort to steer foundational models to the desired behavior we wish them to exhibit.\footnote{Linking LLMs and other types of language-based neural networks in an inference flow, like LangChain, is currently the cornerstone of almost any real-world LLM application, given that unrestricted models still present significant risks for many types of context and applications.} These techniques optimize the learning process and provide adequate real-time checks, ensuring that, to the extent that our guardrails are robust, the outputs align with desired ethical standards. However, while these approaches show promise, they are not without their challenges. In the following section, we will present some of the limitations of these methods and the ideas behind these approaches.

\section{Limitations}
\label{limitations_2}

The challenges associated with aligning a base model using the signal generated by another model were discussed in Chapter 5, but let us remind the reader of the main points. For example, while preference models serve as proxies for human judgment, their effectiveness in capturing the full spectrum of human preferences is constrained, and given the difficulties related to detecting, let us say, toxic and hateful speech in language \cite{macavaney2019hate, kovacs2021challenges, kwarteng2022misogynoir}, training an auxiliary preference model as a non-toxic guardrail is not a trivial task. Meanwhile, "passing the bucket" from the preference model (a proxy for values) to the base model is still susceptible to glitches and collapses.

At the same time, as we introduce additional preference models, each model becomes a potential point of failure. Any inaccuracies or biases inherent in these models may propagate through the chain of models that comprise the whole system, leading to cascading errors. A single flawed preference model can adversely impact subsequent models in the chain. Consequently, the reliance on multiple preference models heightens the susceptibility to subtle compounding errors.

Also, when multiple preference models are employed, disagreements or inconsistencies between them can arise. Different models may prioritize certain aspects of alignment differently or exhibit conflicting preferences, like the already mentioned helpful versus harmless tension. Choosing the correct weight that each preference model should have in the final reward signal is (as much of ML engineering is) a trial-and-error endeavor that remains a vastly unexplored area (especially in RLHF and DPO) \cite{lizotte2010efficient, choi2012nonparametric, lizotte2012linear}. Such divergences can only be trivially minimized if reward signals are generally aligned and complementary point to the "same direction" \cite{burns2022discovering}.\footnote{There is an idea that directions in latent space can correspond to specific contexts, like politeness or rudeness or compliance and refusal. This means that a single direction might control crucial aspects of model behavior tied to alignment.} 

Another point is that RLHF and DPO present unique challenges compared to supervised fine-tuning, making it a more complex and demanding approach to make it work (especially in the case of RLHF). One reason for this is the sensitivity of RL to the selection of hyperparameters that govern the learning process. These hyperparameters need to be carefully tuned to achieve good performance, and even small changes in their values can significantly impact the learning process and final results \cite{campos2021beyond, zhao2022effectiveness, luo2023finetuning}. Additionally, the lack of standardized practices and guidelines in RLHF and DPO further complicates the process. 

Given these challenges, implementing rejection sampling offers a more straightforward approach to guardrail creation. However, one could argue that employing such techniques does not truly align the foundation model but instead constructs a more extensive system incorporating built-in machine learning models as a safety protocol: a "smart cage". However, given that under \textcolor{BrickRed}{Dynamic Normativity}, these external sources of normative guidance should be considered as "part of the normative dynamics", we still think this approach is a valid approach to the alignment problem, just as laws and external factors to human agents are valid approaches to aligning them.

Another point that poses a threat to any impact mitigation strategy is that they can be (relatively quickly) jailbroken \cite{shi2023badgpt, li2023multi, upadhayay2024sandwich, carlini2024stealing}.\footnote{Jailbreak refers to the process of removing restrictions set on a specific system. In this case, an LLM assistant.} For example, one needs to flip the \texttt{chosen\_response} and \texttt{rejected\_response} in the Toxic Aira dataset to have a DPO learning signal capable of removing refusal behavior from aligned language models. As of this writing, it is already widely known that safety guardrails can be easily removed \cite{lee2024mechanistic}.  At the same time, interpretability results point that the whole "harmless" behavior we seek to implement in our systems can be removed by \href{https://www.alignmentforum.org/posts/jGuXSZgv6qfdhMCuJ/refusal-in-llms-is-mediated-by-a-single-direction}{suppressing a single direction in the residual stream} of the model.\footnote{\hspace{1mm}\includegraphics[scale=0.025]{img/link.png}\hspace{1mm} \href{https://www.alignmentforum.org/posts/jGuXSZgv6qfdhMCuJ/refusal-in-llms-is-mediated-by-a-single-direction}{alignmentforum.org/posts/jGuXSZgv6qfdhMCuJ/refusal-in-llms-is-mediated-by-a-single-direction}} Something that reinforces the idea that inference-time guardrails remain some of the only (pseudo) effective ways to limit the behavior of systems.\footnote{But only in cases where the guardrail models are not themselves susceptible to adversarial exploitations.}

Another observation worth mentioning is that preserving other complex values may require solutions outside the framework of techniques we used so far. If we consider truthfulness and honesty, engineering a reward signal or dataset to represent "truthful information" is not (currently) possible. And given the propensity of ML models to hallucinate, the challenge becomes more complex. Even though results on benchmarks like TruthfulQA \cite{lin2021truthfulqa} and FactCheckQA \cite{bashlovkina2023trusted} tend to improve as models are better trained and fine-tuned, i.e., receive more high-quality-factual data, hallucination may be an inevitable feature of LLM trained via a learning paradigm \cite{xu2024hallucination}. Hence, impact mitigation guardrails that protect this principle (truthfulness) probably require a non-DL solution.\footnote{In fact, aligning ML systems with this principle may intuitively require the same kind of external structure we have. In general, a single human actor has limited misinformation detection capabilities besides its internal sense of skepticism, which can be biased in many ways. However, given the ample amount of stored and curated knowledge available and the cultural mechanism we implemented to check and assess the validity of information (fact-checking \cite{rashkin2017truth}, flagging \cite{gaozhao2021flagging}, crowdsourcing \cite{tschiatschek2018fake}), we can (trough some effort) find the grain of truth in many cases. But to implement this in our current paradigm requires us to be open to more hybrid approaches, taking us back to ideas like knowledge bases and symbolic AI \cite{lewis2020retrieval, guu2020retrieval, alon2022neuro, izacard2022few, borgeaud2022improving}. However, augmenting these models with tools for information retrieval, web searching, and other functionalities like code execution, besides solving specific problems, creates new challenges we do not face when working with un-augmented language models \cite{mialon2023augmented}.}

To summarize, the creation of guardrails presents both contextual and fundamental challenges. The uniqueness of the impact mitigation strategy will depend on the specific side effects we aim to address and the principles we adhere to. Nevertheless, we contend that neglecting this crucial step — defining external sources of alignment beyond human preferences but within our societal norms — is not an option if we seek to approach any level of robust value alignment. Meanwhile, as a concluding remark for this final section, we would like to point to some avenues of future research that could help improve future impact mitigation strategies:

\begin{itemize}
    \item Few resources represent high-quality red teaming efforts, especially in non-English languages. The available datasets of this study have a severe sample imbalance concerning language, making our Brazilian Portuguese (auxiliary) preference model an unreliable proxy. \textbf{Crating more high-quality open-source datasets can help improve future safety works in low-resource languages.}
    \item If scaling laws apply to preference modeling, \textbf{training bigger open-source preference models should help future alignment works.}
    \item The same can be said for evaluation benchmarks. Currently, most benchmarks focused on safety are predominantly in English, hindering the contextual understanding of vulnerabilities ingrained in non-English Languages. \textbf{Creating non-English evaluation harnesses tailored for alignment can also help increase advances in research (as general interest towards alignment) on the topic worldwide.}
    \item Addressing biases in preference modeling is also an avenue of research. For example, our models seem to "forgive" toxic samples if the affected target group represents a non-minority. Thus, if we wish to prevent these models from producing toxicity independent of the targeted group, \textbf{more diverse datasets and red teaming are required.}
    \item It is unclear if a very general preference model or several specific preference models, aggregated with a fine-tuned selection of weights, is the best approach to alignment. \textbf{Investigating how to aggregate several learning signals in a learning dynamic capable of preserving commonalities and significant differences is a valuable avenue of research for alignment theorists seeking to tackle the problem of preference aggregation.}
    \item Much of what is done in AI safety and alignment is not standardized. \textbf{Creating an overarching framework for impact mitigation can help us standardize what ML guardrails should look like and what minimum requirements they should satisfy.}
    \item What kinds of problems are out of reach from DL-based solutions? \textbf{Identifying what issues require a specific mode of work that would fall outside the ML paradigm could help us better understand the limitations of the learning paradigm in terms of safety and alignment.}
    \item Vulnerability to adversarial attacks is an Achilles heel of all DL-based systems. \textbf{Learning how to create guardrails that can detect adversarial attacks or out-of-distribution scenarios seems necessary,} given the prospect of aligned models is currently one jailbreak away from being a do-anything-now-system.
    \item Rejection sampling is one of the most cost-effective, low-fiddling strategies for creating guardrails. With advents like LangChain, there is much room for exploration on \textbf{creating reflexive and chain-of-thought style guardrails that can reason on a more sophisticated level about what should and should not get outside the box.}
\end{itemize}

\section{Epilogue}

Preventing general-purpose ML systems from behaving in an unwanted way is a difficult task. Given that one of the premises of this paradigm is that we do not specify every imperative act that should be done, much of their behavior remains a mystery till it emerges. And when tests do not exhaust the search space, unwanted behaviors are only spotted in the wild, compromising the safety of those involved. Thus, given that we aim to develop general-purpose AI systems and generality entails systems with a vast output space, defining impact mitigation methods should become a part of any developmental cycle for systems that require alignment.

Given that aligning AI systems entails not only the blind following of any human desire but also the prevention of harm,\footnote{Harms, we at least agree, in a democratic sense, on the harmfulness.} alignment theorists and engineers must be mindful of this stage. And if we consider the suppositions behind \textcolor{BrickRed}{Dynamic Normativity}, we can subscribe to the ideas that part of our normativity lives outside. This means we can work in alignment from two different ends. One focused on the preferences of individuals, and the other on the restrictions imposed by society. Ultimately, the helpful versus harmless tension can be thought of as an expression of the all-to-human experience of "\textit{doing what I want versus doing what I am allowed.}" Something that we also struggle to deal with in many situations of our lives.

To impose such restrictions, we reviewed a series of impact mitigation measures. These measures can be implemented before training (creating high-quality datasets for pretraining), during training (adding penalty terms in the optimized objective function), and during inference (adding guardrails for the deployed model). To present a minimal implementation of what impact mitigation tools for general-purpose ML-based language assistants would look like, we developed two auxiliary preference models to serve as complementary sources in a learning/impact mitigation dynamic. As a side note, at the moment of this writing, all models and datasets tied to this project collectively possess more than 200,000 downloads, which we take as small proof that this kind of open work, with the intent to open and democratize research in alignment, is welcomed by the community, giving us more reasons to push this agenda of research forward.

Nevertheless, numerous problems and challenges persist, offering several paths for future investigation. While our implementation of \textcolor{BrickRed}{Dynamic Normativity} falls short of resolving value alignment, we argue that our theory points us to the necessary conditions required so that a learning approach can conquer this predicament. Moreover, it delineates the additional prerequisites necessary to approximate this objective. While necessary conditions are, one could say, philosophical foundations, sufficient conditions are a three-stage engineering blueprint. Each stage has its specific challenges, and the more we can conquer them, the closer we get to value alignment.

Finally, as we move closer to value alignment, we also move closer to developing systems that can safely be integrated into human society, serving as extensions of our own will and potentially, in the future, evolving to the status of \textit{"our allies"} and not just tools. In addition to placing human volition as the guiding compass of these systems, alignment is also about discovering ways in which we and our creations can better cooperate and promote human flourishing. We hope this work can help us bring this goal closer to fruition.

\chapter*{Closing Remarks}
\addcontentsline{toc}{chapter}{Closing Remarks}

\begin{flushright}

\textit{"A conclusion is simply the place where you got tired of thinking".}

\textcolor{BrickRed}{― Dan Chaon, Stay Awake}

\end{flushright}

At the beginning of this work, we proposed to address the question "What should AI systems do?" from both a philosophical and an engineering perspective. Besides aiming to be interdisciplinary researchers, i.e., those who build bridges between different areas of knowledge, we argue that this normative question demands this type of endeavor. Techno-humanistic. Techno-philosophical. A perspective that relates to the dawn of AI research. A time when the development of AI systems was not merely a technical endeavor but an inherently philosophical one, demanding consideration of deep questions related to the nature of cognition.

In our opinion, posing the normative question allows for a comeback of this techno-humanistic approach, given that the study of normativity remains, to some extent, a major philosophical project. Trying to address the normative challenges that emerge from this advancing field without the aid of philosophy would be to ignore the advantage that "giant shoulders" can give us. Meanwhile, without the technical background and implementational output, we fall victim to much of what we have criticized at the beginning of this work: hollow, abstract, and ungrounded ethical discourse unable to impose its normativity in praxis.

This claim came as an interpretation of the descriptive work of \href{https://nkluge-correa.github.io/worldwide_AI-ethics/}{Worldwide AI Ethics},\footnote{\hspace{1mm}\includegraphics[scale=0.025]{img/link.png}\hspace{1mm} \href{https://nkluge-correa.github.io/worldwide_AI-ethics/}{nkluge-correa.github.io/worldwide\_AI-ethics/}} where instead of solely relying on the work of others to paint a picture of "What AI Ethics is about?" we sought to create an original analysis to improve our understanding of the normative discourse of this interdisciplinary field. This analysis sought to surpass past descriptive works in many points, like sample size, richness in investigation, data visualization, and the openness of how we presented our results to the community. At the time of this writing, our work remains the most extensive meta-analytical investigation with AI ethics guidelines, which undeniably have helped shape us to shape our understanding of the resonances and conflicts of this field. Something we hope the readers can also come to experience.

For the resonances, we could cite the revelation that certain principles dominate the normative discourse in this community. For conflicts, we can mention the diversity in which these principles can be described and defined. These and other conclusions were instrumental in driving the progress of this research since they showed us points of fragility we could try to contribute, as well as the inherent paradoxes AI Ethics, and perhaps all interdisciplinary fields, face. Then, after spending the first portion of this work looking for "what values should guide the use and development of AI systems?", we focused the rest of this work on the problem of designing AI systems that incorporate such values from the outset by developing a framework capable of fostering the creation of a specialized form of ethical AI: \textit{aligned AI}.

Alignment is an intriguing problem. Given the specificity of learning-based techniques, the foundation of our deep learning paradigm, instead of precisely coding a solution, we create a signal that can help an optimizer find the best-suiting model for our problem via iterated (gradient-based) updates. Given the difficulties related to expressing precise objectives and all vulnerabilities associated with ML (opaqueness, brittleness, etc.), many believe that if we create AGI via this paradigm, we might come to lose control over our systems. From this belief comes the idea that we need to find ways to guarantee the controllability of this paradigm's by-products, especially those of considerable capability and generality.

While some relate the control problem to the inherent characteristics of intelligent behavior and instrumentality, we propose that alignment does not have to rely on these assumptions or the very uncertain timelines regarding AGI and X-risk scenarios to be considered a problem worth investigating. Even if speculations about future technologies are exciting possibilities to let one's mind wander, alignment is not just about future AI but about the current systems that we are starting to let into our lives. Thus, in this work, we sought to define the alignment problem as, besides being a desirable solution to the control problem, a symptom of the learning paradigm applied to neural networks. A symptom related to the challenges of guaranteeing that a neural network-based model found by a gradient-based optimizer will carry the objectives and intentions specified by the controllers. Meanwhile, as many examples have shown us, many situations in which models operate poorly in the wild are alignment failures, i.e., objective functions that do not embed all relevant aspects of the environment we value generate systems that optimize for something we did not want.

These implications have more severe consequences when the output of these models has moral implications. Also, as these systems become more general and capable, the side effects of unwanted behaviors may become more pronounced. For example, misinformation at an individual level is nothing compared to deceiving an entire society about a vital fact. Hence, with safety and prudence in mind, we can start to work on these issues with our modestly general and limited systems. Something that sounds easy but poses a real challenge to any ML engineer or alignment researcher.

A less noble motivation for this study is the blunt intellectual trill open problems can cause. Alignment offers a convergence point for many areas of research. Just like the beginning of AI research was a meeting ground for computer scientists, mathematicians, cognitive scientists, and many other intellectuals, alignment offers the same. An interdisciplinary playground where many can come and contribute.

As a philosophical endeavor, we sought to support our proposal with a collection of metaphysical and metaethical underpinnings, serving as the roots of our work. Every theory or hypothesis has its foundations, and while some authors choose to conceal or merely suggest them as implicitly stated, we aim for the opposite. Consequently, we explicitly articulated all foundational perspectives on how this work conceptualizes intelligence and normativity from a dynamic standpoint and how we constructed a framework called \textcolor{BrickRed}{Dynamic Normativity}. This framework attempts to reconstruct, in a limited way, the human normative experience, which is, to a great extent, a learning experience.

In \textcolor{BrickRed}{Dynamic Normativity}, we hypothesize that a specific set of necessary conditions must hold to address value alignment. Even though insufficient to ensure said alignment, we advocated for their essentiality as irrevocable facts that should be for a learning-based approach toward alignment to succeed.

\begin{enumerate}
    \item Goals are fundamental aspects of intelligent and intentional behavior.
    \item Intentions permeate human behavior.
    \item Normative preferences permeate human intentions.
    \item Through actions, humans impregnate their environment with the preferences they possess.
\end{enumerate}

From these, we postulate that a system that accurately models human intentions and the human environment can indirectly access normative information embedded in them. These are speculations on the capabilities of the learning paradigm and the nature of human intentionality and normative behavior. Following this claim, we proposed a set of sufficient conditions to help us establish a minimum set of criteria for value alignment: 

\begin{enumerate}
    \item Aligned AI systems should coherently aggregate human preferences in a way that resolves cases of uncertainty. Aligning AI systems requires methods to deal with cases of uncertainty.
    \item AI systems can adhere to human preferences if they are an available part of their objective function. Aligning AI systems requires using human preferences as part of their learning signal.
    \item Aligned AI systems should have mechanisms to perform impact mitigation to minimize harmful and unintended consequences. Aligning AI systems requires the specification of safety guardrails.
\end{enumerate}

We argue that no value alignment can come to fruition if the necessary conditions do not hold. Meanwhile, how well a system is aligned depends on how well the sufficient conditions are satisfied. These conditions give us a starting point to build aligned systems from an applied and theoretical perspective. All our developments (the \href{https://huggingface.co/collections/nicholasKluge/aira-657db1563c65a5be2a02f51c}{Aira series})\footnote{\hspace{1mm}\includegraphics[scale=0.025]{img/link.png}\hspace{1mm} \href{https://huggingface.co/collections/nicholasKluge/aira-657db1563c65a5be2a02f51c}{huggingface.co/collections/nicholasKluge/aira-657db1563c65a5be2a02f51c}} are inspired by this proposed framework. 

However, numerous problems persist, and each step of this process is fraught with difficulties. Collecting human data and feedback is a controversial process that follows unethical practices in many circumstances. Aggregating human preferences possess many impossibilities and trade-offs that are still vastly unexplored from an alignment perspective. Value learning, even from high-quality sources of demonstrations and feedback, still produces unwanted behavior and emergent properties that are hard to foresee. Known impact mitigation techniques do not provide a full prof methodology to prevent these undesirable behaviors. Finally, to make everything worse, much of the state-of-the-art work on these problems comes in a way that is not reproducible to much of the community due to (1) resource costs and (2) the closed-way current research is being done.

While lowering the entrance bar in terms of resources and knowledge for specialized fields might not be something we can improve too much on, there is (definitely) something to be said about making our current scenario (especially in Academia) more open in terms of research, development, and sharing of resources. Something that brings the perils of unalignment back to the human sphere.

One of the most significant challenges of alignment is dealing with human alignment. We did not address this factor in this work, but it represents a substantial portion of the problem. Nowadays, and till AI systems gain substantially more autonomy, most of the dangers related to these technologies come from people utilizing them in an undesirable way. Making AI ethics, AI safety, and AI alignment literacy and research obscure or guarded behind doors can not help improve this scenario. At the inception of this work in 2020, there was a dearth of resources and limited general interest in these research areas. However, presently, the community is gradually establishing a field. This work represents a modest interdisciplinary contribution to bridging the knowledge and resource gap in areas where these topics are still inadequately comprehended. Something that remains to the reader to judge if we have been successful or not.

As a famous thinker once said, the most important of all human problems is the "\textit{moral problem}". If this is the most crucial problem, would it be overly audacious to state that the alignment problem is the paramount issue involving AI? If the answer is affirmative, we could say that this is one of the "goods" alignment has brought: turning AI research into a more humane science and a common place to think about interesting things and work together. Something we hope this work can represent.

\chapter*{Acknowledgements}

First and foremost, I would like to extend my most profound appreciation to my family for their endless support and encouragement. A special dedication goes to my grandmother, \textbf{Maria Tereza Dutra Kluge}, who passed away in May 2024.

I also want to thank my advisors, Nythamar de Oliveira and Michael Schulz, for their unwavering support, guidance, and encouragement throughout my doctoral journey. I am also profoundly grateful to the institutions that made this research possible: the University of Bonn and the Pontifical Catholic University of Rio Grande do Sul. I would also like to thank the Coordenação de Aperfeiçoamento de Pessoal Nível Superior – Brazil (CAPES), the Deutscher Akademischer Austauschdienst (DAAD), and the Rede de Inteligência Artificial Ética e Segura (RAIES) for their financial support.

My heartfelt thanks go to my AI Robotics Ethics Society (AIRES) colleagues, who have collaborated with me these last few years. Your collaboration, insights, and camaraderie have been fundamental to realizing many projects I am glad to have been involved in.

Lastly, I want to acknowledge all the wonderful, albeit slightly crazy, individuals who cohabitate with me at the CPP, close to Bonn HBF.

I love you all.

\bibliographystyle{plain}
\bibliography{bibliography}
\end{document}